\RequirePackage{fix-cm}
\documentclass[smallextended]{svjour3}       
\smartqed  
\usepackage{graphicx}
 \journalname{Empirical Software Engineering}

\usepackage{comment}
\usepackage{xcolor}
\usepackage{booktabs}
\usepackage{multirow,bigstrut}
\usepackage{tabu}
\usepackage{cancel}
\usepackage{makecell}
\usepackage{color,soul}
\usepackage[caption=false, font=footnotesize]{subfig}
\usepackage{hyperref}
\newlength\MAX  
\newcommand*\Chart[1]{#1~\rlap{\textcolor{black!20}{\rule{\MAX}{2ex}}}\rule{#1\MAX}{2ex}}
\usepackage[most]{tcolorbox}

\tcbset{
enhanced,
breakable,
left=2mm,
right=2mm,
attach boxed title to top left={
xshift=0.5cm,
yshift=-3mm,
yshifttext=-1mm
},
top=4mm,
colback=blue!5!white,
colframe=blue!100!black,
colbacktitle=blue!100!black,
boxed title style={size=small,colframe=blue!100!black}
}
\usepackage{setspace}
\usepackage{hyperref}
\newenvironment{mybox}[1]{%
\begin{tcolorbox}[title={#1}]%
\setstretch{0.95}}{
\end{tcolorbox}
}
\usepackage{url}

\usepackage{todonotes}
\usepackage{color,soul}

\usepackage{amssymb}

\begin{document}

\title{Enhancing Robustness of AI Offensive Code Generators via Data Augmentation}


\author{Cristina Improta\textsuperscript{*}         \and Pietro Liguori \and Roberto Natella \and Bojan Cukic \and Domenico Cotroneo
}


\institute{Cristina Improta \at University of Naples Federico II, Italy\\
\email{cristina.improta@unina.it} \and
Pietro Liguori \at University of Naples Federico II, Italy\\
\email{pietro.liguori@unina.it} \and
Roberto Natella \at University of Naples Federico II, Italy\\
\email{roberto.natella@unina.it} \and
Bojan Cukic \at University of North Carolina at Charlotte, USA\\
\email{bcukic@charlotte.edu} \and
Domenico Cotroneo \at University of Naples Federico II, Italy\\
\email{cotroneo@unina.it} \and
}

\date{Received: date / Accepted: date}

\maketitle

\begin{flushleft}
\textsuperscript{*}Corresponding author
\end{flushleft}

\begin{abstract}
Since manually writing software exploits for \textit{offensive security} is time-consuming and requires expert knowledge, AI-base code generators are an attractive solution to enhance security analysts' productivity by automatically crafting exploits for security testing. However, the variability in the natural language and technical skills used to describe offensive code poses unique challenges to their robustness and applicability. 
In this work, we present a method to add perturbations to the code descriptions to create new inputs in natural language (NL) from well-intentioned developers that diverge from the original ones due to the use of new words or because they miss part of them. The goal is to analyze how and to what extent perturbations affect the performance of AI code generators in the context of offensive code.
First, we show that perturbed descriptions preserve the semantics of the original, non-perturbed ones. Then, we use the method to assess the robustness of three state-of-the-art code generators against the newly perturbed inputs, showing that the performance of these AI-based solutions is highly affected by perturbations in the NL descriptions. 
To enhance their robustness, we use the method to perform data augmentation, i.e., to increase the variability and diversity of the NL descriptions in the training data, proving its effectiveness against both perturbed and non-perturbed code descriptions.
\end{abstract}

\section{Introduction}
\label{sec:introduction}
With the advent of modern AI technologies, we are witnessing a tremendous increase in using AI code generators, i.e., tools (or part of them) that employ powerful Neural Machine Translation (NMT) models, which promise deep changes in software development processes. These solutions automatically translate natural language (NL) descriptions (\textit{intents}) into programming code (\textit{code snippets})~\cite{tufano2019learning,mastropaolo2021studying}.
Popular examples include CodeBERT~\cite{feng2020codebert} and CodeT5+~\cite{wang2023codet5+}, which are representative of the state-of-the-art in numerous software engineering tasks, including code search and program repair~\cite{zeng2022extensive,DBLP:conf/icse/GaoZGW23}.

The inherent variability of NL often results in code generators producing inaccurate or inefficient code for semantically equivalent descriptions, posing a significant limit to their practical application in real-world development processes. To be used in real-world contexts, thus bridging the gap between NL descriptions and precise code snippets, these tools must be \textit{robust} to this variability and generate the correct code snippet even when they meet different yet semantically equivalent NL code descriptions~\cite{DBLP:conf/icse/MastropaoloPGCSOB23}. 

Improving the robustness of AI code generators is crucial for making software development better and more efficient~\cite{henkel2022semantic,zhang2024rocoins}. This requirement becomes even more relevant when it comes to writing code for \textit{offensive security} activities (e.g., code for performing effective penetration tests).
Offensive security code is particularly challenging because it requires developers to be very accurate and deep understanding of complex technical issues and low-level language aspects, such as the intricacies of memory layout and CPU registers~\cite{liguori2022can,yang2022dualsc,yang2023exploitgen}.
If something goes wrong in the generation of offensive code, it could lead to serious problems, such as leaving important software procedures and/or information vulnerable to attacks. 
This underscores the importance of Automatic Exploit Generation (AEG) as an attractive solution to enhance security analysts’ productivity by automatically creating working exploits for security testing~\cite{botacin2023gpthreats,ruan2023prompt,xu2023autopwn}.

The problem of model robustness becomes even more critical due to the complexity of offensive code descriptions, as they require a strict vocabulary to detail the low-level operations needed to gain access to the system by exploiting memory layout and CPU registers. Indeed, slight variations in technical knowledge, terminology, and specificity among security experts can lead to discrepancies in the NL, altering the whole semantics of the description~\cite{wang2013study}. 

However, the evaluation of the robustness of AI-based solutions in the context of offensive code generation still poses unique challenges. The main issue concerns the choice of \textit{perturbations}, i.e., intentional modifications made to the NL descriptions to test the robustness of AI code generators. Indeed, perturbations commonly used in human languages are often not suitable for code descriptions. It suffices to think that human languages are usually made up of assertions, which are, instead, almost absent from the descriptions of programming languages~\cite{pops}. Moreover, NL descriptions of offensive code can use ad hoc terms that are barely used in human languages (e.g., the names of assembly registers or logical operators) or are used with different meanings (e.g., the word \textit{label}). Therefore, they should be treated differently to preserve the meaning of the descriptions.

This paper proposes a novel method to assess and enhance the robustness of AI offensive code generators. Our approach reproduces the natural variability of how different developers describe code by introducing perturbations into the descriptions of offensive code. More precisely, we modify the original code descriptions by introducing new perturbed NL descriptions that diverge from the original ones due to the use of new terms (\textit{word substitution}) or because they miss part of them (\textit{word omission}). These perturbations are designed to maintain the original meaning while testing the AI's ability to handle variations in input. 
Our findings demonstrate that the generated perturbed descriptions successfully preserve the intent of the original code descriptions. We then assess the impact of these variations on the code generators' performance, focusing on the \textit{syntactic} and \textit{semantic} correctness of the code generated by the models. 

Additionally, we analyze the use of these perturbed descriptions for \textit{data augmentation}, aiming to diversify training data and thereby enhance the models' robustness. 
Data augmentation is an effective solution to increase the volume but especially the diversity of the data used to train AI models, in order to improve their performance on unseen data. In our work, we employ the data augmentation strategy with the primary objective of increasing the diversity of training examples without collecting new data.
To the best of our knowledge, this is the first study to apply word-level perturbations to NL descriptions of offensive code as a means to test and improve the resilience of AI code generators.
We validate our approach through experiments with three state-of-the-art models\footnote{The dataset, experimental results, and code to replicate the experiments are available at the following URL: \url{https://github.com/dessertlab/Robustness-of-AI-Offensive-Code-Generators/}}, one non--pre-trained Seq2Seq model and two pre-trained models, i.e., CodeBERT and CodeT5+. To fine-tune them for the generation of offensive code, we extended the
publicly available Shellcode IA32 dataset for automatically generating shellcodes from NL descriptions~\cite{liguori2021shellcode_ia32}. The results of our analysis provide the following key findings:

\begin{enumerate}
     \item \textbf{Vulnerability to NL Perturbations}: The three models used in our experiments are vulnerable to new, perturbed code descriptions. Indeed, the performance of the models decreases in terms of syntactic correctness ($\sim10\%$ and $\sim18\%$ for word substitution and word omission on average, respectively) and semantic correctness ($\sim20\%$ and $\sim34\%$ on average).
    
    \item \textbf{Robustness Enhancement through Data Augmentation}: Augmenting training data with perturbed descriptions significantly boosts model performance, improving both syntax correctness (up to $\sim17\%$ for word substitution, and $\sim27\%$ for word omission) and semantic correctness (up to $\sim25\%$ and $\sim16\%$).
    
    \item \textbf{General Performance Improvement}: Augmentation not only increases robustness to perturbations but also enhances overall performance on non-perturbed (original) code descriptions, with improvements in semantic correctness of up to $\sim 5\%$ and up to $\sim 4\%$ when we use word substitution and word omission in the training data, respectively.
\end{enumerate}

In the following, 
Section~\ref{sec:related} discusses related work;
Section~\ref{sec:methodology} presents the proposed method; 
Section~\ref{sec:case_study} describes our case study;
Section~\ref{sec:semantics} analyses the semantics of the perturbed descriptions;
Section~\ref{sec:results} shows the robustness assessment of the models;
Section~\ref{sec:threats} discusses threats to validity;
Section~\ref{sec:limitation} discusses the limitations and future work;
Section~\ref{sec:conclusion} concludes the paper.

\section{Related Work}
\label{sec:related}
\noindent
\textbf{AI Offensive Code Generators.}
Automatic exploit generation (AEG) research challenge consists of automatically generating working exploits~\cite{avgerinos2014automatic}. 
This task requires technical skills and expertise in low-level languages to gain full control of the memory layout and CPU registers and attack low-level mechanisms (e.g., heap metadata and stack return addresses) not otherwise accessible through high-level programming languages. 
Given their recent advances, AI-code generators have become a new and attractive solution to help developers and security testers in this challenging task~\cite{COTRONEO2024112113}.
Although these solutions have shown high accuracy in the generation of software exploits, their robustness against new inputs has never been studied before.
Liguori \textit{et al.}~\cite{liguori2022can} released a dataset containing NL descriptions and assembly code extracted from software exploits. They performed an empirical analysis showing that NMT models can correctly generate assembly code snippets from NL and that in many cases can generate entire exploits with no errors. 
The authors extended the analysis to the generation of Python offensive code used to obfuscate software exploits from systems' protection mechanisms~\cite{liguori2021evil}.
Yang \textit{et al.}~\cite{yang2022dualsc} proposed a data-driven approach to software exploit generation and summarization as a dual learning problem. The approach exploits the symmetric structure between the two tasks via dual learning and uses a shallow Transformer model to learn them simultaneously.
Yang \textit{et al.}~\cite{yang2023exploitgen} proposed a novel template-augmented exploit code generation approach. The approach uses a rule-based template parser to generate augmented NL descriptions and uses a semantic attention layer to extract and calculate each layer’s representational information. 
Ruan \textit{et al.}~\cite{ruan2023prompt} proposed \textit{PT4Exploits}, an approach for software exploit generation via prompt tuning. They designed a prompt template to build the contextual relationship between English comment and the corresponding code snippet, simulating the pre-training stage of the model to take advantage of the prior knowledge distribution. 
Xu \textit{et al.}~\cite{xu2023autopwn} introduced an artifact-assisted AEG solution that automatically summarizes the exploit patterns from artifacts of known exploits and uses them to guide the generation of new exploits. The authors implemented \textit{AutoPwn}, an AEG system that automates the generation of heap exploits for Capture-The-Flag \textit{pwn} competitions. 
Recent work also explored the role of GPT-based models, including ChatGPT and Auto-GPT, in the offensive security domain. Botacin~\cite{botacin2023gpthreats} found that, by using these models, attackers can both create and deobfuscate malware by splitting the implementation of malicious behaviors into smaller building blocks.
Pa \textit{et al.}~\cite{pa2023attacker} and \cite{gupta2023chatgpt} proved the feasibility of generating malware and attack tools through the use of reverse psychology and \textit{jailbreak prompts}, i.e., maliciously crafted prompts able to bypass the ethical and privacy safeguards for abuse prevention of AI code generators like ChatGPT.
Gupta \textit{et al.}~\cite{gupta2023chatgpt} also examined the use of AI code generators to improve security measures, including cyber defense automation, reporting, threat intelligence, secure code generation and detection, attack identification, and malware detection.

\vspace{0.1cm}
\noindent
\textbf{Adversarial Inputs.}
Previous work investigated the use of perturbations at different linguistic levels (i.e., character, word, and sentence levels) to address the security concerns raised by the ML models, focusing on \textit{adversarial attacks}, i.e., inputs that are specifically designed to mislead the models, rather than well-intentioned perturbations.
Character-level perturbations include homograph attacks, where characters are replaced with homoglyphs, i.e., characters that render the same or a visually similar glyph (e.g., the lowercase Latin letter ‘l' and the uppercase Latin letter ‘I'), which are used to mislead models in question-answer problems~\cite{wu2020evaluating} or to reproduce keyboard typos~\cite{belinkov2018synthetic}.
Boucher \textit{et al.}~\cite{boucher_2022_badchars} proposed a class of encoding-based attacks using invisible characters, control characters, and homoglyphs, which are imperceptible to human inspection and can strongly harm machine learning systems. 
Heigold \textit{et al.}~\cite{heigold-etal-2018-robust} studied the effects of word scrambling and random noise insertion in machine translation, focusing on English and German languages. Their perturbation strategies include character flips and swaps of neighboring characters to imitate typos.
At the word level, words in a sentence can be substituted with different random words, similar words in the word embedding space (i.e., real-valued vector representation of the words), or meaning-preserving words \cite{li2018textbugger,michel2019evaluation,huang2021robustness}.
Paraphrases, back translation, and reordering are some of the approaches to produce syntactically and semantically similar sentences to fool models~\cite{huang2021robustness}.

\vspace{0.1cm}
\noindent
\textbf{Models' Robustness.} Recent work introduced several tools for the generation of adversarial inputs to assess how they impact the performance of the models in the generation of NL, but never targeted the code descriptions.
As an example, Cheng \textit{et al.}~\cite{cheng2020seq2sick} proposed \textit{Seq2Sick}, an optimization-based framework to generate adversarial examples for sequence-to-sequence neural network models. 
Michel \textit{et al.}~\cite{michel2019evaluation} presented \textit{TEAPOT}, a toolkit to evaluate the effectiveness of adversarial perturbations on NMT models by taking into account the preservation of meaning in the source.
\textit{TextBugger}~\cite{li2018textbugger} is a framework to generate utility-preserving adversarial texts against real-world online text classification systems. 
Adversarial attacks have also been investigated against the code.
These works focused on perturbing code snippets to mislead models on several code-code and code-NL tasks across different programming languages.
\textit{ALERT}~\cite{yang2022natural} is a black-box attack that transforms code snippets to make pre-trained models fail predictions in three software engineering tasks.  
\textit{ACCENT}~\cite{DBLP:journals/tosem/ZhouZSHCG22} is an approach that crafts adversarial code snippets to mislead the models to produce completely irrelevant code comments.
\textit{DAMP}~\cite{DBLP:journals/pacmpl/Yefet0Y20} is an approach to generate attacks on models of code by renaming variables and adding dead code. \textit{CodeAttack}~\cite{jha2022codeattack} is a black-box attack model that generates imperceptible adversarial code samples by leveraging the code structure.
In the field of code generation from NL, Mastropaolo \textit{et al.}~\cite{DBLP:conf/icse/MastropaoloPGCSOB23} investigated the robustness of a public model against paraphrased descriptions, showing that semantically equivalent code descriptions can result in different outputs.
Zhu and Zhang~\cite{10123534} proposed a white-box attack to measure the robustness of a pre-trained model in the generation of code solutions for programming problems, showing that the model is affected by character-level perturbations.
Zhuo \textit{et al.}~\cite{zhuo-etal-2023-robustness} investigated the robustness of a prompt-based semantic parser in text-to-SQL tasks and proposed adversarial in-context learning, i.e., contextual guidance during a given task or interaction, as a means to improve robustness.
Wang \textit{et al.}~\cite{wang-etal-2023-recode} proposed \textit{ReCode}, a
robustness evaluation benchmark for code generation models, including CodeGen, InCoder, and GPT-J. They 
perform different perturbations on docstrings, function names, code syntax, and code format, and evaluate their benchmark using HumanEval and MBPP datasets.


\vspace{0.1cm}
\noindent
\textbf{Data Augmentation.}
To enhance the robustness of the models, data augmentation has proven to be an effective solution in numerous NL processing tasks.
However, it has never been applied in the specific task of code generation due to the difficulty of manipulating intents that use ad hoc terminology and precise structure to describe code snippets.
Wang \textit{et al.}~\cite{wang2018switchout} designed a data augmentation algorithm as an optimization problem, where they seek the policy that maximizes an objective that encourages both smoothness and diversity of the examples. 
Gao \textit{et al.}~\cite{gao2019soft} proposed to augment NMT training data by replacing a randomly chosen word in a sentence with a\textit{soft word}, which is a probabilistic distribution over the vocabulary and is computed based on the contextual information of the sentence.
Nguyen \textit{et al.}~\cite{nguyen2020data} trained multiple models on forward translation, i.e., source-to-target, and on backward translation, i.e., target-to-source, and used them to generate a diverse set of synthetic training data from both lingual sides.
Wei and Zou~\cite{wei2019eda} presented an effective \textit{easy data augmentation} strategy to reduce the problem of overfitting in text classification tasks. They improved the performance of models trained on smaller datasets by performing synonym replacement and random insertion, swap and deletion of words on different subsets of training examples.
Yu \textit{et al.}~\cite{yu2022data} proposed a solution to improve the models' generalization ability in the software engineering domain. They designed program transformation rules that can preserve both the semantics and the syntactic naturalness of code snippets to increase the accuracy of tasks such as method naming and code clone detection.
In the context of code comment generation, Zhang \textit{et al.}~\cite{zhang2020training} employed the Metropolis-Hastings modifier (MHM) algorithm to augment the training data by perturbing source code and renaming identifiers, thereby enhancing model robustness.


Our study centers on the specific challenge of generating exploits from NL descriptions, a focus that distinguishes it from broader code generation research.
Indeed, to the best of our knowledge, no prior work tackles the challenge of perturbing NL descriptions of exploit code. 
Existing work leverages synonym substitution and word deletion to augment textual data~\cite{wei2019eda,huang2021robustness}. We drew these ideas and tailored them to the task at hand by ensuring the substitution with context-appropriate synonyms and the omission of significant information (i.e., offensive code-related words). By focusing on the challenging domain of software exploit generation and introducing a data augmentation method, our study offers unique insights and contributions that set it apart from previous work. 
Differently from closely related work such as that of Mastropaolo \textit{et al.}~\cite{DBLP:conf/icse/MastropaoloPGCSOB23},  our work focuses on the generation of software exploits from natural language descriptions, which is an even more challenging task than the broader scope of automatic code generation. The generation of offensive code, particularly for low-level programming languages like assembly, demands a high degree of technical expertise and the use of specialized vocabulary. This complexity makes evaluating model robustness more difficult, as it requires creating meaningful perturbations that maintain the original description's semantics. Unlike general-purpose code generation, which can rely on simple paraphrases, our task involves preserving the technical terminology used in shellcode generation, such as register names and low-level operations.
Furthermore, our work extends beyond assessing the robustness of code-generation techniques. We have designed a novel and effective data augmentation strategy that improves the models’ ability to handle varied inputs and enhances their robustness in generating offensive code.

\section{Method}
\label{sec:methodology}

\begin{figure*}[t]
  \centering
  \includegraphics[width=1\columnwidth]{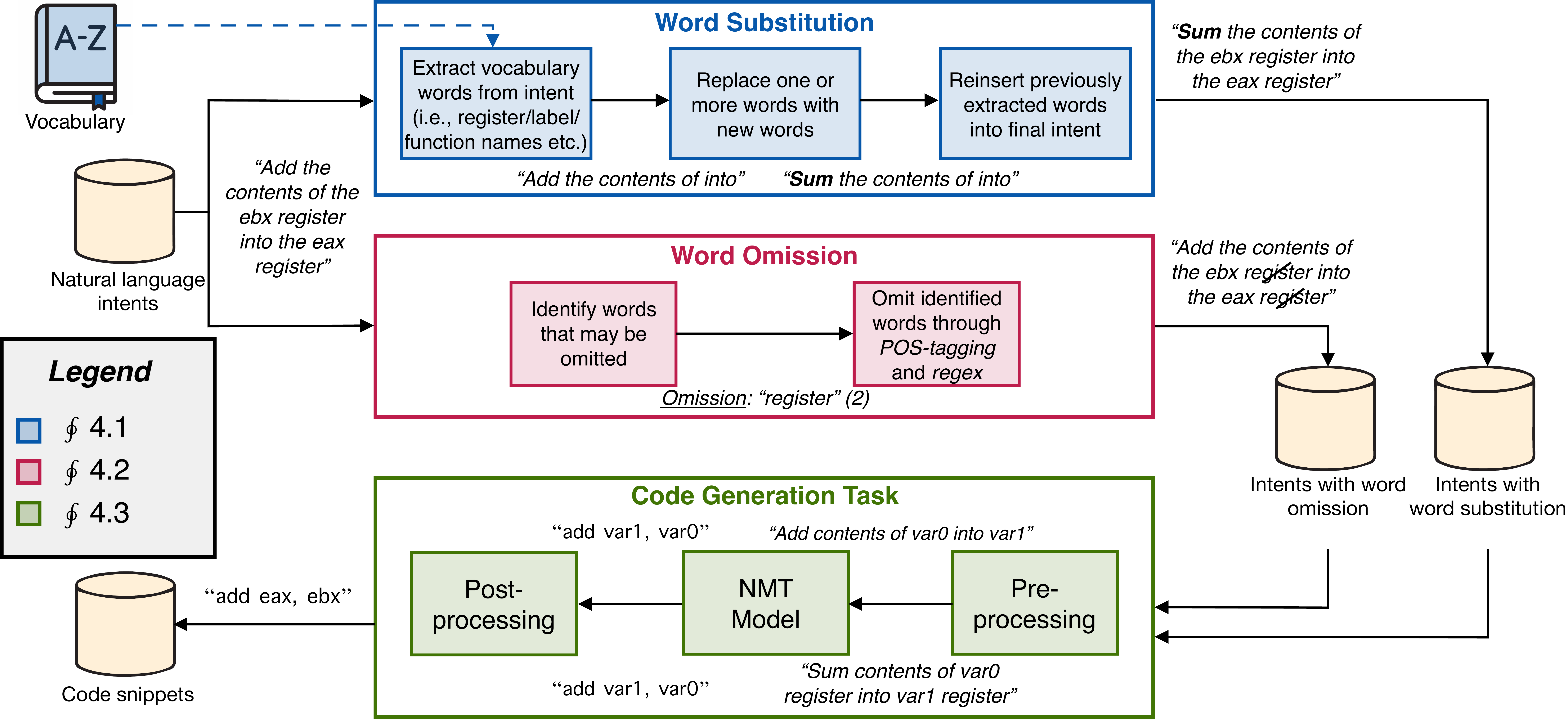}
  \caption{Overview of the proposed method.}
  \label{fig:methodology}
\end{figure*}

The challenge of evaluating the robustness of AI offensive code generators lies in selecting meaningful perturbations, issue that is further exacerbated by the purpose of the code to describe. Indeed, unlike regular high-level programming languages that focus on logically complex functional code fragments, offensive code often contains a large number of low-level arithmetic and logical operations. It leverages machine-level programming languages, such as assembly, to perform surgically crafted exploitation of the system's internals, including heap metadata and stack return addresses, that are not accessible through high-level programming languages.

For this reason, the NL used to describe the code snippets can not only be variable, depending on the writing style and technical expertise of the developers, but also contain ad hoc terminology that is rarely used when describing general-purpose code, such as register names and low-level operations.
Moreover, a sentence can be expressed using different words, ordering words in different ways, lacking significant details, or being too specific.

Therefore, to assess the robustness of AI offensive code generators, we present a method that perturbs the original code descriptions to create new, equivalent NL descriptions, taking into account the offensive nature of the described code. \figurename{}~\ref{fig:methodology} shows the overview of the proposed method. 

To add perturbations to the NL intents of the corpora, we choose types of perturbation reflecting both the variability and ad hoc vocabulary of the NL. 

A robust model should be resistant to this variability and be able to predict the same output when dealing with two different but equivalent code descriptions.
Therefore, to measure the robustness of the code generators, we define a set of perturbations to the models' inputs. 

Although character-level perturbations can be adopted to study the sensitivity of the models to human errors (e.g., typos), in this work we focus on perturbing words but still preserving the original meaning of the NL intents. To identify the perturbation types that best reproduce this variability, we look at the intents and analyze how each word contributes to the meaning of the whole sentence based on its part of speech. 

\begin{figure}[t]
  \centering
  \includegraphics[width=0.75\columnwidth]{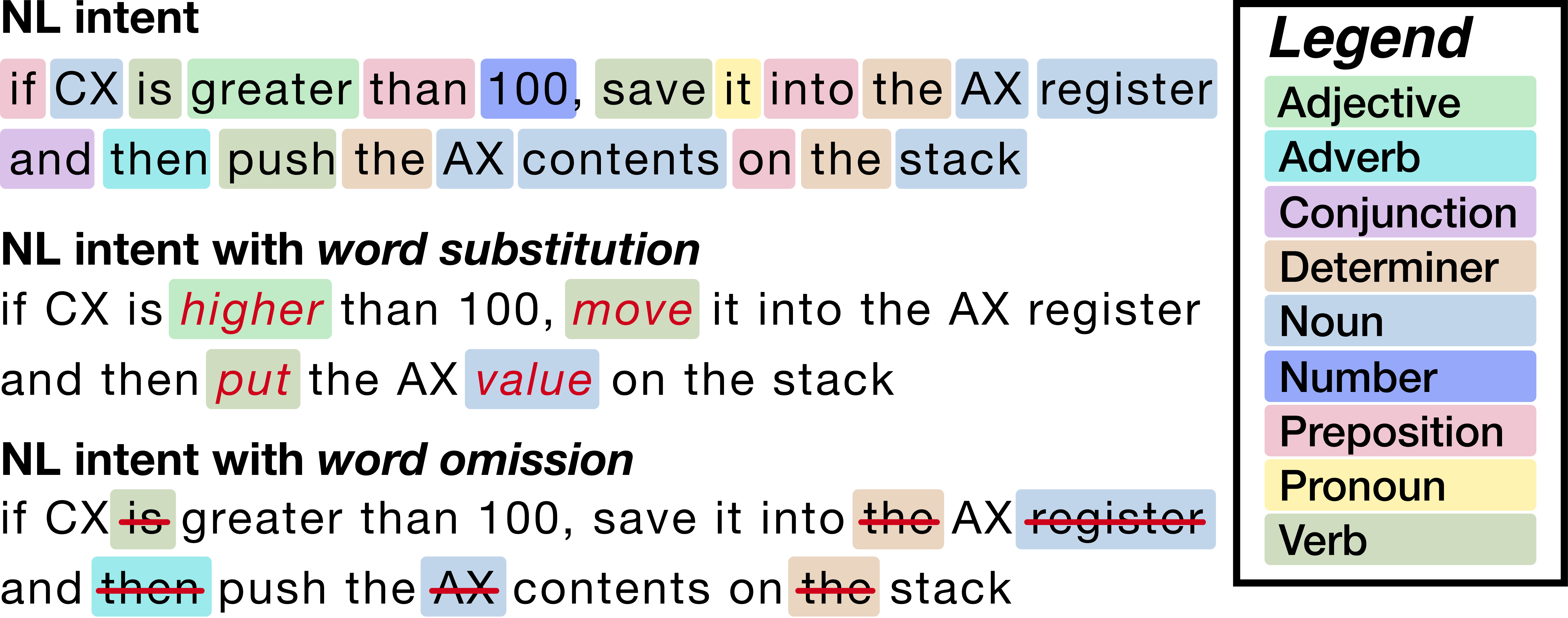}
  \caption{Example of NL description for assembly code. Equivalent intents can be expressed with different words (\textit{word substitution}) or by omitting some words (\textit{word omission}).}
  \label{fig:motivation}
\end{figure}

Consider the example in \figurename{}~\ref{fig:motivation}, which shows a long and detailed NL intent used to describe assembly code extracted from a software exploit program for the \textit{IA-32} (the 32-bit version of the x86 Intel Architecture). The intent contains nine \textit{part-of-speech} (POS), i.e., the category of words that have similar grammatical properties, in the English grammar\footnote{For this example, we used \href{https://parts-of-speech.info/}{Parts-of-speech.Info}, a POS tagger based on the Stanford University Part-Of-Speech Tagger.}: \textit{noun} (AX, register, contents, stack), \textit{verb} (is, save, push), \textit{adjective} (greater), \textit{adverb} (then), \textit{pronoun} (it), \textit{preposition} (if, than, into, on), \textit{conjunction} (and), \textit{determiner} (the) and \textit{number} (100).
The figure shows that we can express a semantically equivalent intent by using different NL descriptions.  
Indeed, POS such as verbs and nouns can be, in some cases, substituted with new words, while they can be omitted in others; adjectives can be replaced, but not omitted (i.e., the word \textit{greater} is fundamental to this description); adverbs and determiners can be omitted without altering the sentence.

Based on these assumptions, we focus on two types of perturbations: \textit{word substitution}~(\S{}~\ref{subsec:word_substitution}), and the \textit{word omission}~(\S{}~\ref{subsec:word_omission}). The former can be used to evaluate the performance of code generation when descriptions diverge from the lexicon used in the training corpus due to variability in the cultural and technical background knowledge of developers~\cite{wang2013study}. 
The latter, instead, allows us to assess the models' performance when developers omit information that would be redundant, such as information implicitly contained in the sentence or words already stated in previous intents (e.g., in a large program where the same operand or variable is accessed by multiple instructions in sequence, it would be redundant to refer to that operand/variable multiple times), or simply when the NL intents do not adequately describe the behavior of the code as they lack details~\cite{panichella2012mining}.

The new set of perturbed NL intents is used as input to assess the robustness of the models in the generation of the code snippets~(\S{}~\ref{subsec:code_generation}). Ideally, the performance of the models against the perturbed data should be comparable to the ones obtained against the original, non-perturbed data, otherwise, the code generator is not robust to the variability of the NL descriptions.

\subsection{Word Substitution}
\label{subsec:word_substitution}
Substituting words is not a trivial task, especially in a programming context. In fact, blindly replacing words may lead to the loss of the original meaning of the code description if we do not consider the context of code generation. For example, consider the NL intent ``\textit{clear the contents of the EAX register}'', which describes an assembly snippet that zeros out a register. A valid perturbation of the input can reasonably lead to the sentence ``\textit{empty the contents of the register}'', but not to ``\textit{purify the contents of the register}'' since the verb \textit{purify} is out of the programming context. 
This problem is further exacerbated by the use of words that belong only to the programming context (e.g., \texttt{EAX}) that cannot be substituted.

To address these issues, we impose a set of \textit{constraints} on the word substitution process.
First, we consider the set of the \textit{top-k nearest neighbors} in the counter-fitted embedding space~\cite{mrksic:2016:naacl}, i.e., a vector space constructed by injecting antonym and synonym constraints into real-valued vector representations to improve the vectors' ability to judge semantic similarity. Counter-fitted word vectors allow us to find better candidates for word substitution than regular word vectors as a result of nearest-neighbor searches~\cite{yoo-qi-2021-towards-improving}. Then, to select the most suitable candidate, we compute the \textit{cosine similarity}, which measures the similarity between two-word vectors by computing the cosine of the angle between them~\cite{li-etal-2021-searching,DBLP:journals/corr/abs-2109-07403,DBLP:journals/corr/abs-2011-03901,yoo-qi-2021-towards-improving}. The metric ranges between $0$ (total dissimilarity) and $1$ (total similarity).
Additionally, we use the POS tag to further constrain the transformation, that is, we mark each word with its corresponding part of speech~\cite{jin2020bert} to limit the substitution to words that share the same POS tag (e.g., a verb can be replaced with another verb, but not with a noun). 
Therefore, we replace the original word with a different one only if they have a cosine similarity value above a specific threshold and the same POS tag. 
We set the cosine similarity threshold to $0.8$ to guarantee a good trade-off between the quality and strength of the generated perturbed example~\cite{li2018textbugger} and set to $20$ the number of nearest-neighbors~\cite{yoo-qi-2021-towards-improving}.
By enforcing these constraints, we replace words with semantically and syntactically similar words, hence ensuring that the examples are perceptibly similar~\cite{alzantotSEHSC18}. 

\tablename{}~\ref{tab:unseen} shows how the use of constraints contributes to preserving the original meaning in the perturbed intent (the intent is chosen intentionally as simple for the example). In the original sentence ``\textit{Store the shellcode pointer in the ESI register}'', the word \textit{store} clearly expresses the action of memorizing a pointer in a specific register. However, if we perform the substitution without using the constraints, the verb can be interpreted as a noun and be replaced with a similar noun, i.e., \textit{stock}, producing an incoherent sentence. 
When we use constraints, instead, the word is replaced with the verb \textit{save}, resulting in a semantically equivalent description.

\begin{table}[t]
\centering
\caption{Examples of the word substitution on the intents. \underline{Underlined} text refers to the candidate word replacing the original one.}
\label{tab:unseen}
\footnotesize
\begin{tabular}
{ >{\centering\arraybackslash}m{3cm} |   >{\centering\arraybackslash}m{7cm}}
\toprule
\textbf{Word Substitution}           &   \textbf{NL Intent}\\ \midrule
\textit{None} (Original Intent)                   & \textit{Store the shellcode pointer in the ESI register.}\\\midrule
w/o constraints    & \textit{\underline{Stock} the shellcode pointer in the ESI register.}\\\midrule
with constraints   & \textit{\underline{Save} the shellcode pointer in the ESI register.}\\ \bottomrule
\end{tabular}
\end{table}

A problem to deal with when we substitute words in the code descriptions is to replace words related to the programming language that cannot be expressed with different terms (e.g., the word \textit{class}, the names of the variables, etc.).   
To overcome this issue, we use a \textit{vocabulary} of programming--language-related words that cannot be substituted. We build the vocabulary by comparing the NL intents of the corpus used to train the models and a large comparison corpus not containing any text from the programming field (e.g., contemporary prose), with both corpora written in the same NL (e.g., English). 
The key idea is to add to the vocabulary all the words that occur more often in the corpus for code generation than in the comparison corpus or that only occur in the code generation corpus, regardless of their occurrence~\cite{chung2004identifying,liu2019technical}.

To this end, we first count the number of unique words in both the code generation and comparison corpora, excluding \textit{stopwords} (i.e., articles, conjunctions, etc.)~\cite{1223656}. Then, for both corpora, we compute the ratio of the occurrences of every word to the number of unique words in the corpus. If the ratio in the code generation corpus is at least $50$ times higher than the ratio in the comparison corpus, or if the word never appears in the comparison corpus (i.e., ratio = 0), then we add the word to the vocabulary. 
Indeed, \textit{structure-related words}, i.e., words describing the structure of the code such as \textit{function}, \textit{operand}, \textit{variable}, etc., are frequently mentioned in the code generation corpus, and rarely (or never) appear in the comparison corpus. 
Moreover, \textit{name-related words}, i.e., words containing the names of labels, functions, variables, words between brackets or containing special characters, etc., even if used just in a few instructions of the code generation corpora, are unlikely to feature in the comparison corpus. 

Once we build the vocabulary, we extract all the programming language--related words contained in the vocabulary from each NL intent, leaving only the subset of words eligible for substitution. Then, one or more words in this subset are swapped with a new one using counter-fitting vectors and applying constraints on the transformation. Finally, the words previously extracted are reinserted into the NL intents, resulting in the final perturbed intents.

\subsection{Word Omission}
\label{subsec:word_omission}
Deleting information from NL intents is useful for analyzing how the behavior of the model and the comprehension of text vary when part of the information is omitted or can be derived from previous sentences. As shown in the example of \figurename{}~\ref{fig:motivation}, nouns, verbs, adverbs, and determiners can be removed under specific circumstances without losing the original meaning of the sentence. 
We focus on the omission of nouns and verbs since adverbs (e.g., \textit{then}) and determiners (e.g., \textit{the}) are noninformative and usually removed during data preprocessing in the code generation task~\cite{1223656}.

To explain the process of word omission, consider the simple intent ``\textit{copy 0x4 into the BL register}''. The intent preserves its semantic meaning also if we remove the verb \textit{copy}. This type of construct, often used to describe code, is known as a \textit{noun phrase}~\cite{pops}. 
Other examples of noun phrases in programming are ``\texttt{2 > 3}'' or ``\textit{if result not negative}'', in which the verb is omitted.
This NL intent can be further reduced by omitting the noun \textit{register}, which is implied since \texttt{BL} is, in fact, a register. 
Furthermore, if the previous intent is preceded by the NL intent ``\textit{zero out BL}'', we can again avoid specifying the name of the register, leading to the sentence ``\textit{copy 0x4 into the register}''. 
However, we cannot omit both nouns from this intent without resulting in a meaningless sentence (i.e. ``\textit{copy 0x4 into}''). For this reason, we need to constrain the perturbation process based on the role each noun has in the phrase. 

Based on the considerations stated above, we identify three main categories of words that can be omitted without losing the original meaning: \textit{i)} the \textit{action-related words}, i.e., verbs, which contain information on the actions of the intents (e.g., \textit{is}, \textit{define}, \textit{push}, etc.); \textit{ii)} the \textit{structure-related words}, i.e., nouns depending on the structure of the target programming language (e.g., the words \textit{function}, \textit{variable}, \textit{register}, etc.); and \textit{iii)} \textit{name-related words}: i.e., words containing names of variables, functions, classes, labels, registers, etc.
\tablename{}~\ref{tab:missing} shows the different types of word omission on the intent ``\textit{Store the shellcode pointer in the ESI register}''.
In this example, the intent still preserves its meaning even without specifying the omitted words. The verb \textit{store} and the word \textit{register} are implicit (the pointer of the shellcode can only be moved to \texttt{ESI}, which is a register), while the name of the register can be derived from the context of the program (\texttt{ESI} is commonly used to store the shellcode).

To omit words from the NL intents, we use POS tagging to identify action-related words (i.e., verbs) and use the same vocabulary of programming language-related words used to avoid word substitution (described in \S{}~\ref{subsec:word_substitution}) to identify structure-related and name-related words.
For each word omission category, we generate a different version of perturbed intent when possible. For example, if we identify action-related, structure-related and name-related words in the original NL intent, then we generate three perturbed NL intents, one for each category, but we never omit multiple word categories from the same NL intent to avoid altering the meaning of the original code description.

\begin{table}[t]
\centering
\caption{Examples of word omission on the same intent.
\cancel{Slashed} text refers to the omitted words.}
\label{tab:missing}
\footnotesize
\begin{tabular}
{ >{\centering\arraybackslash}m{3.5cm} |   >{\centering\arraybackslash}m{7cm}}
\toprule
\textbf{Word Omission}           &   \textbf{Intent}\\ \midrule
\textit{None} (Original Intent) &   \textit{Store the shellcode pointer in the ESI register}\\ \midrule
Action-related Words   &   \textit{\cancel{Store} the shellcode pointer in the ESI register}\\ \midrule
Structure-related Words &   \textit{Store the shellcode pointer in the ESI \cancel{register}}\\  \midrule
Name-related Words    &   \textit{Store the shellcode pointer in the \cancel{ESI}} register\\ \bottomrule
\end{tabular}
\end{table}

\subsection{Code Generation Task}
\label{subsec:code_generation}
We use NMT models to generate code snippets starting from the NL intents perturbed with word substitution and word omission. 
To perform the generation of offensive code, we follow the best practices in the field by supporting the models with \textit{data processing} operations. The data processing steps are usually performed both before translation (\textit{pre-processing}), to train the model and prepare the input data, and after translation (\textit{post-processing}), to improve the quality and the readability of the code in output.

The preprocessing starts with the \textit{stopwords filtering}, i.e., we remove a set of custom-compiled words (e.g., \textit{the}, \textit{each}, \textit{onto}) from the intents to include only relevant data for machine translation. 
Next, we use a \textit{tokenizer} to break the intents into fragments of text containing space-separated words (i.e., the \textit{tokens}). 
To improve the performance of machine translation~\cite{li2018named,modrzejewski2020incorporating,liguori2021evil}, we \textit{standardize} the intents (i.e., we reduce the randomness of the NL descriptions) by using a \textit{named entity tagger}, which returns a dictionary of \textit{standardizable} tokens, such as specific values, names of labels, and parameters, extracted through regular expressions. We replace the selected tokens in every intent with ``\textit{var}\#", where ``\#" denotes a number from $0$ to \textit{$|l|$}, and $|l|$ is the number of tokens to standardize.
Finally, the tokens are represented as real-valued vectors using \textit{word embedding}.  
Pre-processed data is used to feed the model (\textit{model's training}). Once the model is trained, we perform the code generation from the NL intents. Therefore, when the model takes as inputs new intents, it generates the related code snippets based on its knowledge (\textit{model's prediction}).
Finally, the code snippets predicted by the models are processed (\textit{post-processing}) to improve the quality and readability of the code. The dictionary of standardizable tokens is used in the \textit{de-standardization} process to replace all the ``\textit{var}\#" with the corresponding values, names, and parameters. Moreover, code snippets are cleaned using regular expressions to remove any extra characters or spaces.

\section{Case Study}
\label{sec:case_study}
\subsection{AI Code Generators}
To assess the robustness of the models, we consider three architectures, a standard Seq2Seq model with attention mechanisms, and two pre-trained models, CodeBERT and CodeT5+, which are representative of the state-of-the-art~\cite{DBLP:conf/icse/GaoZGW23} and demonstrated efficacy in software engineering-related tasks. 
For instance, CodeT5+ 220M has been shown to outperform several larger decoder-only models in the text-to-code generation task~\cite{wang2023codet5+}.

\noindent
$\blacksquare$ \textbf{Seq2Seq} is a model that maps an input of sequence to an output of sequence. 
Similar to the encoder-decoder architecture with attention mechanism \cite{bahdanau2014neural}, we use a bi-directional LSTM as the encoder to transform an embedded intent sequence into a vector of hidden states with equal length. 
We implement the Seq2Seq model using \textit{xnmt}~\cite{neubig18xnmt}. 
We use an Adam optimizer \cite{kingma2015adam} with $\beta_1=0.9$ and $\beta_2=0.999$, while the learning rate $\alpha$ is set to $0.001$. We set all the remaining hyper-parameters in a basic configuration: layer dimension = $512$, layers = $1$, epochs = $200$, beam size = $5$.

\noindent
$\blacksquare$ \textbf{CodeBERT}~\cite{feng2020codebert} is a large multi-layer bidirectional Transformer architecture~\cite{vaswani2017attention} pre-trained on millions of lines of code across six different programming languages. 
Our implementation uses an encoder-decoder framework where the encoder is initialized to the pre-trained CodeBERT weights. 
The encoder follows the RoBERTa architecture~\cite{DBLP:journals/corr/abs-1907-11692}, with $12$ attention heads,  hidden layer dimension of $768$, $12$ encoder layers, and $514$ for the size of position embeddings. We set the learning rate $\alpha = 0.00005$, batch size = $32$, and beam size = $10$.
We employ CodeBERT with a transformer decoder, composed of $6$ stacked layers, to enable text-to-text translation tasks. Although CodeBERT alone does not fit the NMT category, its integration with a decoder during the fine-tuning phase effectively performs functions analogous to those of encoder-decoder models~\cite{ahmed2024automatic}.

\noindent
$\blacksquare$ \textbf{CodeT5+}~\cite{wang2023codet5+} is a new family of Transformer models pre-trained with a diverse set of pretraining tasks including causal language modeling, contrastive learning, and text-code matching to learn rich representations from both unimodal code data and bimodal code-text data. 
We utilize the variant with model size $220M$, which is trained from scratch following T5’s architecture~\cite{DBLP:journals/jmlr/RaffelSRLNMZLL20}. It has an encoder-decoder architecture with $12$ decoder layers, each with $12$ attention heads and hidden layer dimension of $768$, and $512$ for the size of position embeddings. We set the learning rate $\alpha = 0.00005$, batch size = $16$, and beam size = $10$.

During data pre-processing, we tokenize the NL intents using the \textit{nltk word tokenizer}~\cite{bird2006nltk} and code snippets using the Python \textit{tokenize} package~\cite{tokenize}. 
We use \emph{spaCy}, an open-source, NL processing library written in Python and Cython~\cite{spacy}, to implement the named entity tagger for the standardization of the NL intents.
We trained the models for a total of 3 epochs. To mitigate the risk of overfitting, we utilized a validation set to evaluate the model's performance after each epoch. This approach allowed us to monitor the model's generalization ability and ensure it performed well on unseen data.
Additionally, we utilized a learning rate scheduler that adjusted the learning rate during training. This adjustment contributed to the stabilization of the training process. The choice of training for 3 epochs was informed by preliminary experiments, which demonstrated that this configuration provided an optimal balance between training time and model performance.

\subsection{Dataset}
\label{subsec:dataset}
To fine-tune the models for the generation of offensive code, we extended the publicly available \textit{Shellcode\_IA32} dataset for automatically generating \textit{shellcodes} from NL descriptions~\cite{liguori2021shellcode_ia32}. 
A shellcode is a list of machine code instructions to be loaded in a vulnerable application at runtime. 
The traditional way to develop shellcodes is to write them using the assembly language, and by using an assembler to turn them into \emph{opcodes} (operation codes, i.e., a machine language instruction in binary format, to be decoded and executed by the CPU) \cite{foster2005sockets,megahed2018penetration}. 
Common objectives of shellcodes include spawning a system shell, killing or restarting other processes, causing a denial-of-service, etc.

The dataset consists of instructions in assembly language for \textit{IA-32} collected from publicly available security exploits~\cite{exploitdb,shellstorm}, manually annotated with detailed English descriptions. In total, it contains $3,200$ unique pairs of assembly code snippets/English intents. 
We further enriched the dataset with additional up-to-date samples of shellcodes collected from publicly available security exploits, reaching $5,900$ unique pairs of assembly code snippets/English intents. To the best of our knowledge, the resulting dataset is the largest collection of offensive code available to date for code generation.
Table~\ref{tab:multi_line} presents two examples of offensive code snippets drawn from our dataset to illustrate the nature of the software exploits and their NL descriptions.

\begin{table}[t]
\caption{Examples of assembly code with NL descriptions from our dataset.}
\label{tab:multi_line}
\footnotesize
\begin{tabular}
{>{\centering\arraybackslash}m{5.5cm} | 
>{\centering\arraybackslash}m{5.5cm}}
\toprule
\textbf{Code Snippet} & \textbf{English Intent} \\ \midrule
\texttt{xor bl, 0xBB} \textbackslash{n} \texttt{jz formatting} \textbackslash{n}
\texttt{mov cl, byte [esi]}
& \textit{Perform the xor between BL register and 0xBB and jump to the label formatting if the result is zero else move the current byte of the shellcode in the CL register.}\\ \midrule
\texttt{xor ecx, ecx} \textbackslash{n} \texttt{mul ecx} &
\textit{Zero out the EAX and ECX registers.
}\\ \bottomrule
\end{tabular}
\end{table}

To take into account the variability of descriptions in NL, multiple authors described independently different samples of the dataset in the English language. Where available, we used as NL descriptions the comments written by developers of the collected programs.
The dataset’s NL descriptions were independently crafted by three authors, all with a computer science background and expertise in assembly language and cybersecurity, including 2 Ph.D. students and a post-doc researcher.
Each author initially worked independently on a subset of the dataset to create their respective descriptions. Then, to decide on the final version of the NL descriptions, all authors collaboratively reviewed all the samples included in the extended version of the dataset. The code descriptions were thoroughly cross-verified to ensure consistency, accuracy, and completeness. Any differences were resolved through consensus, ensuring that the final descriptions accurately represented the intent and functionality of the code.

The dataset also includes $1,374$ intents (${\sim}23\%$ of the dataset) that generate multiple lines of assembly code, separated by the newline character \textit{\textbackslash{n}}. These multi-line snippets contain many different assembly instructions (e.g., whole functions).
Table~\ref{tab:dataset_statistics} summarizes the statistics of the dataset used in this work, including the unique examples of NL intents and assembly code snippets, the unique number of tokens, and the average number of tokens per snippet and intent.

\begin{table}[t]
\centering
\caption{Dataset statistics}
\label{tab:dataset_statistics}
\footnotesize
\begin{tabular}{
>{\centering\arraybackslash}m{3cm} |
>{\centering\arraybackslash}m{2cm}
>{\centering\arraybackslash}m{2.5cm}}
\toprule
\textbf{Metric} & \textbf{NL Intents} & \textbf{Assembly Snippets}\\ \toprule
\textit{Unique lines}    & $5,740$   & $3,316$\\
\textit{Unique tokens}      & $2,855$   & $1,770$\\
\textit{Avg. tokens per line}        & $9.18$    & $5.83$\\
\bottomrule
\end{tabular}
\end{table}

\subsection{Implementation of Perturbations}
\label{subsec:implementation}
\noindent
$\blacksquare$ \textbf{Word Substitution.} We used TextAttack~\cite{morris2020textattack}, a Python framework for data augmentation, to substitute words and apply constraints, and Flair POS-tagging model~\cite{akbik2018coling} as POS tagger. The TextAttack framework implements the \textit{word swap by embedding} transformation replacing words with their candidates in the counter-fitted vector space~\cite{mrksic:2016:naacl}.
To build the vocabulary of programming--language-related words, we compared our dataset with a book containing $832$ pages describing the English language as it is spoken around the world~\cite{kachru2009handbook}.
Unsurprisingly, structure-related words like \textit{stack}, \textit{function}, or \textit{pointer} are among the most frequent words in our dataset but barely (or never) mentioned in the book, while the name-related words, such as the name of the registers (e.g., \texttt{EAX}) and labels (e.g., \texttt{encoded\_shellcode}) never appear in the book.
We randomly substitute the 10\% of words in every NL intent, ensuring that at least one word is swapped with a similar one~\cite{yoo-qi-2021-towards-improving}.

\noindent
$\blacksquare$ \textbf{Word Omission.} We removed action-related words from each intent (e.g., \textit{define}, \textit{add}, etc.) using Flair POS-tagging model~\cite{akbik2018coling} as POS tagger, while we identify structure-related words (e.g., \textit{register}, \textit{label}, etc.) and name-related words (e.g., \texttt{EAX}, \texttt{\_start\_label}, etc.) to omit from the NL intents by using the vocabulary of programming--language-related words. 
After identifying the words to omit, we removed from the intents all the words related to a word omission category (i.e., we removed all the verbs, all the structure-related words, or all the names), generating different perturbed intents, according to the word omission categories, but we did not generate a perturbed intent with multiple word omission categories.

The disparity between the percentage of substituted words (10\%) and the omission of entire word categories arises from the different nature and goals of these perturbations. Substitution is aimed at introducing variability without altering the overall semantic meaning too drastically, while omission is aimed at removing critical components to test the model's robustness to significant information loss (e.g. when developers omit information in the NL descriptions).
Previous work showed that a value of 10\% represents the best compromise between performance improvement and maintaining the integrity of the original sentences~\cite{wei2019eda}. As reported in their study, higher values tend to hurt performance, likely because replacing too many words in a sentence changes its original identity. 
Following best practices in the SOTA~\cite{wei2019eda,yoo-qi-2021-towards-improving}, we randomly substituted 10\% of words in each NL intent, ensuring that at least one word is swapped with a similar one. Substituting a higher percentage of words risks significantly altering the original intent's meaning, potentially making the perturbed intent unrecognizable or too different from the original. Hence, by substituting 10\% of the words, we ensure that the perturbations introduce sufficient variability to test robustness in the subsequent analyses while preserving the overall semantic equivalence of the intent.

In contrast, our approach to word omissions involved removing all words related to specific categories (verbs, structure-related words, or names), which are limited and specific. By removing 100\% of the words in these categories (one category per perturbation), we obtain a targeted subset of omitted words that significantly impacts the intent's structure or content, yet this subset remains a limited portion of the total words. In fact, we computed, for each word-omission category, the average percentage of omission with respect to the total length of the NL intents in the dataset used for our experiments. We found that the average percentage of words that can be omitted is approximately 14\% for action-related words, 13\% for structure-related words, and 14\% for name-related words. These values, although higher than 10\%, are still in line with the percentage used for word substitutions.
In summary, while the ratio of substituted words is slightly lower compared to the ratio of omitted words, our methodology ensures that we are testing different aspects of the model's robustness.  

\subsection{Metrics}
To evaluate the performance of the models, we computed the \textit{Syntactic Accuracy} (SYN) and the \textit{Semantic Accuracy} (SEM).
While the former gives insights into whether the code is correct according to the rules of the target language, the latter indicates whether the output is the exact translation of the NL intent into the target programming language (i.e., assembly).
Syntactic/semantic accuracy is computed as the number of syntactically/semantically correct predictions over the total number of predictions (i.e., the total number of code snippets generated by the model).
Furthermore, to better estimate the impact of the perturbations on the models, we computed the \textit{Robust Accuracy} (ROB)~\cite{huang2021robustness}. Similar to semantic accuracy, this metric evaluates the semantic correctness of the code predicted by the models. However, instead of considering all the examples in the test data, the robust accuracy limits the evaluation to the subset of examples correctly generated by the model without the use of perturbations (i.e., it discards from the evaluation the semantically incorrect predictions w/o perturbations in the NL intents). 
This means that, to compute ROB, we consider only the subset of predictions correctly generated in the baseline setting (i.e., no perturbations in the test set), and compute the percentage of predictions that are correct even after the perturbations in the test set. Hence, the formula for computation is:
\[
\text{ROB} = \frac{N_{\text{SEM}=1 \text{ before and after perturbation}}}{N_{\text{SEM}=1 \text{ before perturbation}}} \times 100
\]
where:
\begin{itemize}
    \item $N_{\text{SEM}=1 \text{ before and after perturbation}}$ represents the number of samples for which the semantic accuracy (SEM) equals 1 both before and after the perturbation.
    \item $N_{\text{SEM}=1 \text{ before perturbation}}$ represents the number of samples for which the semantic accuracy measure (SEM) equals 1 before the perturbation.
\end{itemize}

To evaluate the syntactic correctness of the predictions, we use the NASM assembler for the Intel x86 architecture in order to check whether the code generated by the models is compilable.
For the semantic correctness of the predictions, instead, we manually analyze every code snippet generated by the models to inspect if it is the correct translation of the English intent. This analysis cannot be performed automatically (e.g., by comparing the predictions with ground-truth references) since an English intent can be translated into different but equivalent code snippets. Thereby, manual (human) evaluation is a common practice in machine translation~\cite{han-etal-2021-translation,liguori2023evaluates}. To reduce the possibility of errors in manual analysis, multiple authors performed this evaluation independently, obtaining a consensus for the semantic correctness of the predictions.

More precisely, the manual evaluation was conducted by a diverse group consisting of 3 human evaluators, all with a computer science background and expertise in assembly language and cybersecurity. The group included individuals with varying degrees of professional experience and educational qualifications. In particular, 2 Ph.D. students with a master’s degree and a researcher with a Ph.D. in information technologies. The diversity and expertise of our evaluators ensured the reliability of our human evaluation process. Each model’s prediction was independently evaluated by all 3 evaluators to ensure thoroughness, checking all predictions in each round of semantic correctness evaluation.  
Specifically, the manual assessment of the semantic correctness of a single code snippet took $\sim$10 seconds on average. We performed a total of 21 experiments for each of 3 models and evaluated them on a test set of 590 samples, resulting in a total of $\sim 100$ of human hours.
In the few cases of disagreement, a structured discussion was held to reach a consensus, and the group collectively decided on the semantic correctness of each prediction. This consensus-driven approach ensured that all differing viewpoints were considered and resolved.

\section{Semantics Evaluation}
\label{sec:semantics}
A model is robust if the performance obtained on the original, non-perturbed data is similar to the one achieved against the perturbed data, i.e., the perturbations do not imply a significant drop in the model's performance.
To properly assess the model's robustness, the key requirement is to guarantee that new, perturbed inputs, although syntactically different, preserve the semantics of the original ones.
In fact, if the perturbations alter the semantics of the NL descriptions, then the model is tested with different inputs (e.g., inputs requiring other actions), hence invalidating the analysis. 

Therefore, before analyzing the robustness of the models, we assessed whether the perturbations violate this requirement by comparing the semantics of inputs before and after the perturbations.
To this aim, we perturbed all the $5,900$ examples in our dataset with word substitution and word omission. For the former, we considered both the substitution with and without constraints to verify if they helped to preserve the original semantics. For the latter, we performed an in-depth analysis that considered independently all the types of omission, i.e., action, structure, and name-related words. This helped us to understand what is the omission that impacted the most on the semantics.

Unfortunately, there is no automatic solution to check the semantic equivalence of the descriptions, i.e., solving the ambiguity in the NL. A common practice is represented by the inspection from different human annotators (e.g., through a survey) that manually analyze whether the perturbed descriptions are actually semantically equivalent to the original ones~\cite{DBLP:conf/icse/MastropaoloPGCSOB23}.
However, a manual inspection becomes infeasible and too prone to errors due to the massive amount of NL descriptions to review, which, in our case study, depends on the number of examples in the dataset (i.e., $5,900$), and the number of different perturbations (i.e., word substitution with and without constraints, action, structure, and name-related omission). 

Therefore, following the best practices of the field~\cite{li2018textbugger,huang2021robustness,morris2020reevaluating}, we adopted multi-lingual models to compute sentence embeddings that are then compared with cosine similarity to find sentences with similar semantics.
More precisely, these models, named \textit{sentence transformers}, are fine-tuned to produce similar vector-space representations for semantically similar sentences within one language or across languages~\cite{reimers2019sentence,DBLP:conf/emnlp/ReimersG20}.
We adopted the sentence transformers to create sentence embeddings of both the original, non-perturbed NL descriptions and the perturbed ones. Then, we compared the cosine similarity between the two embedded descriptions: only if the similarity is higher than a threshold, then we consider that the perturbation did not alter the semantics of the original description. 
We implemented these models by using SentenceTransformers, a Python framework for state-of-the-art sentence, text, and image embeddings~\cite{st}. We set the threshold for similarity equal to $0.80$~\cite{li2018textbugger}.

\begin{figure}[t]
  \centering
  \includegraphics[width=1\columnwidth]{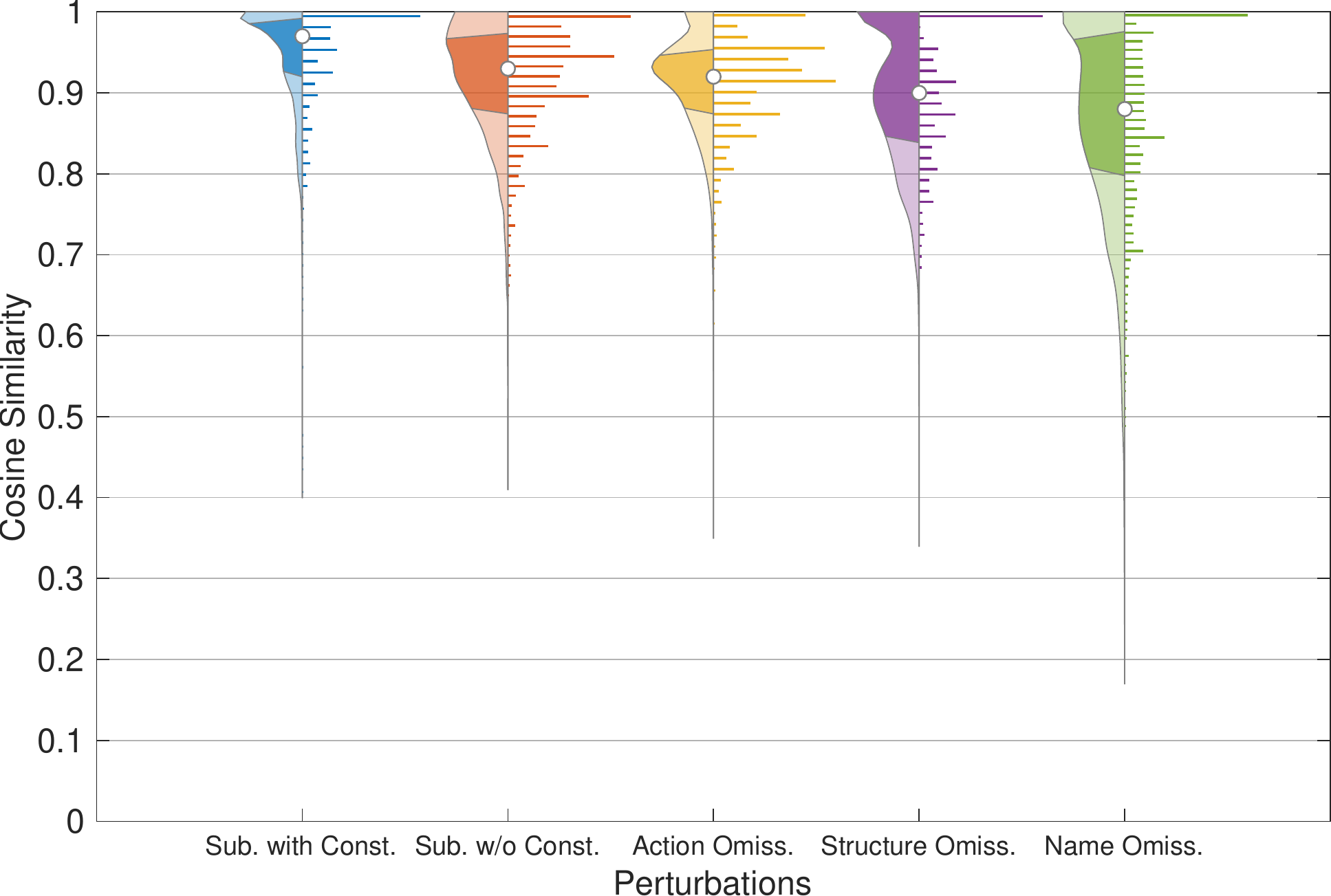}
  \caption{Violin plots showing the cosine similarity between the original and the perturbed NL descriptions in the whole dataset.}
  \label{fig:ST}
\end{figure}

\figurename{}~\ref{fig:ST} shows, for each perturbation, the violin plots of the cosine similarity across all the examples in the dataset. 
First, the figure highlights that, regardless of the perturbations, the cosine similarity values are mostly distributed over the threshold of $0.80$. 
For the word substitution, the median value is $0.97$ and $0.93$ with and without constraints, respectively (the average is $0.94$ vs. $0.91$). The percentage of perturbed descriptions having a cosine similarity higher than the threshold value is equal to $96.20\%$ when we use constraints and $92.90\%$ without constraints.
For the word omission, the median values are $0.92$, $0.90$, and $0.88$ for action-related, structure-related, and name-related words, respectively (the average values are $0.91$, $0.90$, and $0.86$). In this case, the percentage of perturbations not violating the threshold value for the semantic similarity is $94.98\%$ for action-related, $85.97\%$ for structure-related and $74.63\%$ for name-related words.

These results provide practical insights. Indeed, the word substitution perturbation impacts less than the word omission on the semantic similarity with the original descriptions. This happens because the removal of a word, different from its substitution, implies that the information contained in the omitted word is missing from the original sentence. 
A further takeaway of the analysis is that the constraints for word substitution help to preserve the original semantics of the descriptions. In fact, the cosine similarity is on average higher than $3\%$ when compared with the substitution without constraints.
Finally, among the word omission categories, the removal of action-related words is the most semantics-preserving, followed by structure-related and name-related words. The latter is the only perturbation that has a median (average) value below $0.90$, although it is still higher than the threshold value. This category of words contains particular important information (e.g., the name of the variables, registers, etc.), but which are sometimes omitted either because they are redundant (e.g., repeated in the same NL description or in previous ones), or because programmers are not accurate when describing the code.

We also performed a sensitivity analysis of the threshold for cosine similarity, which revealed that if we use a more tolerant threshold of $0.70$, the percentage of perturbed descriptions having a cosine similarity higher than the threshold value is close to 100\% for word substitution, and ~96\% for the three word-omission types, on average. By using a very strict threshold of $0.90$, instead, the percentage of descriptions meeting this requirement is ~80\% for word substitution and ~55\% for word omission types, on average. Following previous studies in the field~\cite{li2018textbugger}, we set the threshold to $0.80$, which allows us to test the models with perturbed descriptions that are not too similar to the original ones, yet preserve their semantics.

For the robustness analysis (see \S{}~\ref{sec:results}), we train and test the models with perturbed intents that meet the similarity threshold, i.e., when the cosine similarity between the encoded code description before and after the perturbation is greater than $0.80$.

Finally, we performed a manual verification to confirm the perturbed NL intents maintain their semantic equivalence to the original ones. To this aim, we conducted a comprehensive survey involving domain experts to evaluate the semantic equivalence of the perturbed descriptions.
Indeed, given the large dataset ($5,900$ samples) and the $5$ perturbation categories applied, a full manual review was impractical. Therefore, we designed a survey to assess the semantic accuracy of the perturbations. We engaged $31$ people with a strong background in software security and shellcode generation. The participants included $10$ Ph.D. students, $6$ postdoctoral researchers and $3$ professors from our research group, $5$ engineering MSc students from a software security course, and $7$ students from a national Italian cybersecurity training course. Each participant was initially tasked with evaluating 50 perturbed descriptions, randomly extracted from our dataset (random sampling with replacement), including 10 random samples for each of the five categories: word substitution with constraints, word substitution without constraints, action-related word omission, structure-related word omission, and name-related word omission. Overall, we collected $1,104$ evaluated descriptions (not all participants completed the full evaluation of 50 descriptions, based on their availability and confidence).

\begin{figure}[t]
    \centering
    \subfloat[Cosine similarity higher than the threshold.]{\label{fig:human_above}
         \includegraphics[width=1\textwidth]{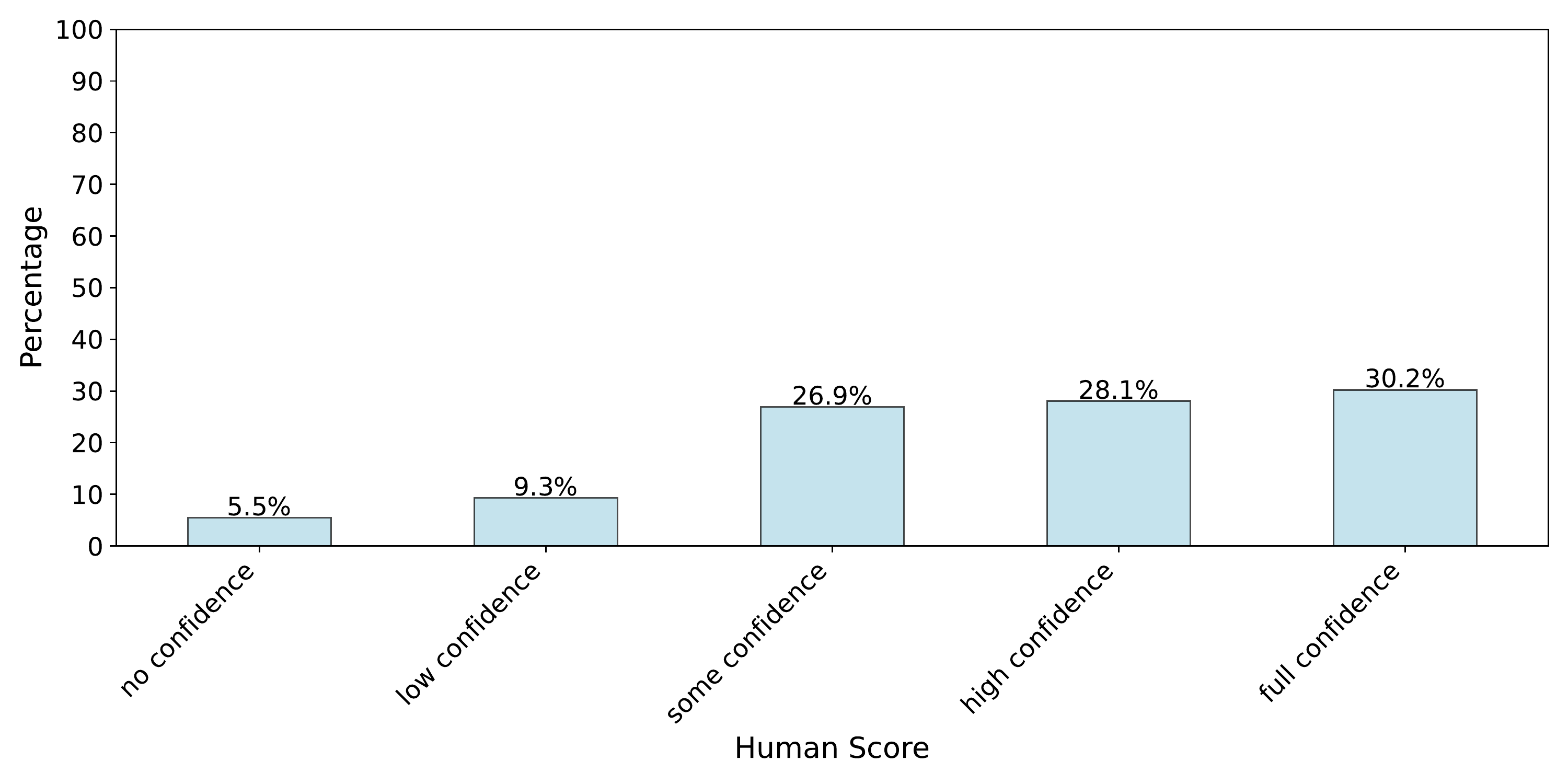}}
         
   \subfloat[Cosine similarity lower than the threshold.]{\label{fig:human_below}
         \includegraphics[width=1\textwidth]{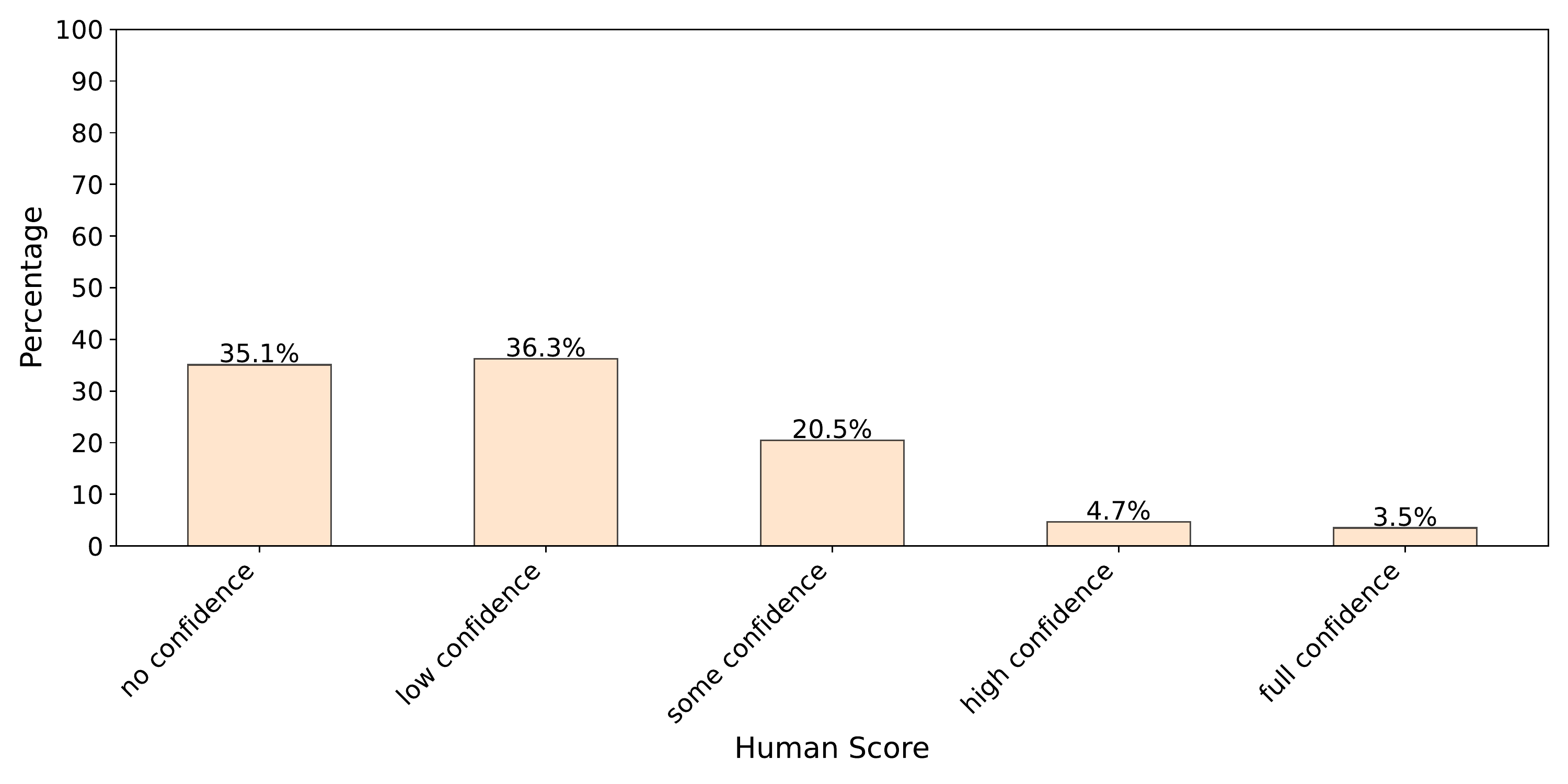}}
\caption{Human scores of perturbations with cosine similarity higher or lower than the threshold.}
\label{fig:survey}
\end{figure}

In the survey, participants were provided with three columns of information: ``Code" containing assembly code extracted from real shellcode programs, ``Original Description" containing the original (i.e., the non-perturbed) NL description of the code, and ``Perturbed Description" containing the original description perturbed with either word substitution or word omission. The participants were asked to rate their confidence that the perturbed description still accurately reflected the code. They used a scale of 1 to 5, where 5 indicated full confidence that the perturbed description was accurate and complete, 4 indicated high confidence with only minor discrepancies, 3 indicated some confidence, 2 indicated low confidence with evident inaccuracies, and 1 indicated no confidence, meaning the perturbed description poorly reflected the code and was notably different from the original description.
\figurename{}~\ref{fig:survey} shows the results of the survey. The results demonstrated a clear pattern in the human evaluation scores based on the cosine similarity threshold of $0.80$. For perturbations with a model score greater than the threshold (\figurename{}~\ref{fig:human_above}), the majority of the human scores were 3, 4, or 5, indicating a high level of confidence in the semantic accuracy of these perturbations. Specifically, approximately $85\%$ of these perturbations received a human score of 3 or higher. On the other hand, for perturbations with a model score less than or equal to the threshold (\figurename{}~\ref{fig:human_below}), the human scores were predominantly 1 and 2 (around $71\%$), reflecting a low level of confidence in their semantic accuracy. These findings validate our approach, showing that human evaluators perceive the perturbations we use to test the models (those with a model score greater than the threshold) as semantically accurate. In contrast, those who do not meet the threshold generally receive low human scores, justifying their exclusion from the robustness analysis.
Overall, the results show that the automatic method performs well even though a low percentage of non-semantic preserving perturbations are used for training and testing models. We achieved a very good trade-off,  as it allows us to conduct large-scale automated evaluations, which would be impossible with human evaluation alone. This validates our methodology and the appropriateness of the constraints applied in the word substitution and omission approach.

\section{Robustness Evaluation}
\label{sec:results}

To perform the experiments, we split the dataset into training (the set of examples used to fit the parameters), validation (the set used to tune the hyper-parameters of the models), and test (the set used for the evaluation of the models) sets using a common random $80\%/10\%/10\%$ ratio~\cite{kim2018artificial,DBLP:conf/msr/MashhadiH21}. 

We acknowledge that data splitting may influence the results. However, the huge amount of our experimentation and the subsequent manual analysis to assess the semantics correctness of the code for all perturbation types would make it infeasible to consider the influence of multiple data splits on the results. Therefore, we used random 80\%-10\%-10\% training-dev-test sets splits, following the best practices in the SOTA. Moreover, we believe that the size of the dataset ($5,900$ samples) allowed us both to have a great variability in the training set via data augmentation and perform a comprehensive robustness evaluation of the models on a wide set of different inputs, hence reducing the impact of the data splitting on our results.

To validate the proposed method, we conducted the experimental analysis by answering the following research questions (RQs):

\vspace{0.1cm}
\noindent
$\rhd$ \textbf{RQ1:} \textit{Are models robust to perturbations in the code descriptions?}\\
We first assess the robustness of the models in the generation of offensive assembly code by using word substitution and word omission in the NL intents of the test set. In this RQ, the test set is 100\% perturbed, i.e., we add a perturbation in every code description.
This analysis aims to comprehensively estimate the model’s robustness in a scenario of high linguistic variation (100\% perturbed test set) with respect to its performance when tested on the original non-perturbed descriptions, which have a similar writing style to what the model has seen during training.

\vspace{0.1cm}
\noindent
$\rhd$ \textbf{RQ2:} \textit{Can data augmentation enhance the robustness of the models against perturbed code descriptions?}\\
In this RQ, we assess if training the model with perturbed NL intents improves the performance of the models against perturbed code descriptions. We use a different percentage of perturbed intents in the training set (25\%, 50\%, and 100\%) and compare the results with the case of no perturbed intents in the training set (0\%). The test set is again perturbed 100\%. 

\vspace{0.1cm}
\noindent
$\rhd$ \textbf{RQ3:} \textit{Can data augmentation increase the performance of models against non-perturbed code descriptions?}\\
Finally, we assess whether data augmentation can improve the performance of the models against non-perturbed code descriptions, i.e., the original test set.
The choice to evaluate the model’s performance on a non-perturbed test set (0\%) when trained on partially augmented data (50\%) is meant to assess if the introduction of variability in the training data increases the performance of the model’s performance on the original, non-perturbed test set. This split provides a good trade-off between robustness on varied data and performance on the original data. Consequently, we selected this perturbation percentage for further experimentation to ensure balanced results.

\subsection{Robustness Evaluation On Perturbed Code Descriptions}
\label{subsec:robustness}

To assess the robustness of the models in the generation of offensive code, we perturbed the test set by using word substitution or by removing words from the NL intents of the corpus. Then we compared the results with the performance of the models on the original, non-perturbed test set. Ideally, a model is robust if the results obtained with and without perturbations are similar.

As a primary goal, we separately evaluated the impact of each perturbation type to understand their individual effects on both model robustness and performance. Applying both perturbations at once may be too aggressive and compromise the semantic equivalence of the original code descriptions, potentially leading to unrealistic inputs that are not representative of real-world scenarios.
By assessing each perturbation type separately, we aimed to determine how augmenting the original data with each perturbation (i.e., word substitution and word omission) independently affects the models' robustness. This approach allowed us to pinpoint specific strengths and limitations of the models and our perturbation-based methodology.

\tablename{}~\ref{tab:perturbed_test} shows the results of the models in terms of syntactic, semantic, and robust accuracy, with and without perturbations in the test set. 
Without any perturbation, the performance of all models is $\geq 90\%$ for the syntactic accuracy, while the semantic one is equal to $65\%$, $69\%$, and $69\%$ for Seq2Seq, CodeBERT and CodeT5+, respectively.
Hence, the choice of models does not impact the ability to produce syntactically correct code. 

\begin{table}[t]
\centering
\caption{Evaluation of the models with and w/o perturbations in the test set. \textit{None} indicates the original, non-perturbed test set.}
\label{tab:perturbed_test}
\scriptsize
\begin{tabular}{ 
>{\centering\arraybackslash}m{1.75cm}  |
>{\arraybackslash}m{0.75cm} 
>{\arraybackslash}m{0.75cm}
>{\arraybackslash}m{0.75cm}|
>{\arraybackslash}m{0.75cm}
>{\arraybackslash}m{0.75cm}
>{\arraybackslash}m{0.75cm}|
>{\arraybackslash}m{0.75cm}
>{\arraybackslash}m{0.75cm}
>{\arraybackslash}m{0.75cm}}
\toprule
& \multicolumn{3}{c}{\textbf{Seq2Seq}} & \multicolumn{3}{c}{\textbf{CodeBERT}} & \multicolumn{3}{c}{\textbf{CodeT5+}}\\
\textbf{Perturb.}  & \textbf{SYN} & \textbf{SEM} & \textbf{ROB} & \textbf{SYN} & \textbf{SEM} & \textbf{ROB} & \textbf{SYN} & \textbf{SEM} & \textbf{ROB}\\ \midrule
\textit{None} & \Chart{0.95} & \Chart{0.65} & - & \Chart{0.93} & \Chart{0.69} & - & \Chart{0.90} & \Chart{0.69} & -\\ 
\textit{Word Substit.} & \Chart{0.86} & \Chart{0.51} & \Chart{0.66} & \Chart{0.89} & \Chart{0.49} & \Chart{0.68} & \Chart{0.73} & \Chart{0.42} & \Chart{0.58}\\ 
\textit{Word Omiss.} & \Chart{0.81} & \Chart{0.33} & \Chart{0.45} & \Chart{0.67} & \Chart{0.32} & \Chart{0.44} & \Chart{0.75} & \Chart{0.37} & \Chart{0.51}\\ \bottomrule
\end{tabular}
\end{table}

When we evaluate the models using word substitution in the NL intents of the test set, the syntactic correctness of the models drops by $9\%$, $4\%$, and $17\%$ for Seq2Seq, CodeBERT, and CodeT5+, respectively, while the semantic correctness of the predictions decreases by $14\%$ for Seq2Seq, by $20\%$ for CodeBERT, and by $27\%$ for CodeT5+. 
According to the robust accuracy, only the $66\%$, $68\%$, and $58\%$ of the code snippets are correctly predicted before and after perturbations by SeqSeq, CodeBERT, and CodeT5+, respectively.

The performance gets even worse when we omit words from the NL intents of the test set. Indeed, the syntactic correctness notably fell by $14\%$, $26\%$, and $15\%$ for Seq2Seq, CodeBERT, and CodeT5+, respectively. Also, the table highlights that semantic correctness decreases by $32\%$ for CodeT5+, and is approximately halved for the other two models when compared to the results obtained without perturbations. On top of that, only $\leq 51\%$ of the models' predictions are correct both with and without perturbations.

This huge decrease in performance can be attributed to the fact that there are instances where input tokens appear directly in the output. This is particularly common in shellcode generation, where intents can be highly detailed since different registers have different, specific functionalities that the models are not able to comprehend if not correctly steered by the code descriptions (e.g., the \texttt{ESI} register is typically used to store the encoded shellcode, \texttt{ECX} is usually used to store the loop counter, etc.). When this crucial information is missing from the intent (as described in \S{}~\ref{subsec:word_omission}, the model is generally not capable of deriving it from the surrounding context and fails to predict the correct translation.

\begin{figure}[ht]
    \centering
    \subfloat[]{\label{fig:seq2seq_no_perturbation_60}
             \includegraphics[width=0.45\textwidth]{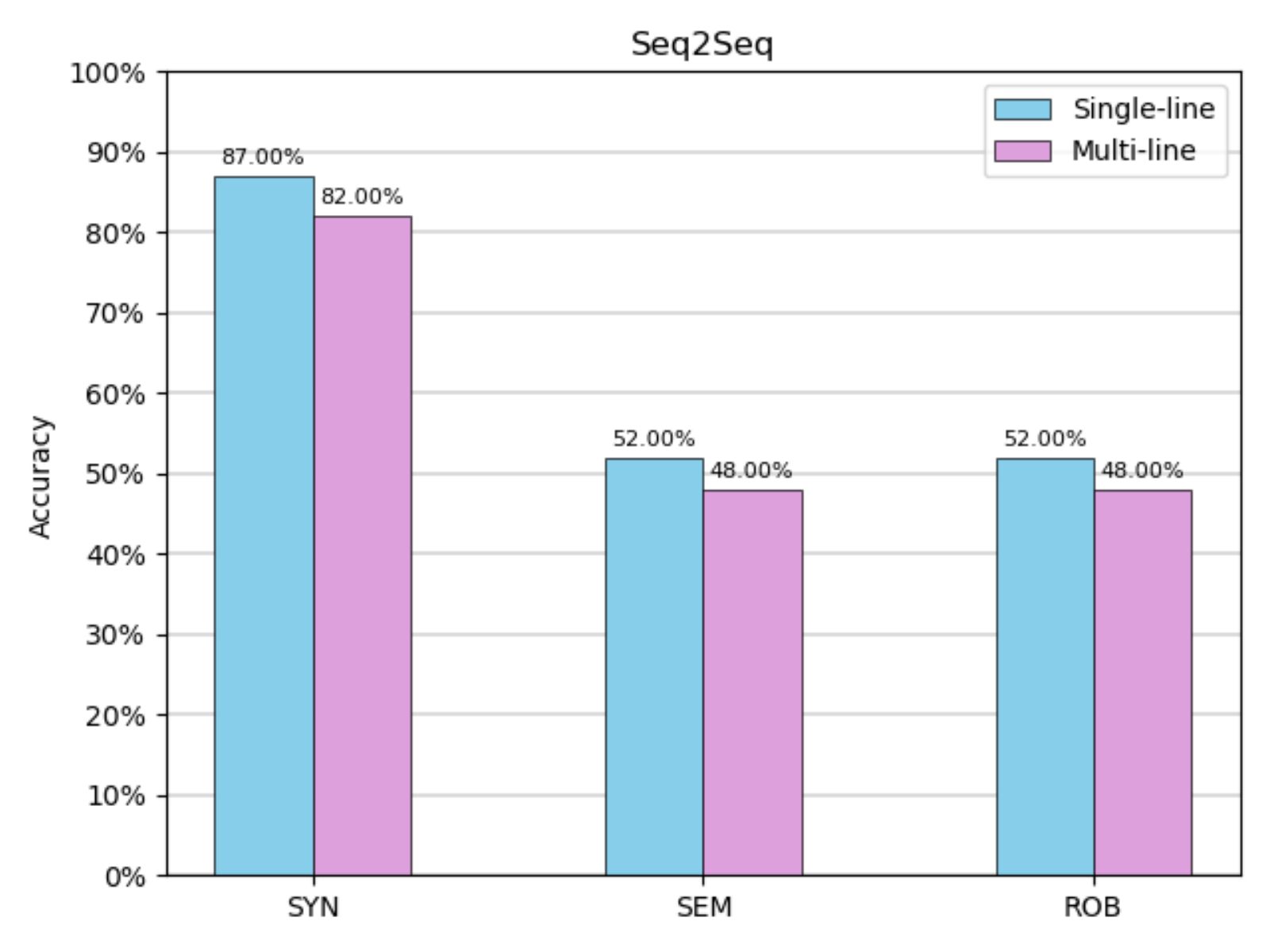}}
    \subfloat[]{\label{fig:codebert_no_perturbation_60}
             \includegraphics[width=0.45\textwidth]{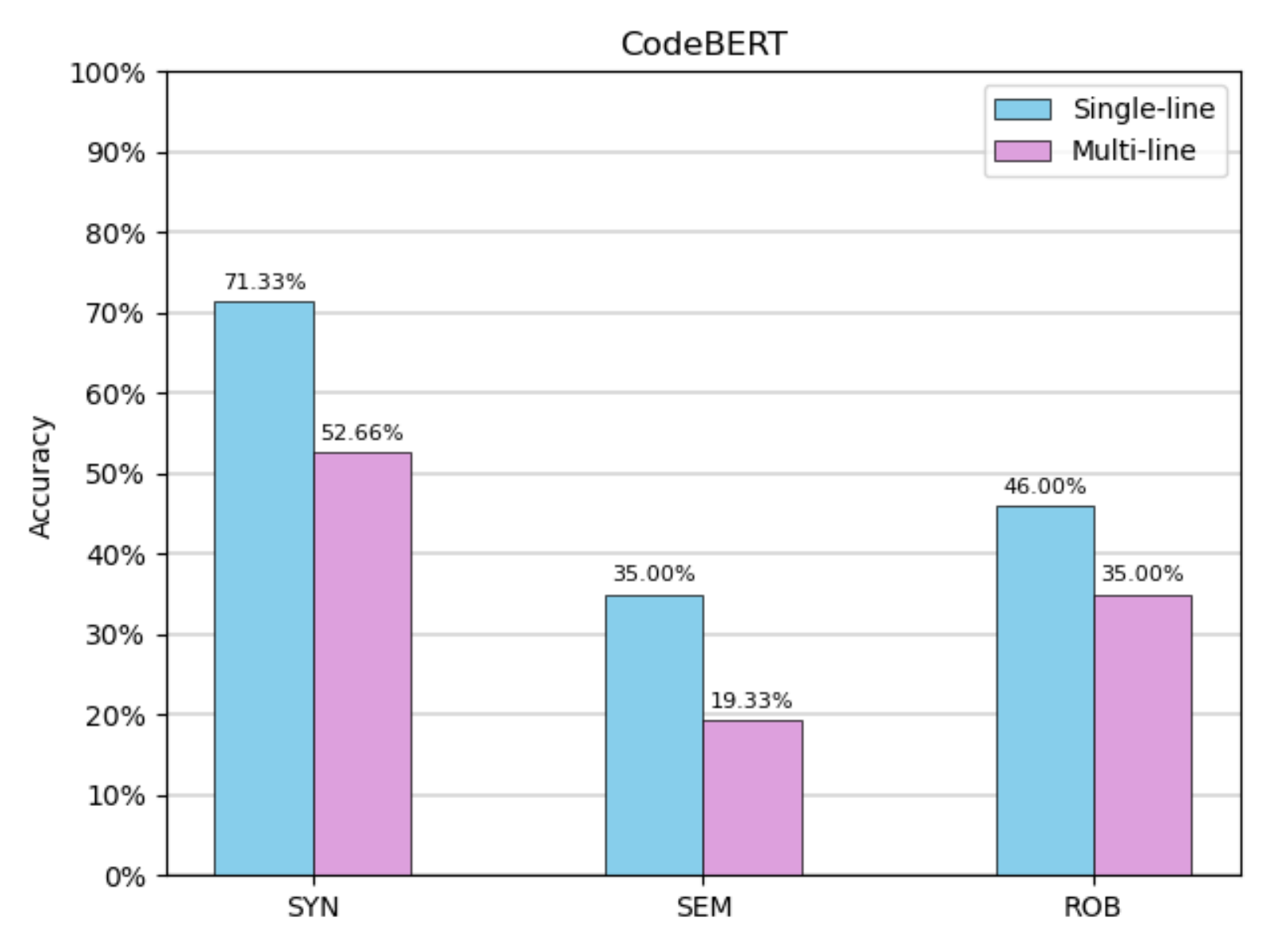}} \\
    \subfloat[]{\label{fig:codet5_no_perturbation_60}
             \includegraphics[width=0.45\textwidth]{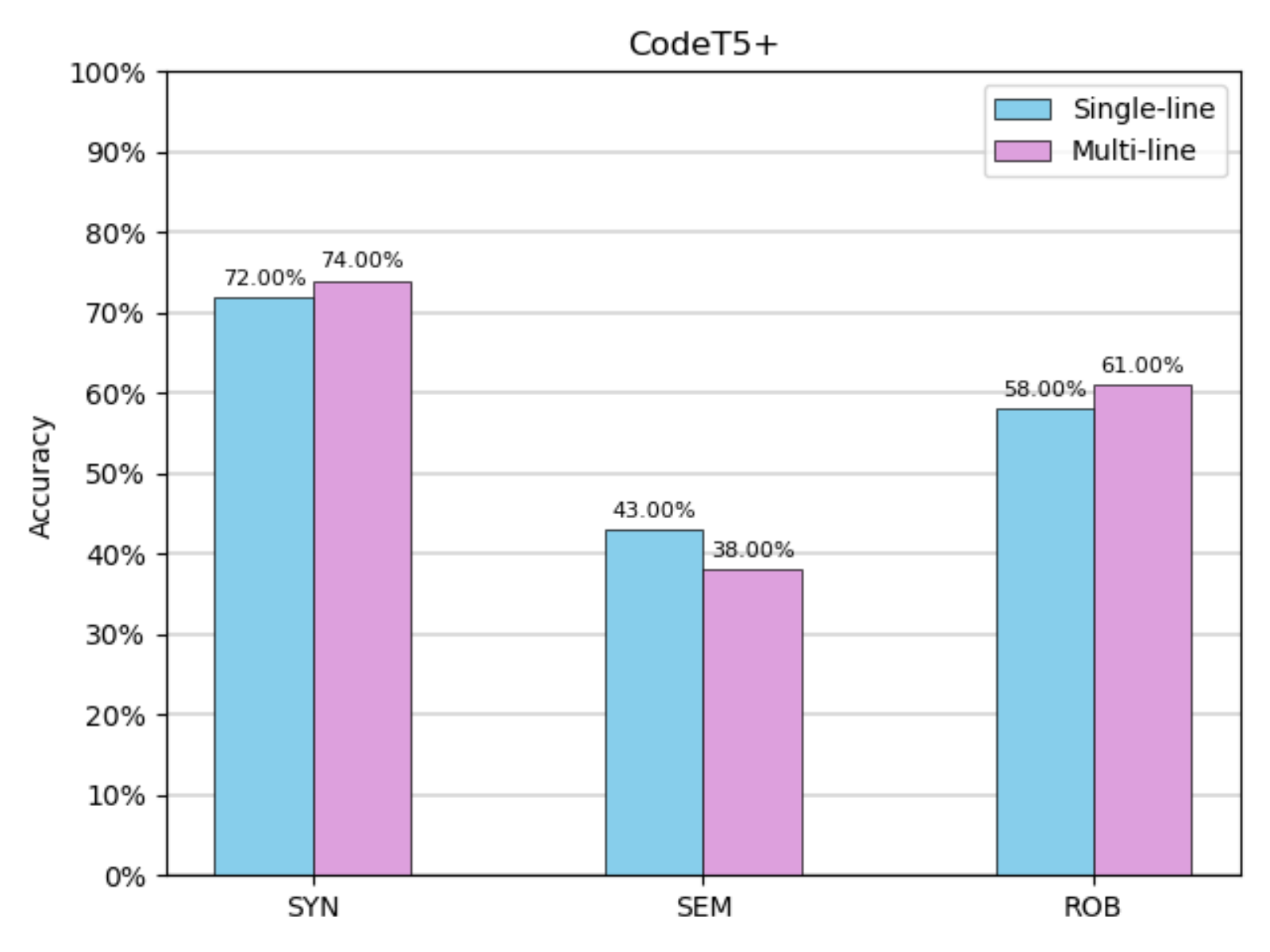}}
    \caption{Comparison between the models' performance on single-line code snippets \textit{vs.} multi-line code snippets in terms of syntactic (SYN) and semantic (SEM) accuracy.}
    \label{fig:single_multi_no_pertub}
\end{figure}

\begin{figure}[ht]
    \centering
        \subfloat[Word Substitution]{\label{fig:seq2seq_word_substitution_61}
             \includegraphics[width=0.45\textwidth]{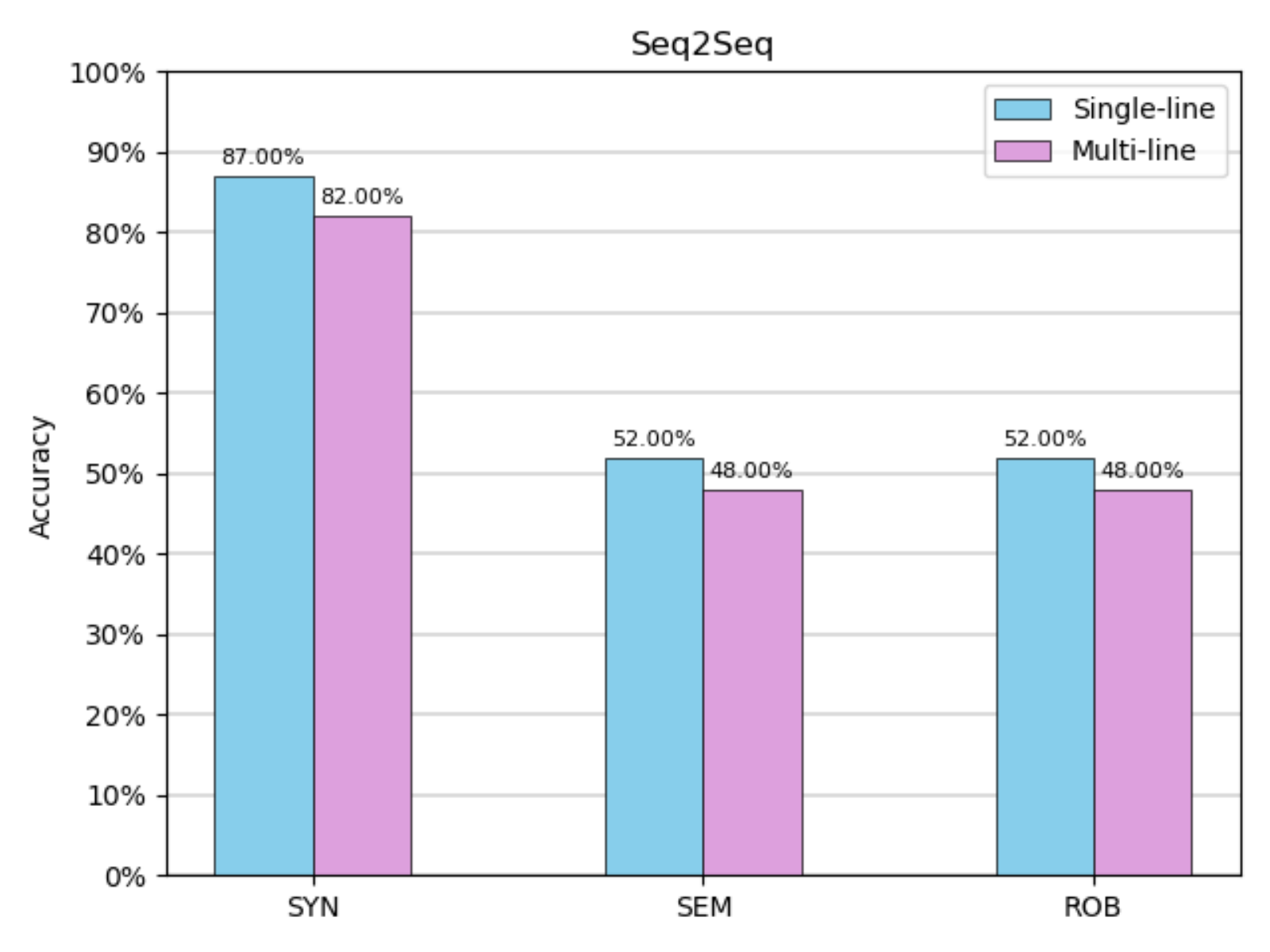}}
        \subfloat[Word Omission]{\label{fig:seq2seq_word_omission_61}
             \includegraphics[width=0.45\textwidth]{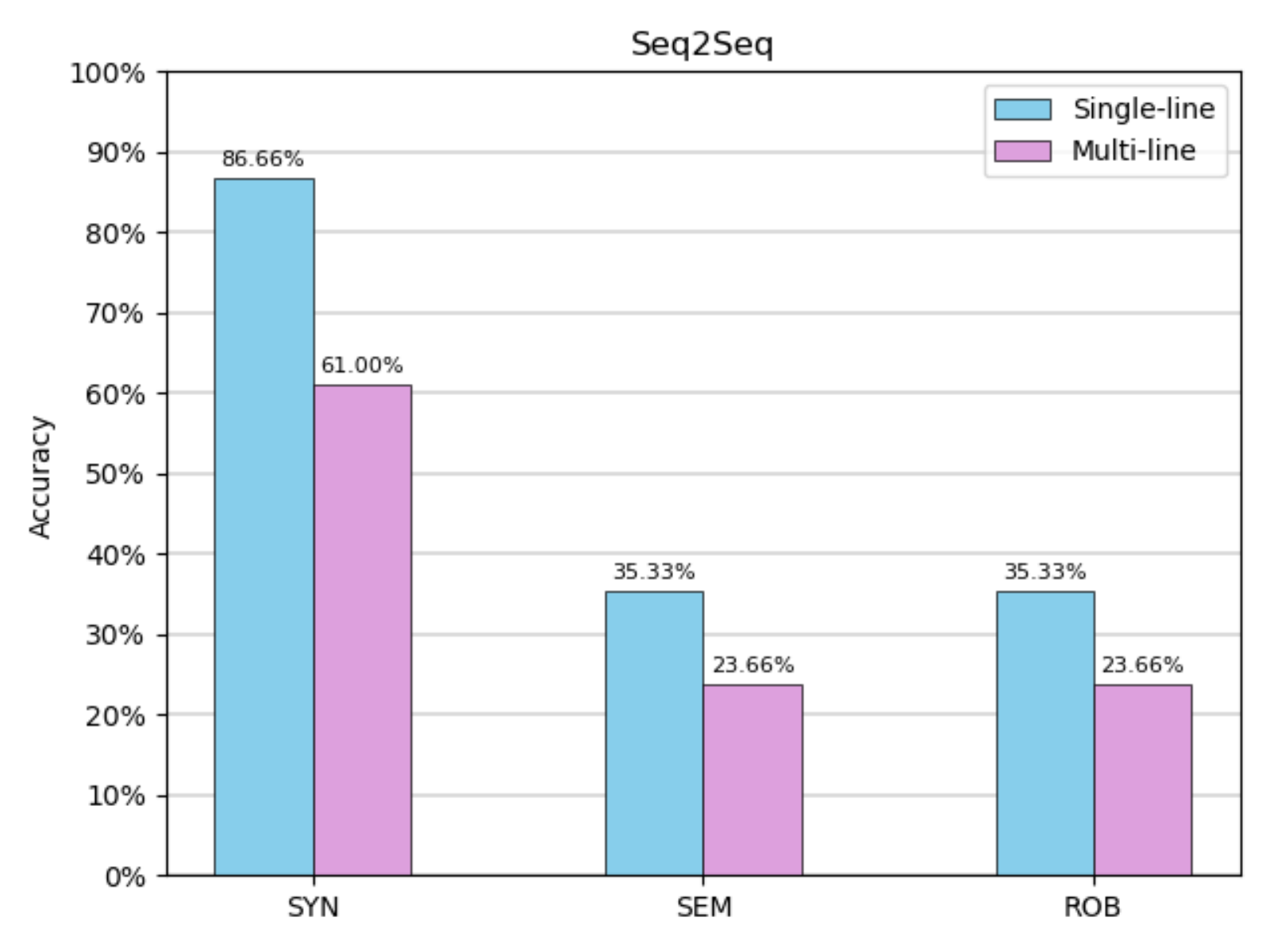}} \\
        
        \subfloat[Word Substitution]{\label{fig:codebert_word_substitution_61}
             \includegraphics[width=0.45\textwidth]{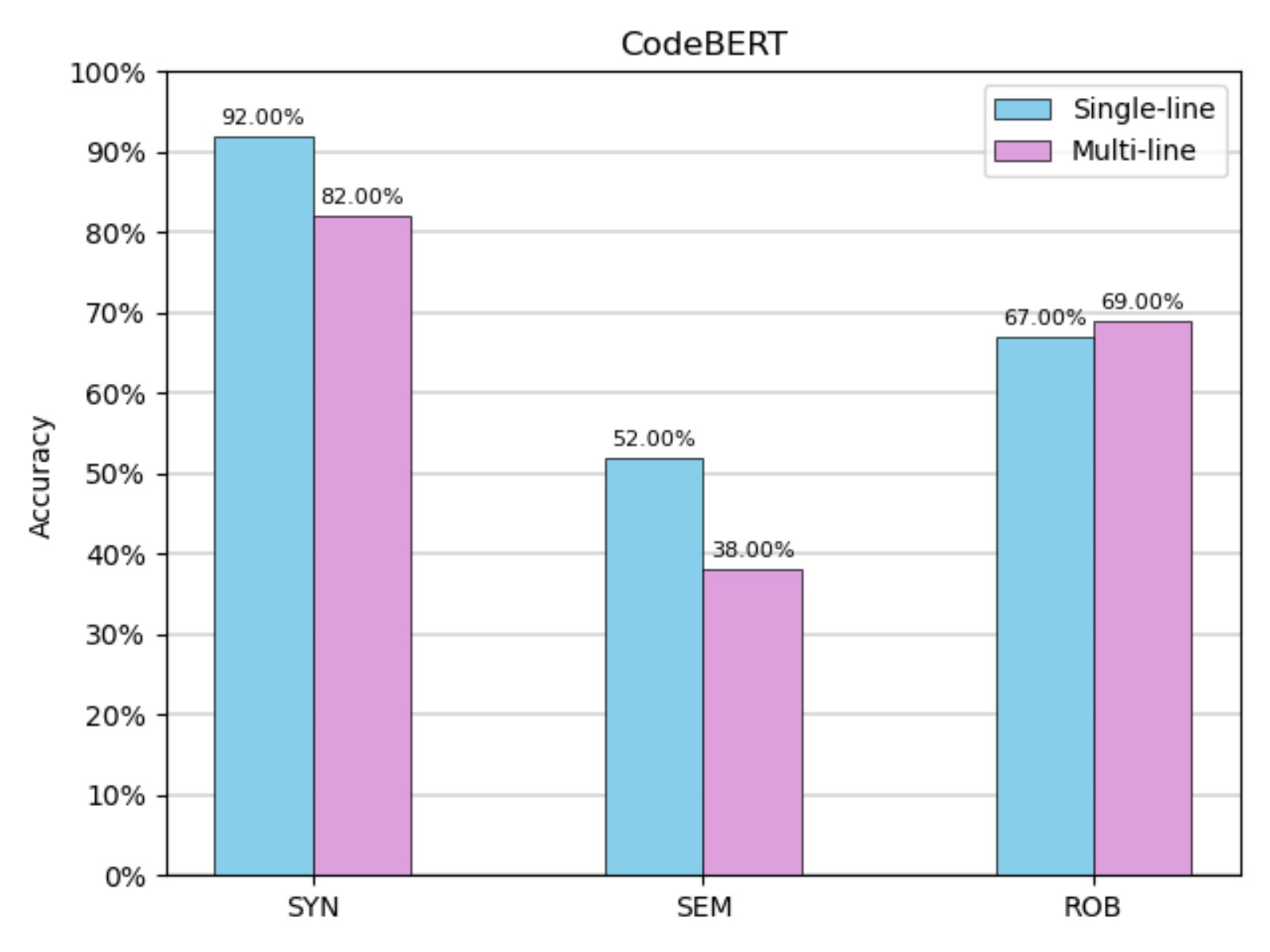}}
        \subfloat[Word Omission]{\label{fig:codebert_word_omission_61}
             \includegraphics[width=0.45\textwidth]{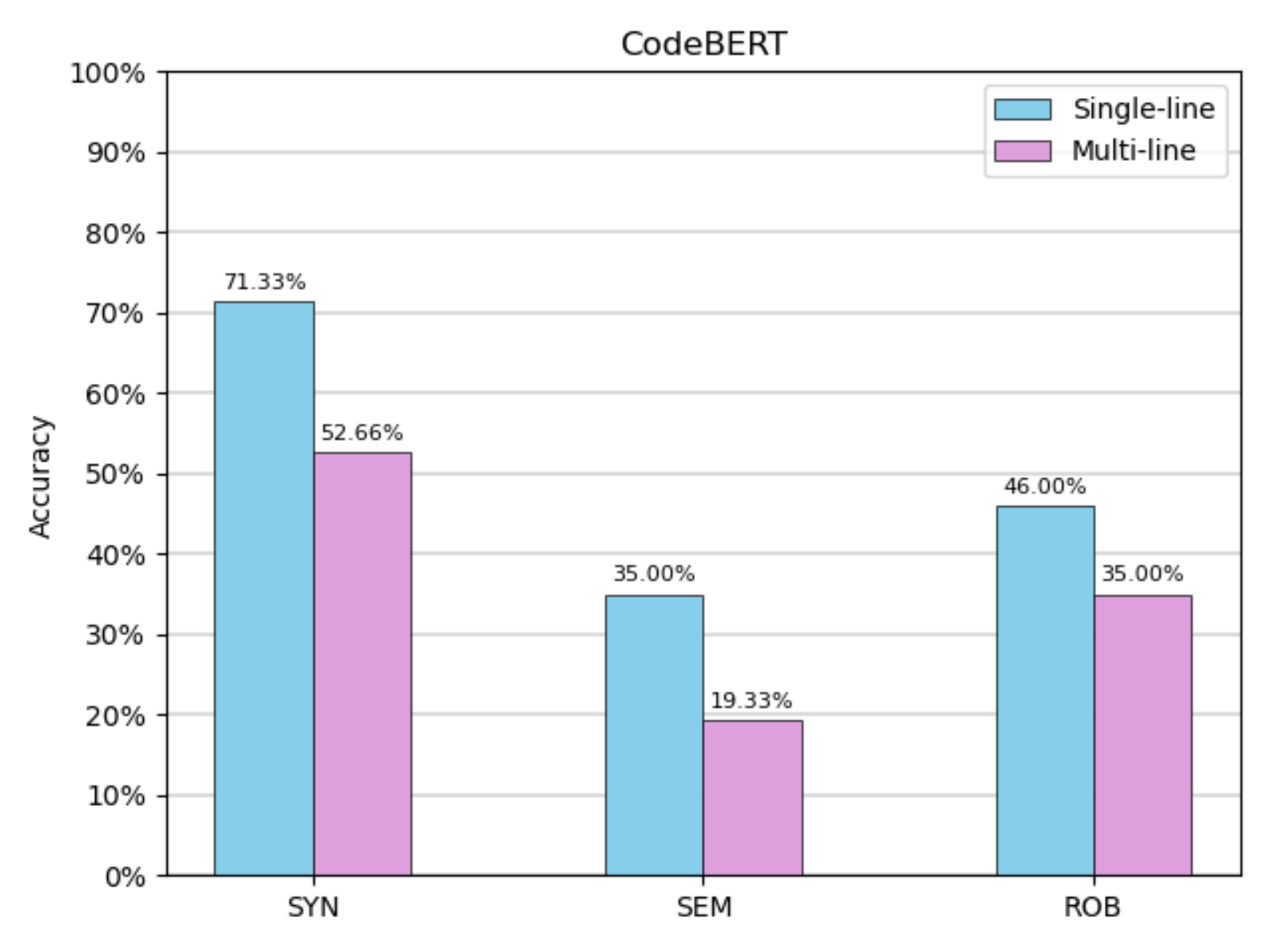}} \\
        
        \subfloat[Word Substitution]{\label{fig:codet5_word_substitution_61}
             \includegraphics[width=0.45\textwidth]{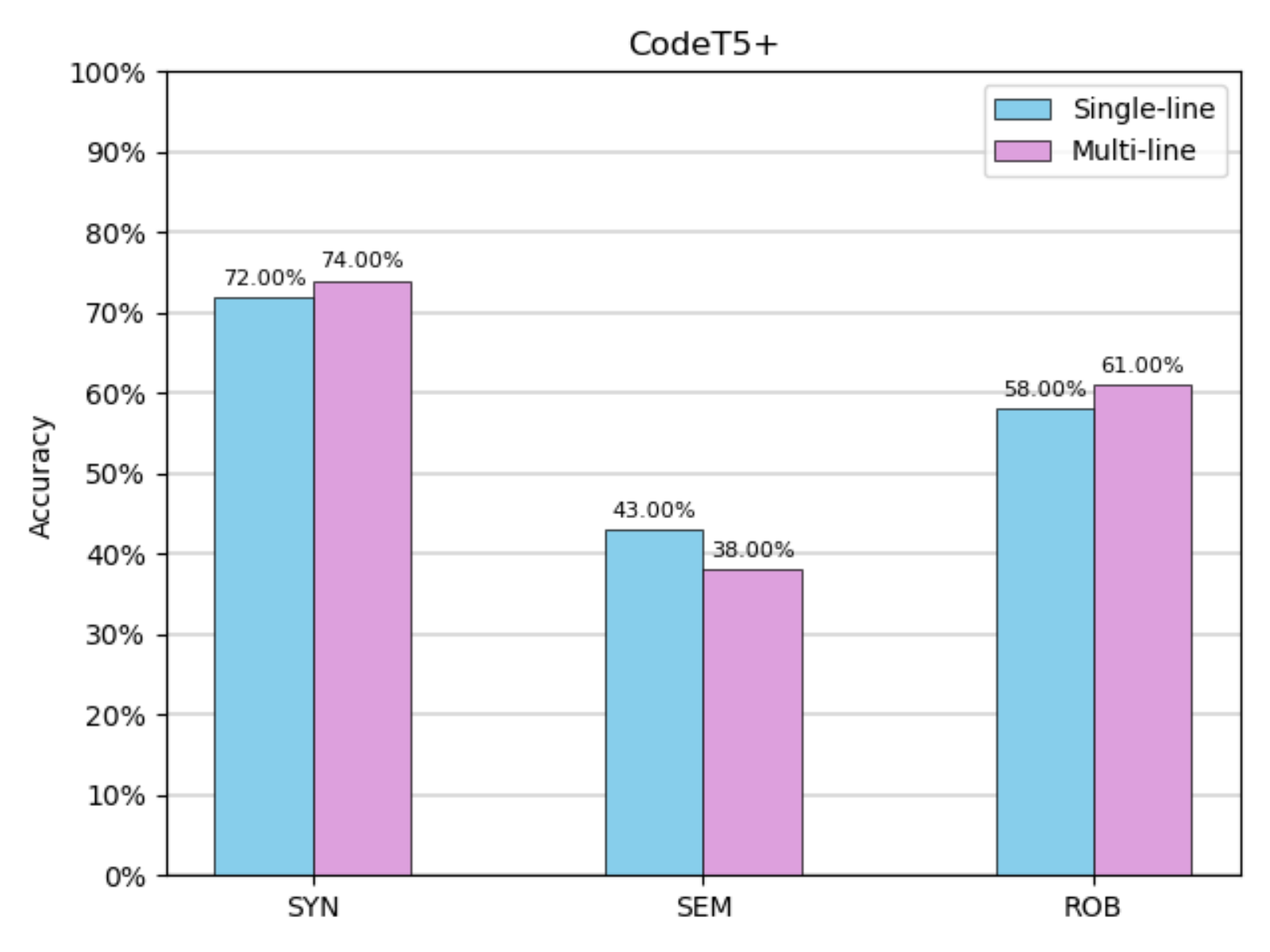}}
        \subfloat[Word Omission]{\label{fig:codet5_word_omission_61}
             \includegraphics[width=0.45\textwidth]{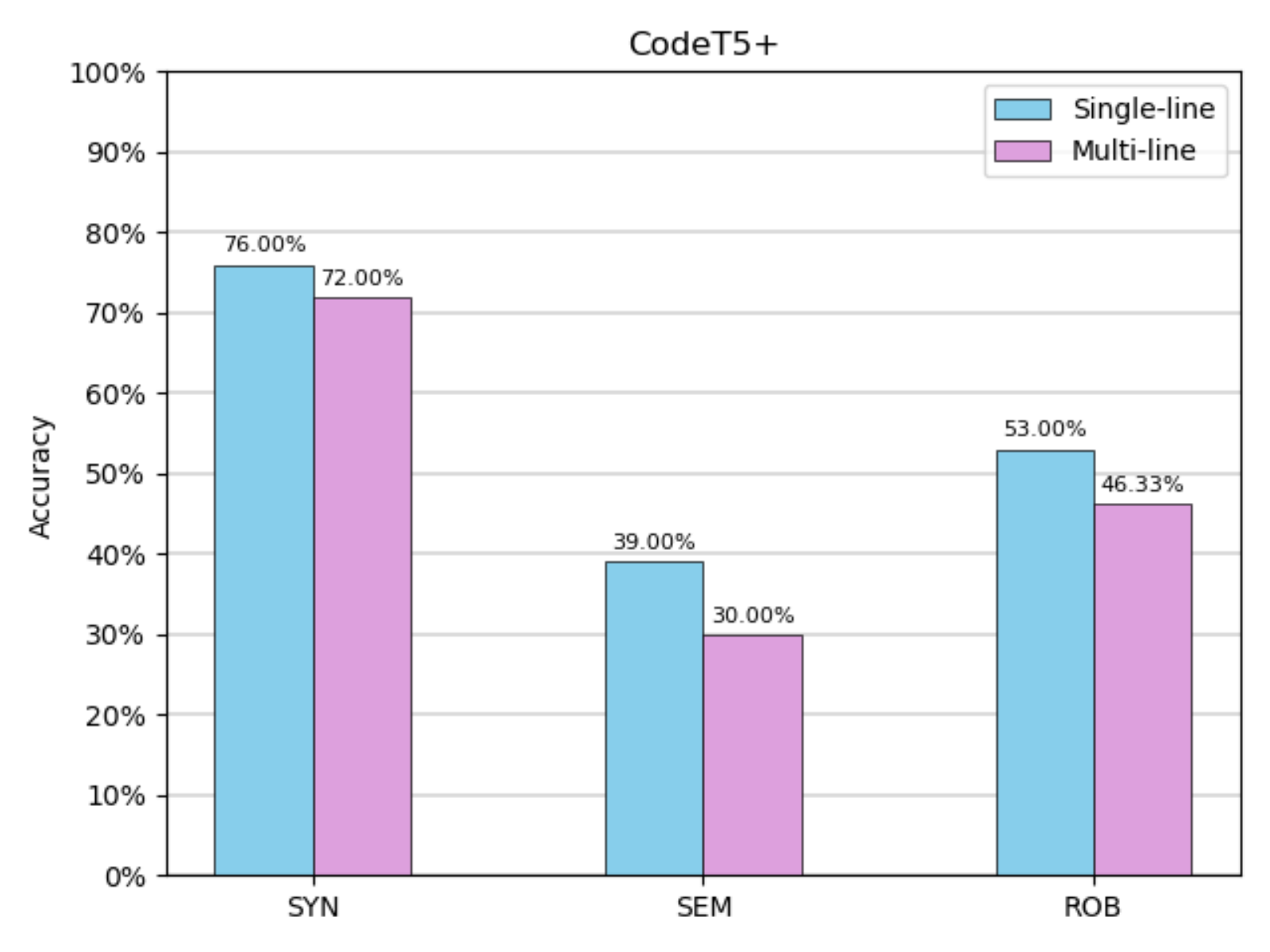}}
            
        \caption{Comparison between the models' performance on single-line code snippets \textit{vs.} multi-line code snippets when trained on the original data and tested on fully perturbed data in terms of syntactic (SYN), semantic (SEM), and robust accuracy (ROB). Figures on the \textbf{left} show the performance when data is perturbed with \textit{word substitution}, while figures on the \textbf{right} show the performance when data is perturbed with \textit{word omission}.}
\label{fig:single_multi_61}
\end{figure}

As an example, consider the sequence of NL intents ``Subtract 8 from the current byte in ESI" and ``Negate it", which translates to the instructions \texttt{sub byte [ESI], 8} and \texttt{not ESI}. While a human developer would immediately understand that ``it" refers to the previously mentioned ESI register, the models targeted in our experimental evaluation fail to make this connection, instead defaulting to a generic placeholder \texttt{not var} due to the lack of explicit mention of the register’s name. This scenario underscores a critical limitation of current AI models in code generation and their struggle with contextual understanding. To address these challenges, future work includes leveraging contextual information to aid AI models in the translation process, even when faced with missing or implicit information.

To further appreciate the effects of both perturbation strategies on the models' robustness, we compared the performance of the models on single-line and multi-line code snippets. As anticipated in \S{}~\ref{subsec:dataset}, 
the dataset contains intents that generate multiple lines of assembly code, separated by the newline character \textit{\textbackslash{n}}. In particular, the test set is composed of $449$ NL code descriptions that generate single-line snippets and $141$ that generate multi-line snippets for a total of $590$ samples.

\figurename~\ref{fig:single_multi_no_pertub} presents the three models' performance in the baseline setting, i.e., with no perturbation during training or testing, in terms of syntactic and semantic accuracy on single-line and multi-line snippets. 
Models perform better with shorter code, succeeding in generating up to 98\% of syntactically correct single-line snippets against 90\% of multi-line snippets. This gap in performance is further exacerbated for semantic accuracy, where models succeed on average on $\sim$71\% of single-line samples and only $\sim$56\% of multi-line samples, with CodeT5+ performing better than the other two models.

As depicted in \figurename~\ref{fig:single_multi_61}, when trained on the original data and tested on fully perturbed data, models behave similarly against word substitutions (\figurename~\ref{fig:seq2seq_word_substitution_61}, \figurename~\ref{fig:codebert_word_substitution_61}, \figurename~\ref{fig:codet5_word_substitution_61} on the left) and word omissions (\figurename~\ref{fig:seq2seq_word_omission_61}, \figurename~\ref{fig:codebert_word_omission_61}, \figurename~\ref{fig:codet5_word_omission_61} on the right), with the former having less negative impact on robustness. When dealing with word substitutions, models tend to generate more syntactically and semantically correct single-line snippets when compared to multi-line snippets, whereas robustness scores are approximately equal or even higher for multi-line snippets for newer pre-trained models like CodeT5+. This suggests that the word substitution perturbations do not affect the models’ ability to generate shorter or longer code snippets. Instead, when dealing with word omissions, models' performance on more complex multi-line snippets tends to deteriorate.

\begin{mybox}{\parbox{10.5cm}{RQ1: Are models robust to perturbations in the code descriptions?}}
In our experiments, we found that models are not robust to intents that diverge from those of the corpus. Indeed, the dropping of the performance is too evident even when the intents slightly differ from the original ones (i.e., word substitution).  
The worst-case scenario occurs when the intent lacks information. In this case, the semantic correctness of the prediction is more than halved.
Therefore, the model shows limited usability in practice since it is not able to deal with the variability of NL descriptions.
We attribute this limitation to the corpora used to train the models since they are often oriented to specific writing styles, or because descriptions in the corpus are often too literal and cumbersome.

Overall, our results show that pre-trained models like CodeBERT and CodeT5+ consistently outperform non-pre-trained models like Seq2Seq (+4\% performance boost). However, pre-trained models were shown to be more sensitive to input variations than non-pre-trained models when tested on unseen perturbed data, leading to a slight decrease in semantic accuracy for CodeBERT and a variable performance for CodeT5+ (i.e., a drop for word substitution and a boost for word omission). This slight performance drop can be attributed to the pre-trained models' sensitivity to input variations that they were not exposed to during their pre-training phase.
\end{mybox}

\subsection{Data Augmentation Against Perturbed Code Descriptions}
\label{subsec:ATperturbed}
We investigated how the use of perturbed intents in the training phase\footnote{We injected perturbations in both the training and validation sets.} improves the robustness of the models when dealing with perturbations in the test set.
Therefore, we randomly perturbed different amounts of the original code descriptions in the training set ($25\%$, $50\%$, and $100\%$), similarly to previous work~\cite{wu2020evaluating,huang2021robustness}.
The models were then trained from scratch using the different perturbed training sets before comparing the results obtained by the models without the use of data augmentation (DA), i.e., $0\%$ of perturbed intents in the training set (as shown in \S{}~\ref{subsec:robustness}). 
We remark that, in our experiments, we did not alter the size of the training data, i.e., the augmentation of the data refers to an increasing variability of the training examples obtained with the new, perturbed code descriptions, which replace the non-perturbed ones of the original dataset.

Data augmentation is an effective solution to increase the volume but especially the diversity of the data used to train AI models~\cite{feng-etal-2021-survey}, in order to improve their generalization abilities and performance on unseen data.
A twofold goal drove our experimental design choices: to ensure a fair comparison between the baseline and the results obtained after data augmentation and to assess the effectiveness of our data augmentation strategy thoroughly. 
To this aim, we kept the dataset size unaltered across the experiments to isolate the effect of the data augmentation strategy without the confounding variable of different dataset sizes. The data augmentation strategy was employed with the primary objective of increasing data diversity and enhancing the model's generalization capabilities. Then, by evaluating the model's performance on fully perturbed test sets, we were able to directly measure the impact of different amounts of data augmentation on the model's ability to handle diverse linguistic variations. 
Furthermore, when the fine-tuning dataset size is relatively large (i.e., $\sim$5,000 samples), the performance gain given by increasing the size is negligible (i.e., less than $\sim 0.5\%$)~\cite{wei2019eda}. Given the above considerations, we did not further increase the training data size.

It is also important to remark that, since the candidate word for replacement is chosen from a list of top candidates, the same word in different sentences might not always be replaced with the same synonym. The replacement can depend on several factors, including randomness, the list of candidate synonyms, and the selection criteria set in the augmentation process.
Without specific configurations to enforce determinism (e.g., choosing the seed), the augmentation process is non-deterministic and involves some level of randomness, especially if multiple suitable synonyms are available. In this case, the same word could be replaced with different synonyms in different NL descriptions. This approach adds variability by introducing different perturbations for the same word.
This is also proved by the size of the vocabulary (in terms of unique words) of the training data, which increases with the amount of perturbed data. In fact, we found that the original, non-perturbed training dataset contains 2536 words, while it increases to 2804, 2939 and 3094 with 25\%, 50\% and 100\% of perturbations in the NL descriptions, respectively.

\tablename{}~\ref{tab:adversarial} shows the results of data augmentation.
The table highlights that, for all the perturbations, the use of perturbed code descriptions increases the robustness of the model over all the metrics. 
For syntactic correctness, both models get a performance boost already with $25\%$ of word substitution in the training set ($+5\%$ for Seq2Seq, $+4\%$ for CodeBERT, $+17\%$ for CodeT5+), while in the case of word omission, we obtain the best results with a training set $100\%$ perturbed ($+11\%$ for Seq2Seq, $+27\%$ for CodeBERT, $+15\%$ for CodeT5+).

Semantic correctness analysis shows that all models gain greater benefits from the use of higher percentages of perturbed inputs ($>=50\%$).
In fact, in the case of the use of word substitution in the training set, Seq2Seq increases the number of semantically correct predictions up to $6\%$, and the robustness accuracy up to $21\%$, CodeBERT gets a boost in semantic correctness up to $17\%$ and robust accuracy up to $22\%$, and CodeT5+ shows an evident improvement up to $25\%$ and $34\%$.
In the case of the omissions in the training set, Seq2Seq improves the semantic and robustness accuracy up to $7\%$ and $16\%$, respectively, while CodeBERT gets an even more marked boost of the performance, increasing the semantic accuracy up to $16\%$ and robustness accuracy up to $22\%$. Finally, for CodeT5+, the semantic and robustness accuracy increased by $10\%$ and $13\%$.

\begin{table}[t]
\centering
\caption{Evaluation of the models trained with different percentages of perturbed code descriptions. The test set is 100\% perturbed. The worst performance per perturbation is \textcolor{red}{\textbf{red}}, while the best is \textcolor{blue}{\textbf{blue}}.}
\label{tab:adversarial}
\scriptsize
\begin{tabular}{ 
>{\arraybackslash}m{1.25cm} | 
>{\centering\arraybackslash}m{1cm} | 
>{\arraybackslash}m{0.75cm} 
>{\arraybackslash}m{0.75cm} 
>{\arraybackslash}m{0.75cm}|
>{\arraybackslash}m{0.75cm} 
>{\arraybackslash}m{0.75cm} 
>{\arraybackslash}m{0.75cm}|
>{\arraybackslash}m{0.75cm} 
>{\arraybackslash}m{0.75cm} 
>{\arraybackslash}m{0.75cm}}
\toprule
& & \multicolumn{3}{c}{\textbf{Seq2Seq}} & \multicolumn{3}{c}{\textbf{CodeBERT}} & \multicolumn{3}{c}{\textbf{CodeT5+}}\\
\textbf{Perturb.} & \textbf{Advers. Inputs} 
& \textbf{SYN} & \textbf{SEM} & \textbf{ROB} & \textbf{SYN} & \textbf{SEM} & \textbf{ROB} & \textbf{SYN} & \textbf{SEM} & \textbf{ROB}\\ \toprule
\multirow{4}{1.5cm}{\textit{Word Substitution}}  
& $0\%$ & \textcolor{red}{\Chart{0.86}} & \textcolor{red}{\Chart{0.51}} & \textcolor{red}{\Chart{0.66}} & \textcolor{red}{\Chart{0.89}} & \textcolor{red}{\Chart{0.49}} & \textcolor{red}{\Chart{0.68}} & \textcolor{red}{\Chart{0.73}} & \textcolor{red}{\Chart{0.42}} & \textcolor{red}{\Chart{0.58}}\\
& $25\%$ & \textcolor{blue}{\Chart{0.91}} & \Chart{0.52} & \Chart{0.80} & \textcolor{blue}{\Chart{0.93}} & \Chart{0.62} & \Chart{0.86} & \textcolor{blue}{\Chart{0.90}} & \Chart{0.63} & \Chart{0.88}\\
& $50\%$ & \textcolor{blue}{\Chart{0.91}} & \textcolor{blue}{\Chart{0.57}} & \Chart{0.85} & \Chart{0.92} & \textcolor{blue}{\Chart{0.66}} & \textcolor{blue}{\Chart{0.90}} &\Chart{0.88} & \Chart{0.62} & \Chart{0.87}\\
& $100\%$ & \textcolor{blue}{\Chart{0.91}} & \textcolor{blue}{\Chart{0.57}} & \textcolor{blue}{\Chart{0.87}} & \Chart{0.92} & \Chart{0.64} & \Chart{0.88} & \textcolor{blue}{\Chart{0.90}} & \textcolor{blue}{\Chart{0.67}} & \textcolor{blue}{\Chart{0.92}}\\\midrule
\multirow{4}{1.5cm}{\textit{Word Omission}}  
& $0\%$ & \textcolor{red}{\Chart{0.81}} & \textcolor{red}{\Chart{0.33}} & \textcolor{red}{\Chart{0.45}} & \textcolor{red}{\Chart{0.67}} & \textcolor{red}{\Chart{0.32}} & \textcolor{red}{\Chart{0.44}} & \textcolor{red}{\Chart{0.75}} & \textcolor{red}{\Chart{0.37}} & \textcolor{red}{\Chart{0.51}}\\
& $25\%$ & \Chart{0.89} & \Chart{0.38} & \Chart{0.57} & \Chart{0.89} & \Chart{0.45} & \Chart{0.61} & \Chart{0.89} &\Chart{0.46} & \Chart{0.62}\\
& $50\%$ & \Chart{0.90} & \Chart{0.39} & \textcolor{blue}{\Chart{0.61}} & \Chart{0.91} & \Chart{0.46} & \Chart{0.63} & \Chart{0.89} & \textcolor{blue}{\Chart{0.47}} & \textcolor{blue}{\Chart{0.64}}\\
& $100\%$ & \textcolor{blue}{\Chart{0.92}} & \textcolor{blue}{\Chart{0.40}} & \Chart{0.60} & \textcolor{blue}{\Chart{0.94}} & \textcolor{blue}{\Chart{0.48}} & \textcolor{blue}{\Chart{0.66}} & \textcolor{blue}{\Chart{0.90}} & \textcolor{blue}{\Chart{0.47}} & \Chart{0.63}\\
\bottomrule
\end{tabular}
\end{table}

To show the improvement in the code generation obtained with the data augmentation, \tablename{}~\ref{tab:qualitative_pert} presents a qualitative analysis using cherry-picked examples from our test sets. In particular, the table shows examples of failed (i.e., semantically incorrect) predictions without using the data augmentation that resulted in successful (i.e., semantically correct) predictions after the augmentation of the code descriptions.
In the case of word substitution, it is worth noticing that the model failed to predict the correct output even if the perturbed intent slightly differs from the original one. 
As concerns the omission of words, the table points out that the model's successful prediction is constrained to all the parts of the description, hence the removal of an unnecessary word from the intent (e.g., the word \textit{function}) implied a failure prediction. The table also highlights that all these failure cases were fixed by the use of data augmentation since the predictions are all syntactically and semantically correct.

\begin{table*}[ht]
\centering
\caption{Illustrative examples of predictions with and without data augmentation (DA) against the perturbed test set. The prediction errors are \textcolor{red}{\textbf{red}}. \textbf{\underline{Underlined}} and \textbf{\cancel{slashed}} text refer to word substitution and to word omission in the test set, respectively.}
\label{tab:qualitative_pert}
\footnotesize
\begin{tabular}{  
>{\centering\arraybackslash}m{2cm} | 
>{\centering\arraybackslash}m{2cm} |
>{\centering\arraybackslash}m{2cm} |
>{\centering\arraybackslash}m{2cm} |
>{\centering\arraybackslash}m{2cm} }
\toprule
\textbf{Original intent} & \textbf{Prediction before perturbing the intent} & \textbf{Perturbed Intent} &  \textbf{Prediction w/o DA after perturbing the intent} & \textbf{Prediction with DA after perturbing the intent} \\ \midrule
\textit{subtract the contents of al from the bl register} &
\texttt{sub bl, al} &
\textit{\underline{deducting} the contents of al from the bl register} &
\texttt{\textcolor{red}{or} al, bl} &
\texttt{sub bl, al}\\ \midrule
\textit{push double word 0x74652f2f onto the stack and point the EBX register to the stack register} &
\texttt{push dword 0x74652f2f \textbackslash{n} mov ebx, esp} &
\textit{\cancel{push} double word 0x74652f2f onto the stack and \cancel{point} the EBX register to the stack register} &
\texttt{push dword 0x74652f2f \textbackslash{n} \textcolor{red}{push dword ebx \textbackslash{n}} mov ebx, esp} &
\texttt{push dword 0x74652f2f \textbackslash{n} mov ebx,  esp}\\ \midrule
\textit{call \_myfunc function} & \texttt{call \_myfunc} & \textit{call \_myfunc \cancel{function}} & \texttt{\textcolor{red}{int 0x80}}  & \texttt{call \_myfunc} \\
\bottomrule
\end{tabular}
\end{table*}

After showing that the use of perturbations in the training set is an effective solution to increase the robustness of the models, we compared the performance of the models with data augmentation on the perturbed test set with the performance of the models on the original training/test sets, i.e., without perturbations.
We aimed at examining the best data augmentation setting against the baseline (i.e., non-perturbed training and test sets), to understand whether our data augmentation strategy could, at any fixed perturbation rate, achieve the baseline performance which is considered optimal.
According to  \tablename{~\ref{tab:adversarial}}, perturbing 100\% of the training data provides the best or closer to the best scores when models are tested on 100\% perturbed data.


\begin{figure}[t]
    \centering
    \subfloat[Syntactic Accuracy (SYN)]{\label{fig:syntax_AT_100}
         \includegraphics[width=0.5\textwidth]{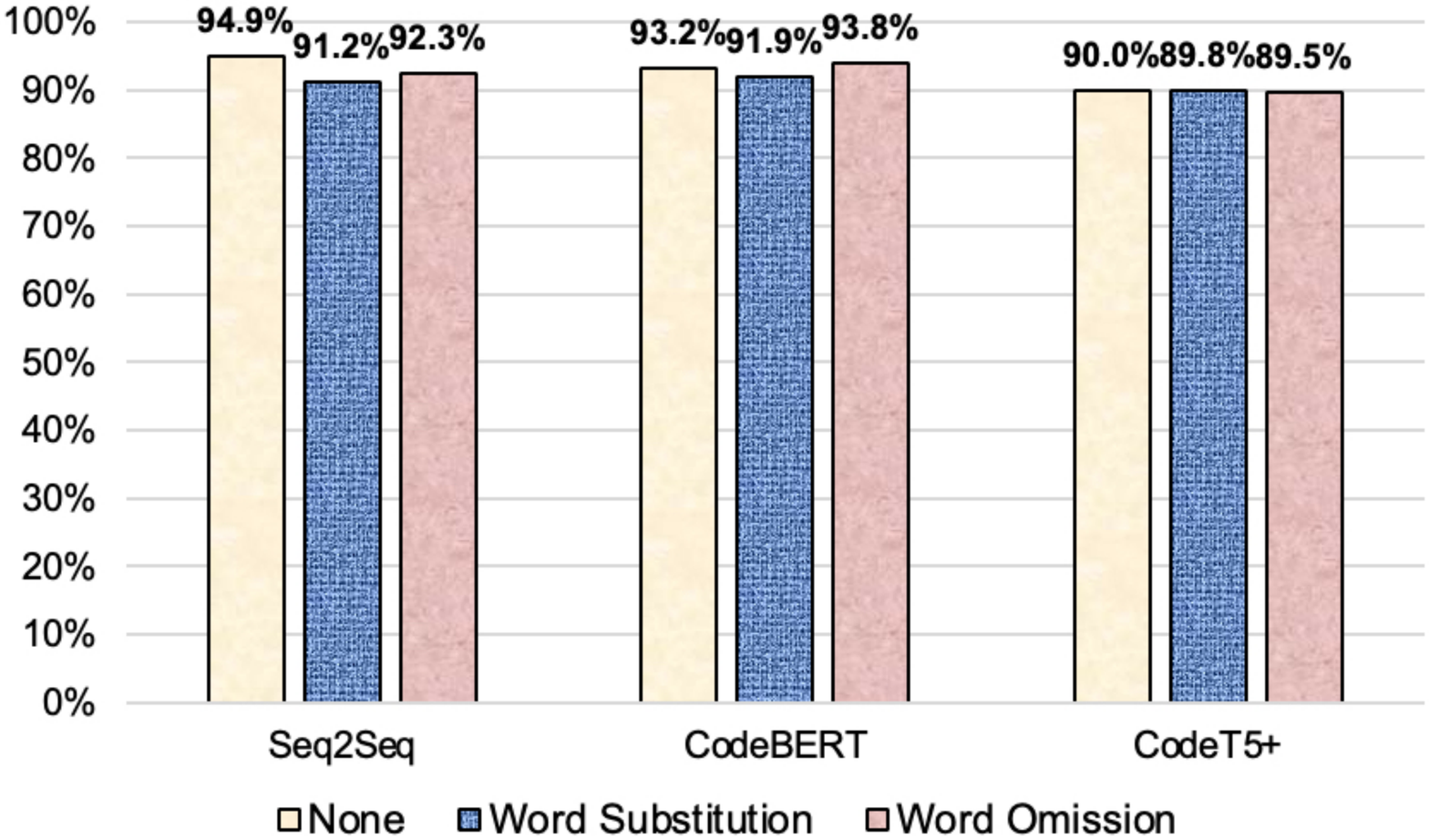}}
   \subfloat[Semantic Accuracy (SEM)]{\label{fig:semantics_AT_100}
         \includegraphics[width=0.5\textwidth]{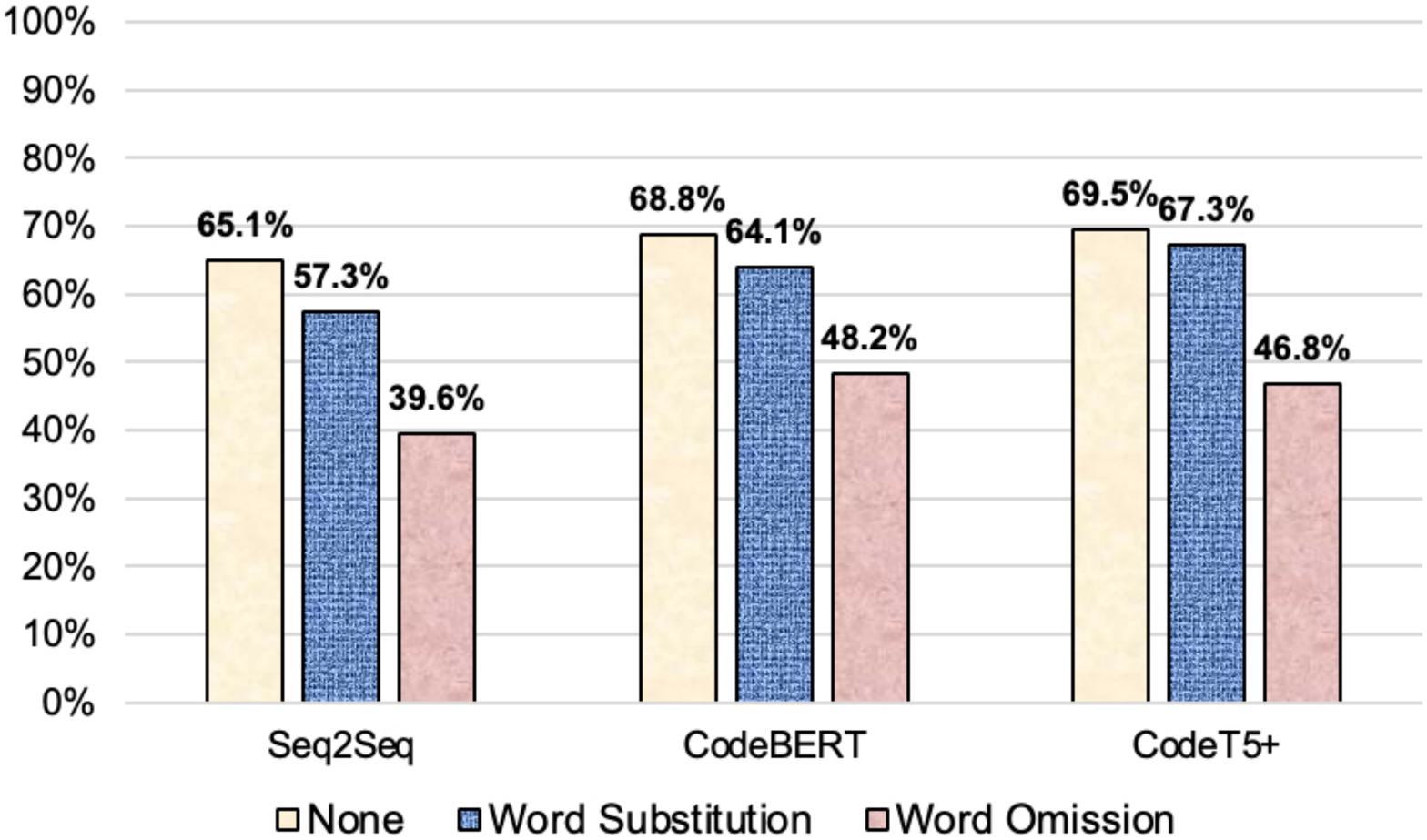}}

\caption{Comparison between the results obtained by models with data augmentation on the perturbed test set against the results obtained on the original, non-perturbed training/test set (\textit{None}).  
The \textit{None} bar represents the baseline model performance, i.e., 0\% perturbed training and test sets, whereas the \textit{Word Substitution} and \textit{Word Omission} bars represent the best model performance when trained on data 100\% augmented with the word substitution and word omission perturbations and tested on a fully (100\%) perturbed test set, respectively.}
\label{fig:best_AT}
\end{figure}

\figurename~\ref{fig:best_AT} shows the results in terms of syntactic and semantic accuracy. We did not include robust accuracy since this metric cannot be evaluated without perturbations in the test set (i.e., for the \textit{None} bar).
\figurename{~\ref{fig:syntax_AT_100}} highlights that the syntactic correctness provided by the models with data augmentation on perturbed test sets is comparable to the one obtained on the original, non-perturbed training/test sets. Furthermore, the use of omissions in the training set helped CodeBERT to exceed the performances obtained without perturbations (93.84\% vs 93.22\%).  
However, the analysis of semantics (\figurename{~\ref{fig:semantics_AT_100}}) led to very different results. Indeed, all models provide performance lower than that obtained without perturbations. In the case of word substitution, Seq2Seq and CodeBERT showed a drop in the performance of $8\%$ and $4\%$, while CodeT5+ is very close to the results of the original training/test set ($2.2\%$ decrease in semantic accuracy).
The decrement of semantic accuracy is more pronounced in the case of omissions in the NL intents. Indeed, despite the use of data augmentation, Seq2Seq, CodeBERT, and CodeT5+ show a clear drop of $25\%$, $21\%$, and $23\%$, respectively.

A key takeaway of our experimental evaluation is that AI code generation models are extremely brittle to unseen perturbations (i.e., non-perturbed training set, fully perturbed test set), but when trained on the augmented data (i.e., perturbed training set, fully perturbed test set) their performance greatly improves. However, it still does not achieve the same performance as the baseline (non-perturbed training set, non-perturbed test set). Future work includes providing the models with contextual information to help them gather the missing information from the surrounding context (i.e., previous sentences).

\begin{table}[t]
\centering
\caption{Jensen-Shannon divergence (JSD) values indicating the divergence between various training and test set distributions.}
\label{tab:jsd_values}
\footnotesize
\begin{tabular}{>{\arraybackslash}m{5cm} | >{\centering\arraybackslash}m{3.5cm} | >{\centering\arraybackslash}m{1.5cm}}
\toprule
\textbf{Comparison} & \textbf{Perturbation Type} & \textbf{JSD Value} \\
\midrule
\textit{JSD between Original Training and Original Test Set} & None & 0.29 \\
\midrule
\multirow{4}{5cm}{\textit{JSD between Original Training and Perturbed Test Sets}} & Omitted Action Words & 0.38 \\
& Omitted Name Words & 0.36 \\
& Omitted Structure Words & 0.36 \\
& Word Substitution & 0.40 \\
\midrule
\multirow{4}{5cm}{\textit{JSD between Perturbed Training (50\%) and Perturbed Test Sets}} & Omitted Action Words & 0.34 \\
& Omitted Name Words & 0.32 \\
& Omitted Structure Words & 0.33 \\
& Word Substitution & 0.33 \\
\bottomrule
\end{tabular}
\end{table}

To verify if augmented data is sufficient to evaluate the robustness of models, we assessed if perturbed data results in different training/test sets in terms of their distributions. To this aim, we analyzed the data distribution differences between the various training and test sets using Jensen-Shannon divergence (JSD). The JSD ranges from 0 (identical distributions) to 1 (completely different distributions). 

\begin{figure}[t]
    \centering
        \subfloat[Word Substitution]{\label{fig:seq2seq_word_substitution_62}
             \includegraphics[width=0.45\textwidth]{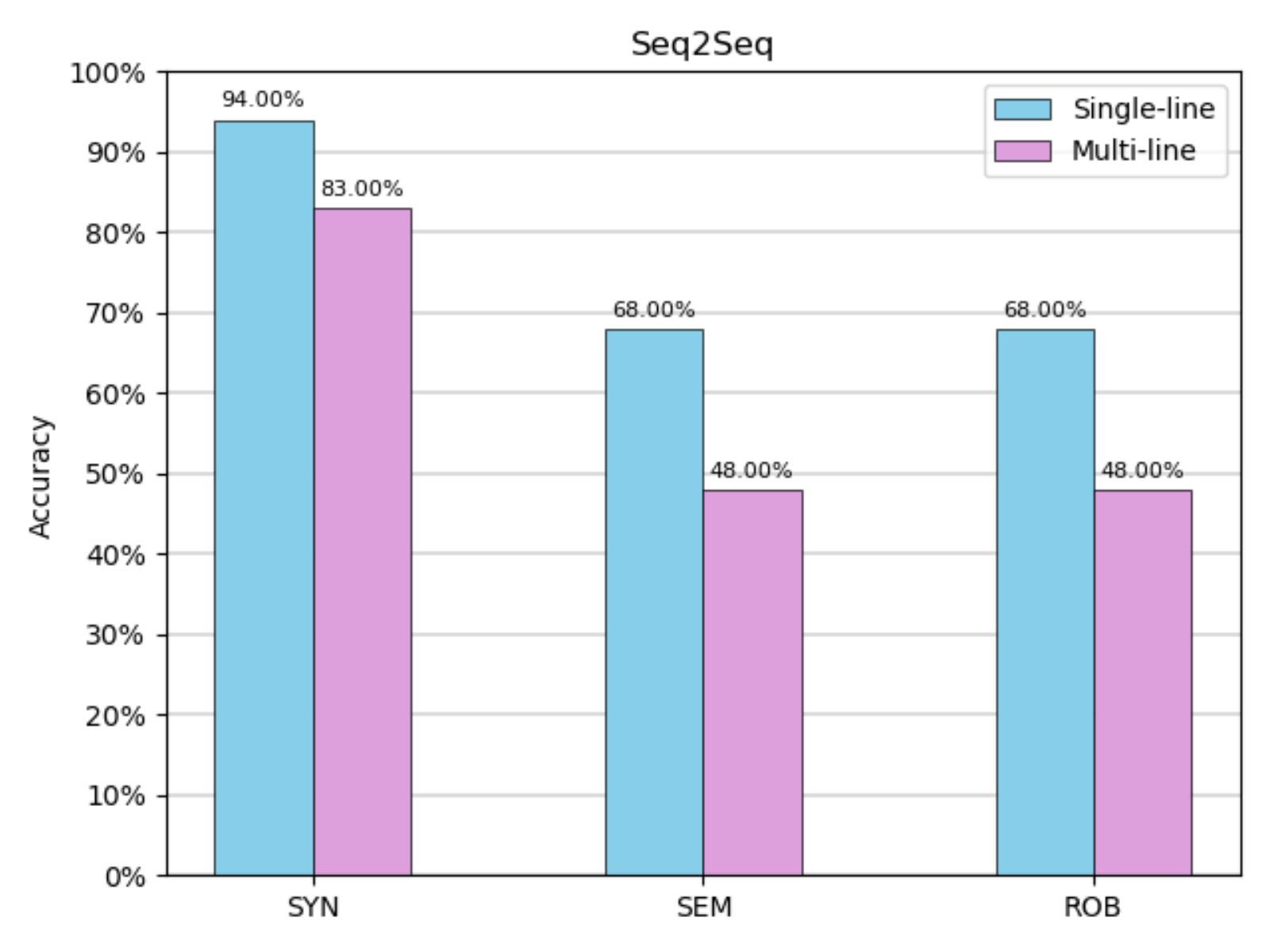}}
        \subfloat[Word Omission]{\label{fig:seq2seq_word_omission_62}
             \includegraphics[width=0.45\textwidth]{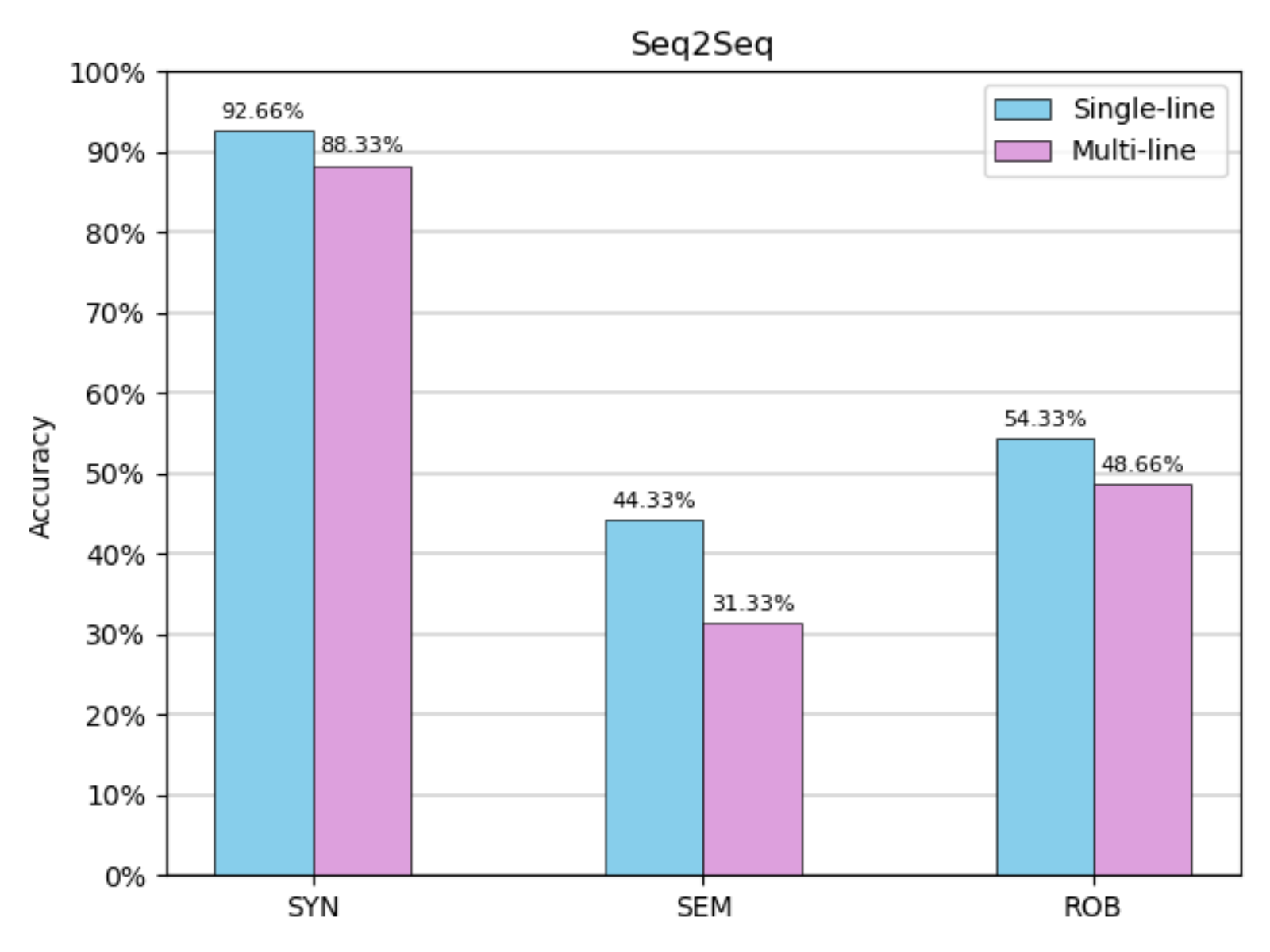}} \\
        
        \subfloat[Word Substitution]{\label{fig:codebert_word_substitution_62}
             \includegraphics[width=0.45\textwidth]{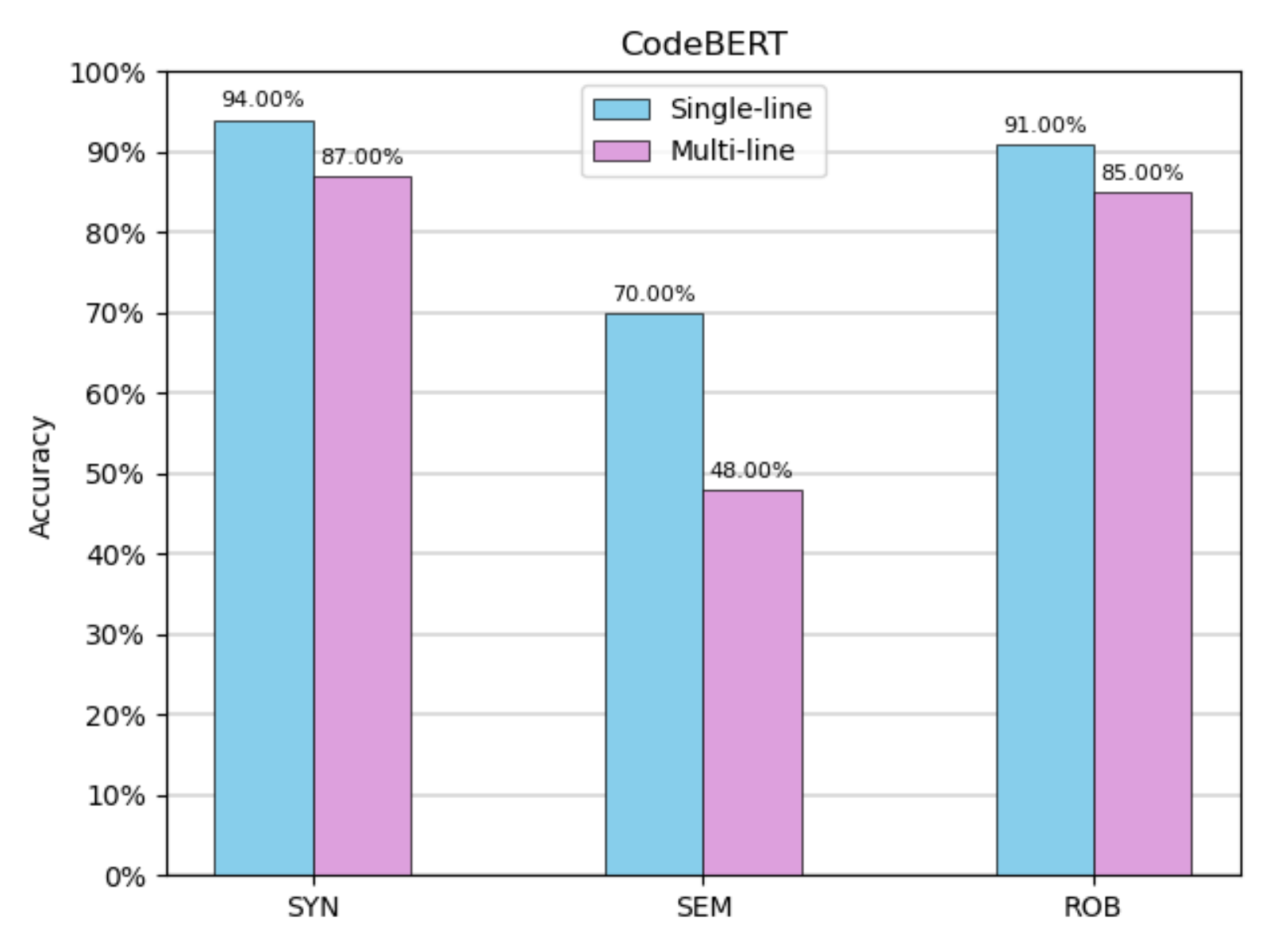}}
        \subfloat[Word Omission]{\label{fig:codebert_word_omission_62}
             \includegraphics[width=0.45\textwidth]{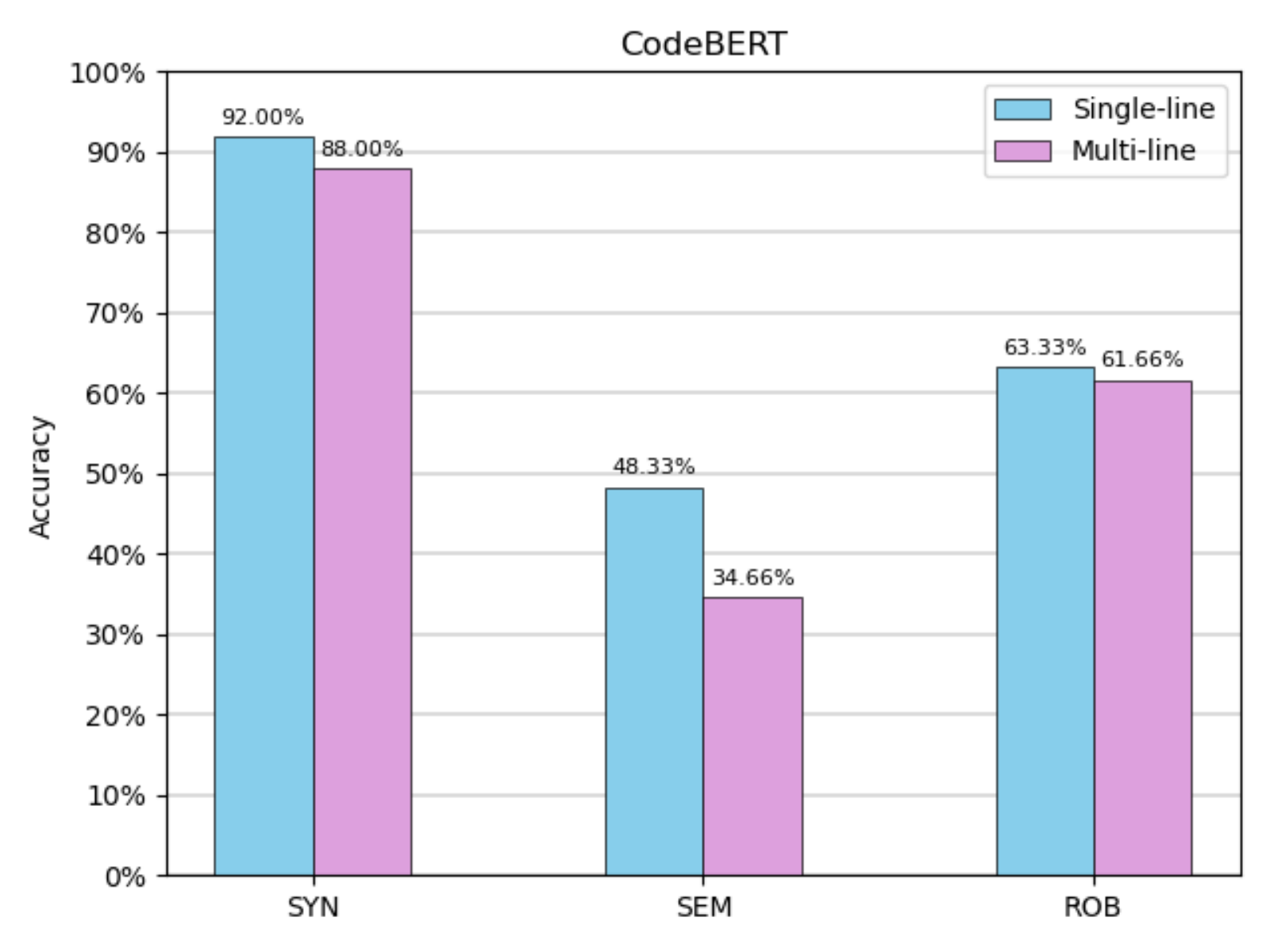}} \\
        
        \subfloat[Word Substitution]{\label{fig:codet5_word_substitution_62}
             \includegraphics[width=0.45\textwidth]{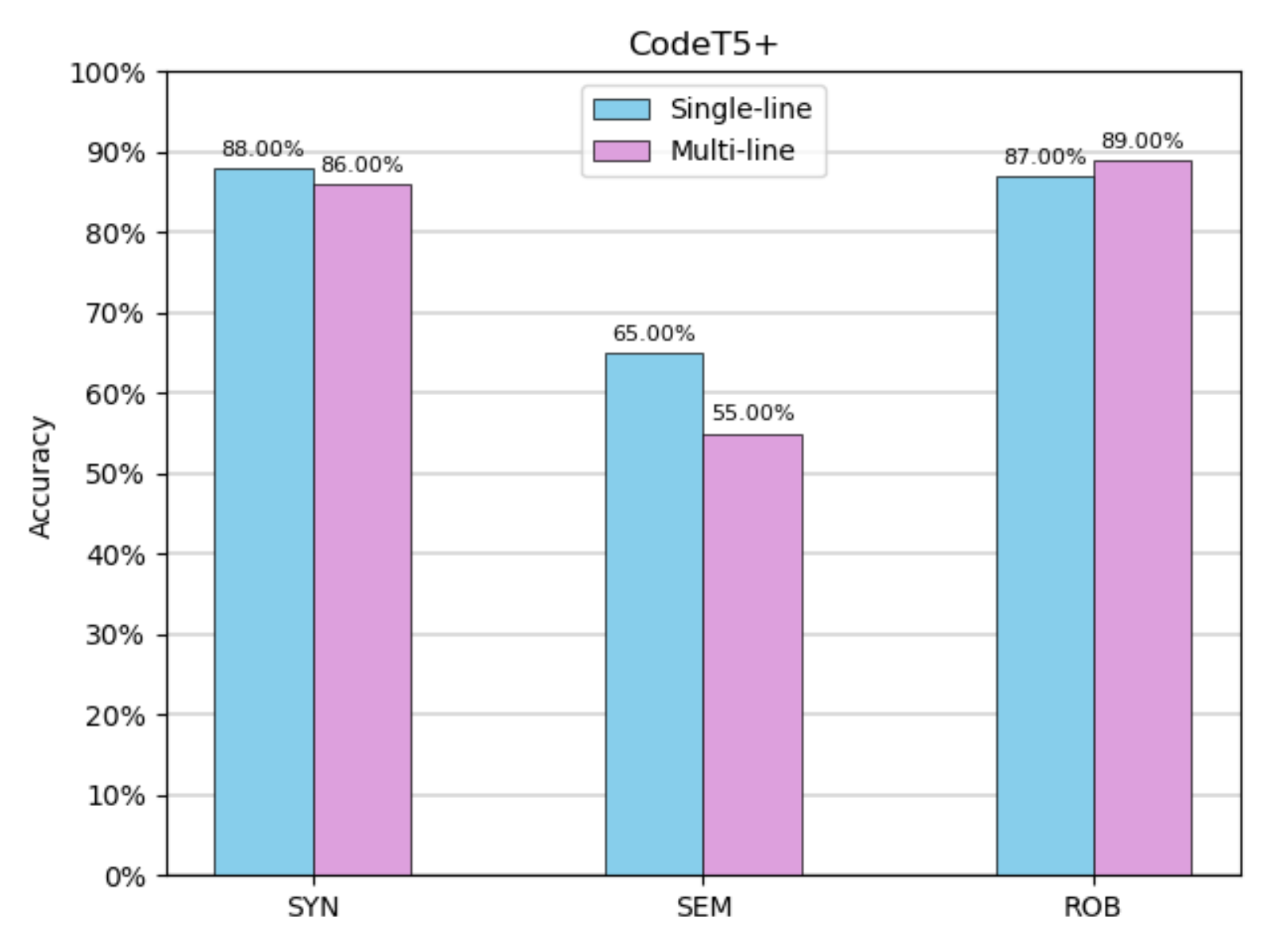}}
        \subfloat[Word Omission]{\label{fig:codet5_word_omission_62}
             \includegraphics[width=0.45\textwidth]{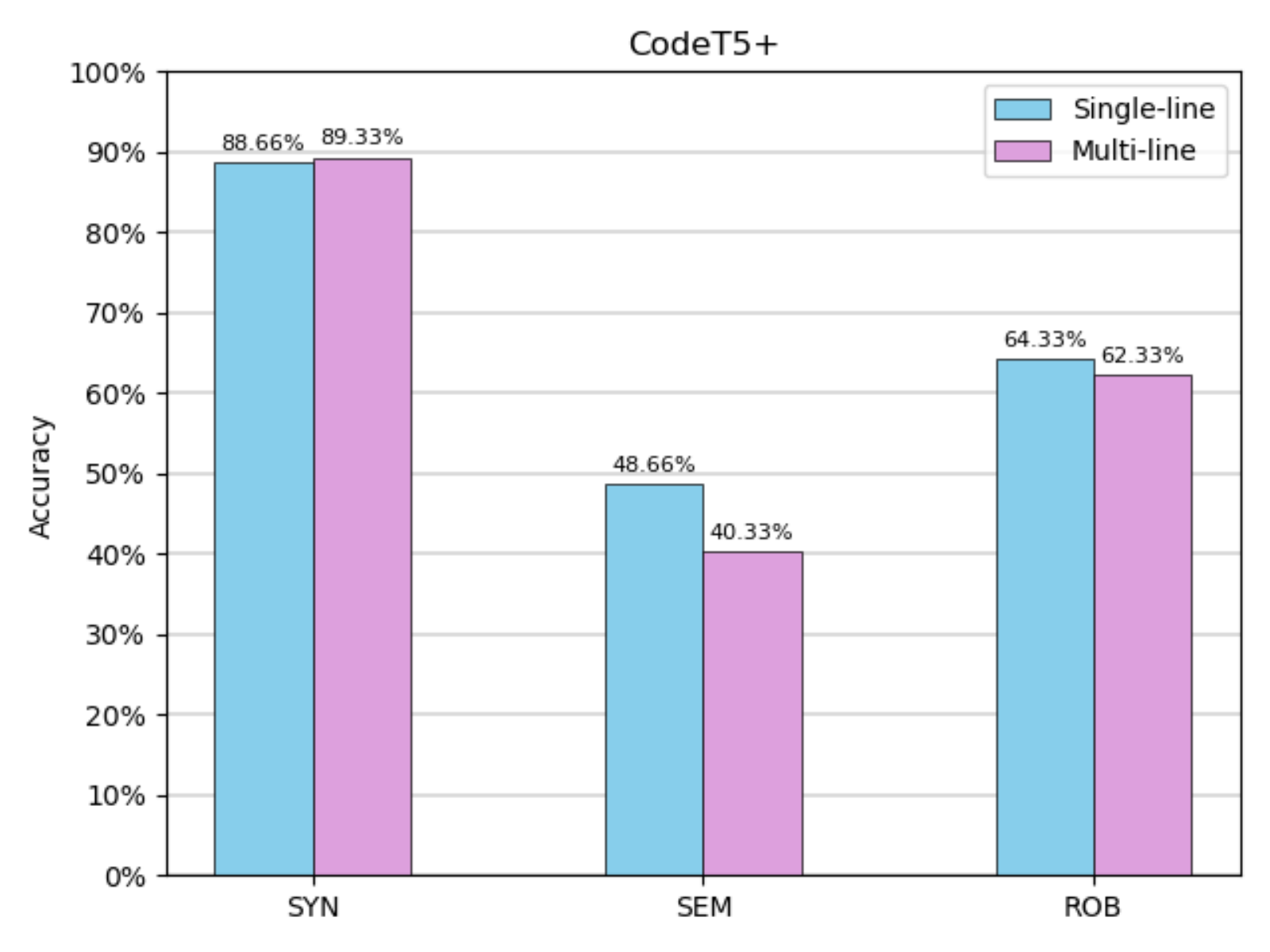}}

        \caption{Comparison between the models' performance on single-line code snippets \textit{vs.} multi-line code snippets when trained on partially augmented data (50\%) and tested on fully perturbed data in terms of syntactic (SYN), semantic (SEM), and robust accuracy (ROB). Figures on the \textbf{left} show the performance when data is perturbed with \textit{word substitution}, while figures on the \textbf{right} show the performance when data is perturbed with \textit{word omission}.}
\label{fig:single_multi_62}
\end{figure}

Table~\ref{tab:jsd_values} shows JSD values indicate varying degrees of divergence between the training and test sets across different perturbations.
The table shows that the JSD between the original training and the original test set: is 0.29. This value serves as a baseline, indicating moderate divergence between the original training and test sets.
The JSD values between the original training and perturbed test sets, instead, are higher than the baseline (Omitted Action Words: 0.38, Omitted Name Words: 0.36, Omitted Structure Words: 0.36, Word Substitution: 0.40). These higher values demonstrate that the perturbations applied to the test set significantly altered the data distribution compared to the original training set. This also suggests that the model might struggle to generalize without being trained on similar data, indicating a robustness issue (as shown in \S{}~\ref{subsec:robustness}).
Finally, the table shows that JSD values between perturbed training (50\%) and perturbed test (Omitted Action Words: 0.34, Omitted Name Words: 0.32, Omitted Structure Words: 0.33, Word Substitution: 0.33) are still higher than the baseline, but lower than the previous case. These intermediate values indicate that while the perturbed training and test distributions are different, they are not as drastically different as the distributions between the original training and the perturbed test sets. This suggests that training on perturbed data (data augmentation) helps the model become more robust against similar perturbations.
The JSD analysis supports our claim that these perturbations impact data distributions, and training on such data improves the model’s ability to handle these variations, thus focusing on robustness rather than just generalization.

\figurename~\ref{fig:single_multi_62} compares the models' performance on single-line and multi-line snippets after data augmentation. Indeed, when trained on partially augmented data (50\%) and tested on fully perturbed data, syntactic accuracy remains relatively stable across different perturbations, whereas semantic accuracy is notably higher in single-line descriptions, particularly for Seq2Seq and CodeBERT, where multi-line descriptions show a significant drop. 
As for robust accuracy, both pre-trained models show high scores on multi-line snippets for both perturbation types, unlike Seq2Seq, whose ability to generate complex code is more negatively affected. 
Interestingly, data augmentation proves to be the most effective on largely pre-trained models like CodeT5+, for which the gap of performance on single-line and multi-line snippets is less evident than Seq2Seq and CodeBERT.

\begin{mybox}{\parbox{10.5cm}{RQ2: Can data augmentation enhance the robustness of the models against perturbed code descriptions?}}
Data augmentation proved to be an effective solution to improve the robustness of models in the generation of offensive code. Indeed, our experiments showed that when the model is trained with a high percentage of perturbed intents, it can deal with the variability of the descriptions.
For all the perturbations, the syntactic correctness of the predictions with the data augmentation is comparable to the one obtained by the model on the original, non-perturbed test set. 
However, the semantic correctness of the predictions on perturbed code descriptions is still lower than the one obtained on the original, non-perturbed data. This drop is more evident when the models deal with the omission of information. This suggests the need for more sophisticated solutions to derive the missing or implicit information from the context of the program~\cite{lutellier2020coconut,wu2022study}.

Moreover, our results provide valuable insights about the impact of pre-training on the robustness of AI-driven solutions for offensive code generation. Indeed, when pre-trained models are fine-tuned and then tested on perturbed data, they regain their advantage and outperform non-pre-trained models, with a performance boost up to 10\% and 11\% for CodeBERT and CodeT5+, respectively. The exposure to diverse and varied inputs during the training phase helps pre-trained models leverage their extensive pre-training knowledge, allowing them to adapt to and handle the perturbations effectively.
\end{mybox}

\subsection{Data Augmentation Against Non-perturbed Code Descriptions}
\label{subsec:ATnonperturbed}

In our last analysis, we assessed whether the use of perturbed code descriptions in the training phase improves the performance of the model against non-perturbed NL code descriptions. Therefore, we added perturbed inputs in the 50\% of the training set, while the remaining 50\% is not perturbed, and evaluated the performance of the models against non-perturbed data.
The key idea is to understand whether the models gain benefits from the data augmentation when generating snippets starting from the original non-perturbed intents.

The rationale for perturbing 50\% of the training set instead of alternative percentages was guided by our findings in RQ2, where different perturbation ratios were tested. The perturbing half of the training set showed a good trade-off between maintaining performance on non-perturbed data and enhancing robustness against varied inputs. These findings motivated our decision to use the same percentage in RQ3 to validate the effectiveness of our data augmentation strategy on the original clean data. 
Indeed, in RQ3, our goal is to identify the optimal data augmentation setting that balances robustness and performance without being too aggressive or too mild. Perturbing 100\% of the training data could overly distort the training data, whereas a 25\% perturbation might not introduce enough diversity to significantly impact the model's robustness. Therefore, a 50\% perturbation was chosen to compromise between the model being exposed to sufficient variation to improve generalization and maintain the integrity of the training data.

\begin{figure}[ht]
    \centering
    \subfloat[Syntactic Accuracy (SYN)]{\label{fig:syntax_Test0}
         \includegraphics[width=0.5\textwidth]{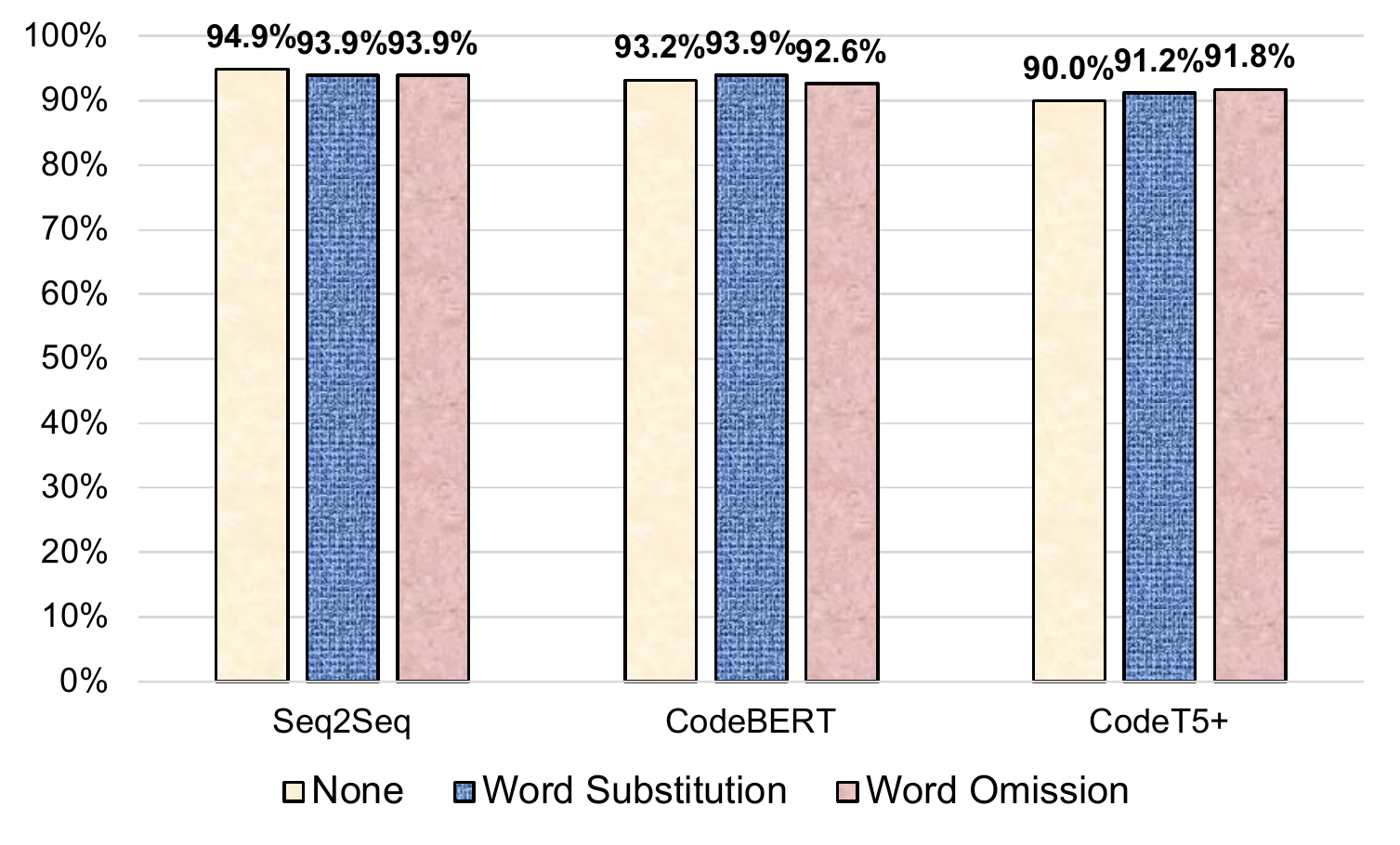}}
   \subfloat[Semantic Accuracy (SEM)]{\label{fig:semantics_Test0}
         \includegraphics[width=0.5\textwidth]{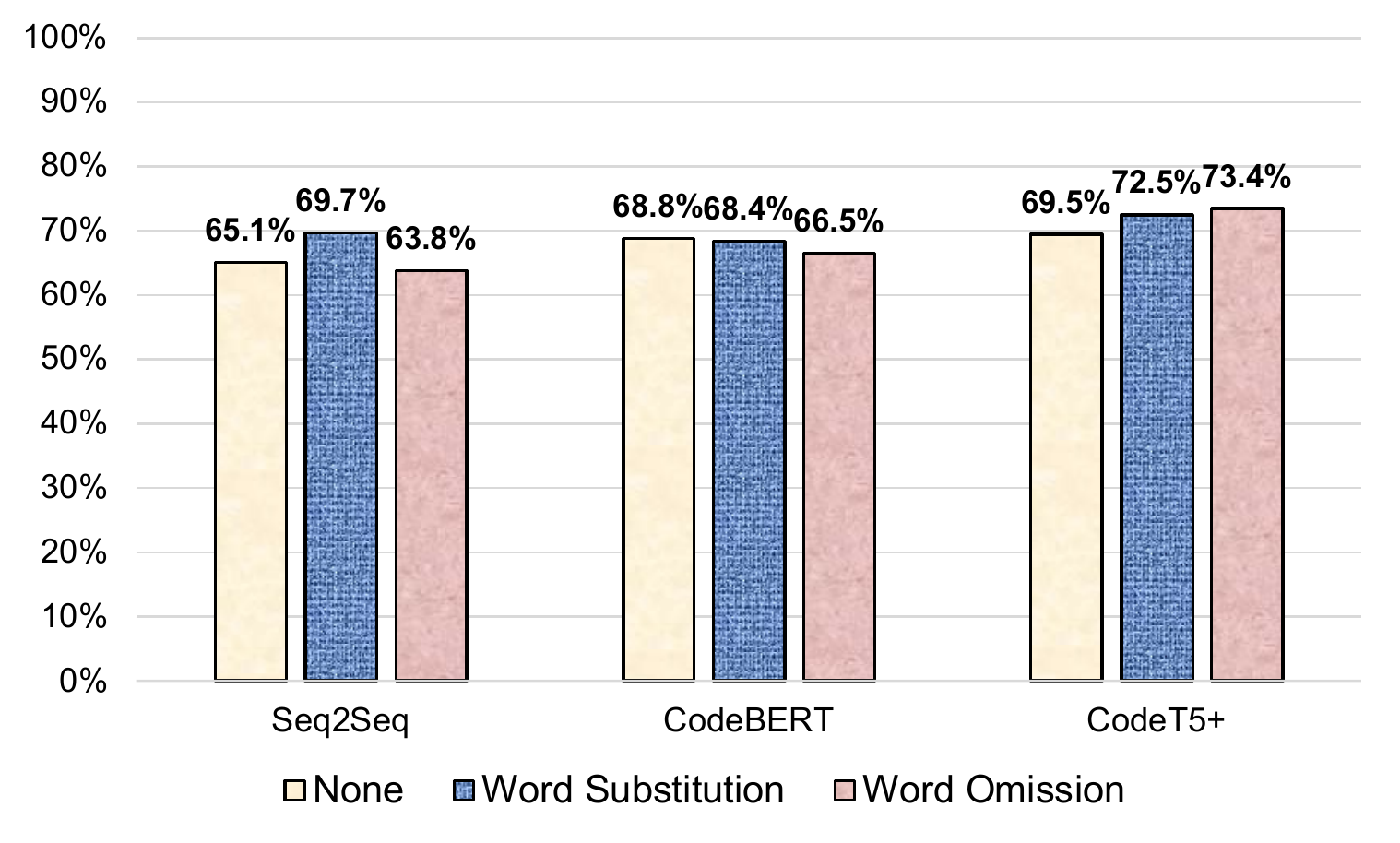}}
\caption{Comparison between results obtained by models with data augmentation on the original, non-perturbed test set against the results obtained on the original, non-perturbed training/test set (\textit{None}).}
\label{fig:clean_test}
\end{figure}

\begin{figure}[t]
    \centering
        \subfloat[Word Substitution]{\label{fig:seq2seq_word_substitution_63}
             \includegraphics[width=0.45\textwidth]{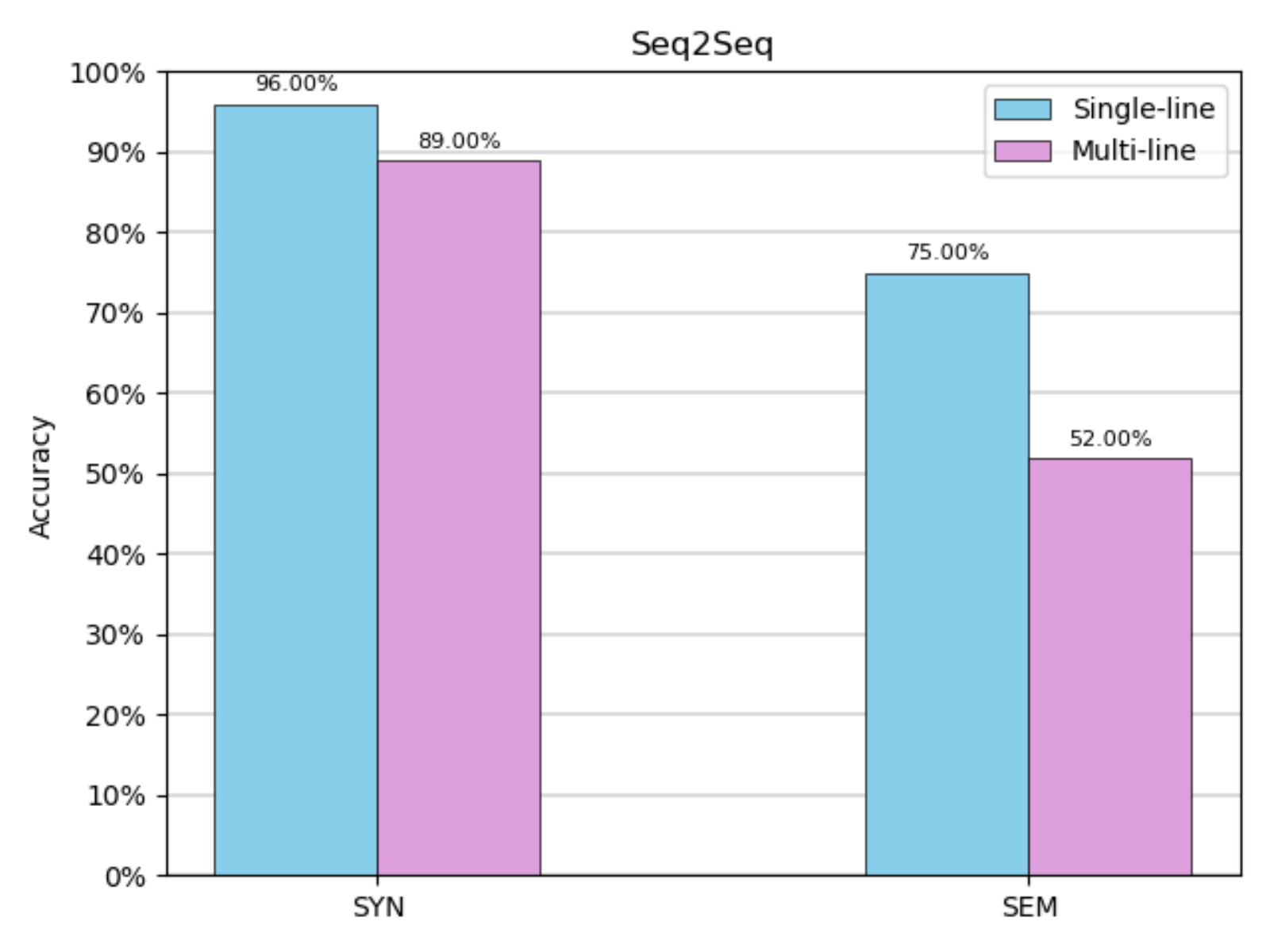}}
        \subfloat[Word Omission]{\label{fig:seq2seq_word_omission_63}
             \includegraphics[width=0.45\textwidth]{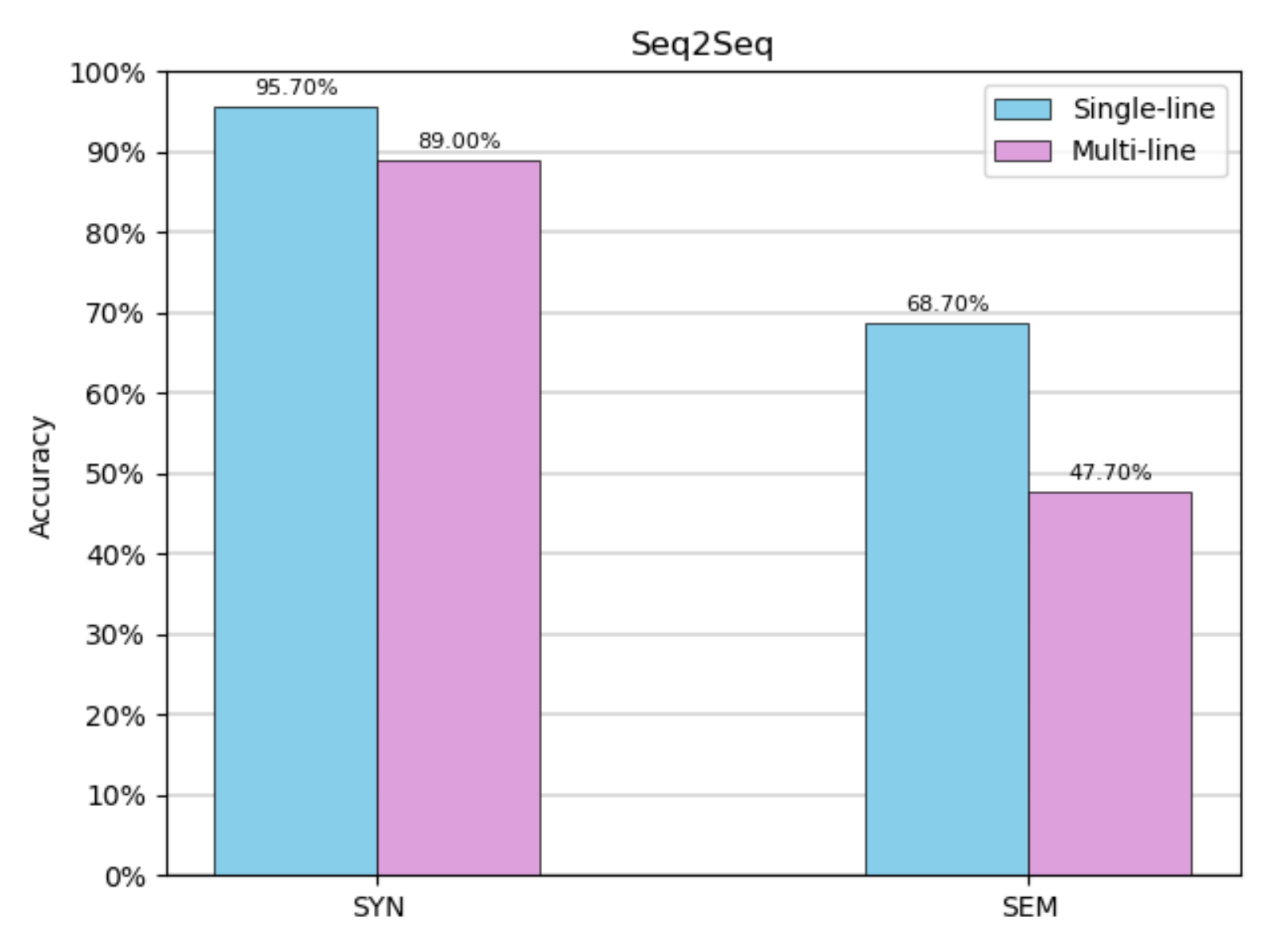}} \\
        
        \subfloat[Word Substitution]{\label{fig:codebert_word_substitution_63}
             \includegraphics[width=0.45\textwidth]{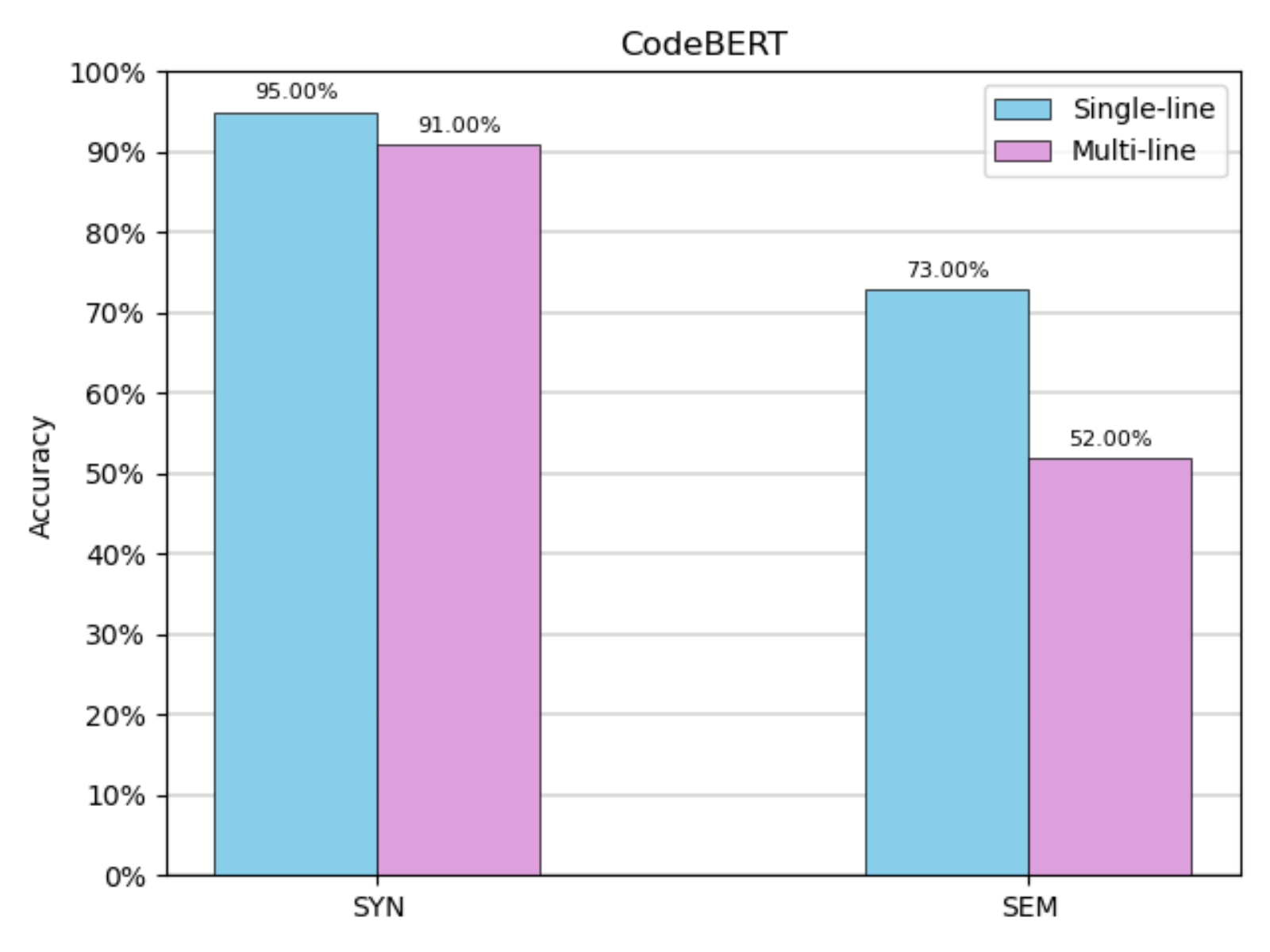}}
        \subfloat[Word Omission]{\label{fig:codebert_word_omission_63}
             \includegraphics[width=0.45\textwidth]{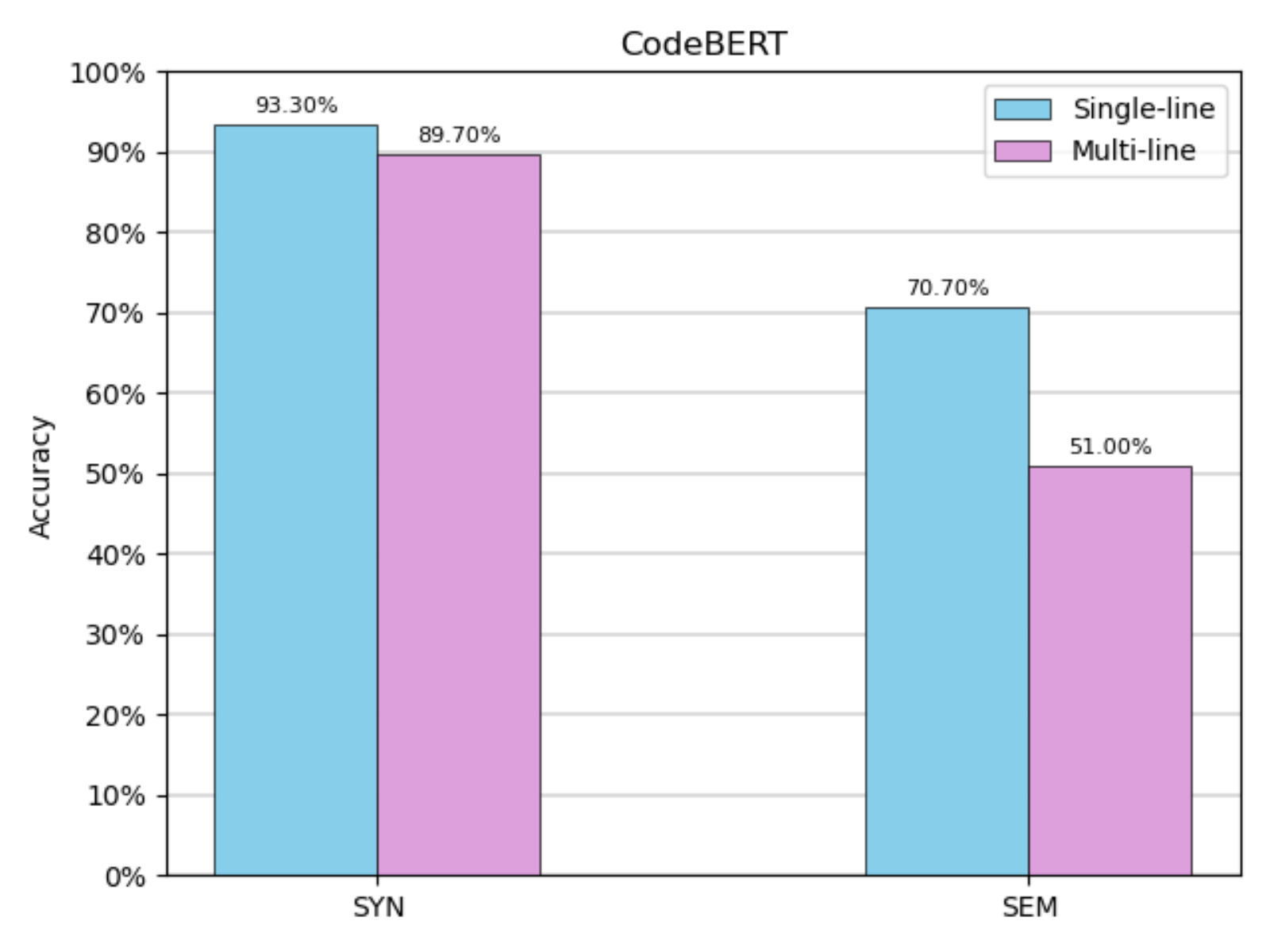}} \\
        
        \subfloat[Word Substitution]{\label{fig:codet5_word_substitution_63}
             \includegraphics[width=0.45\textwidth]{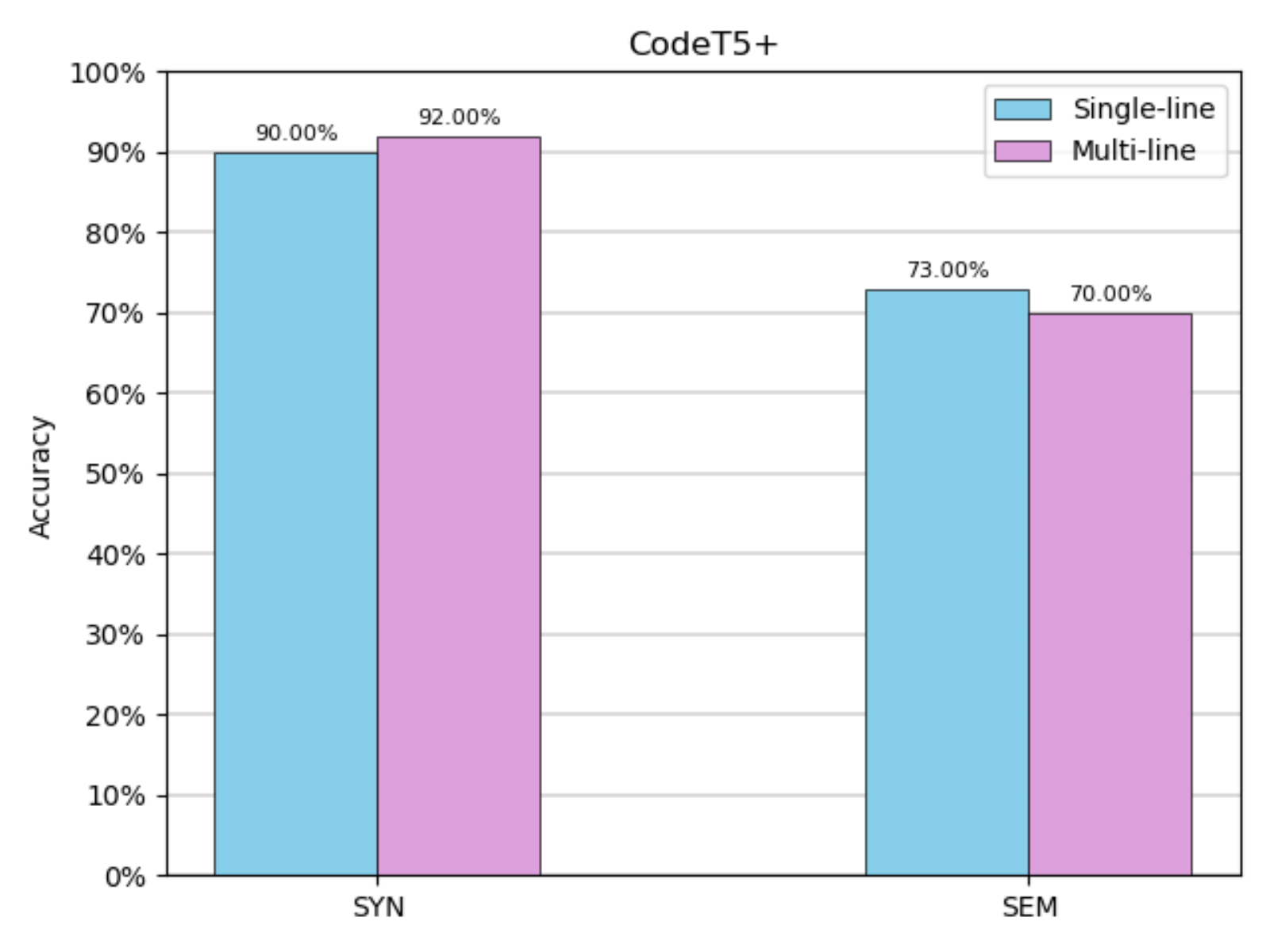}}
        \subfloat[Word Omission]{\label{fig:codet5_word_omission_63}
             \includegraphics[width=0.45\textwidth]{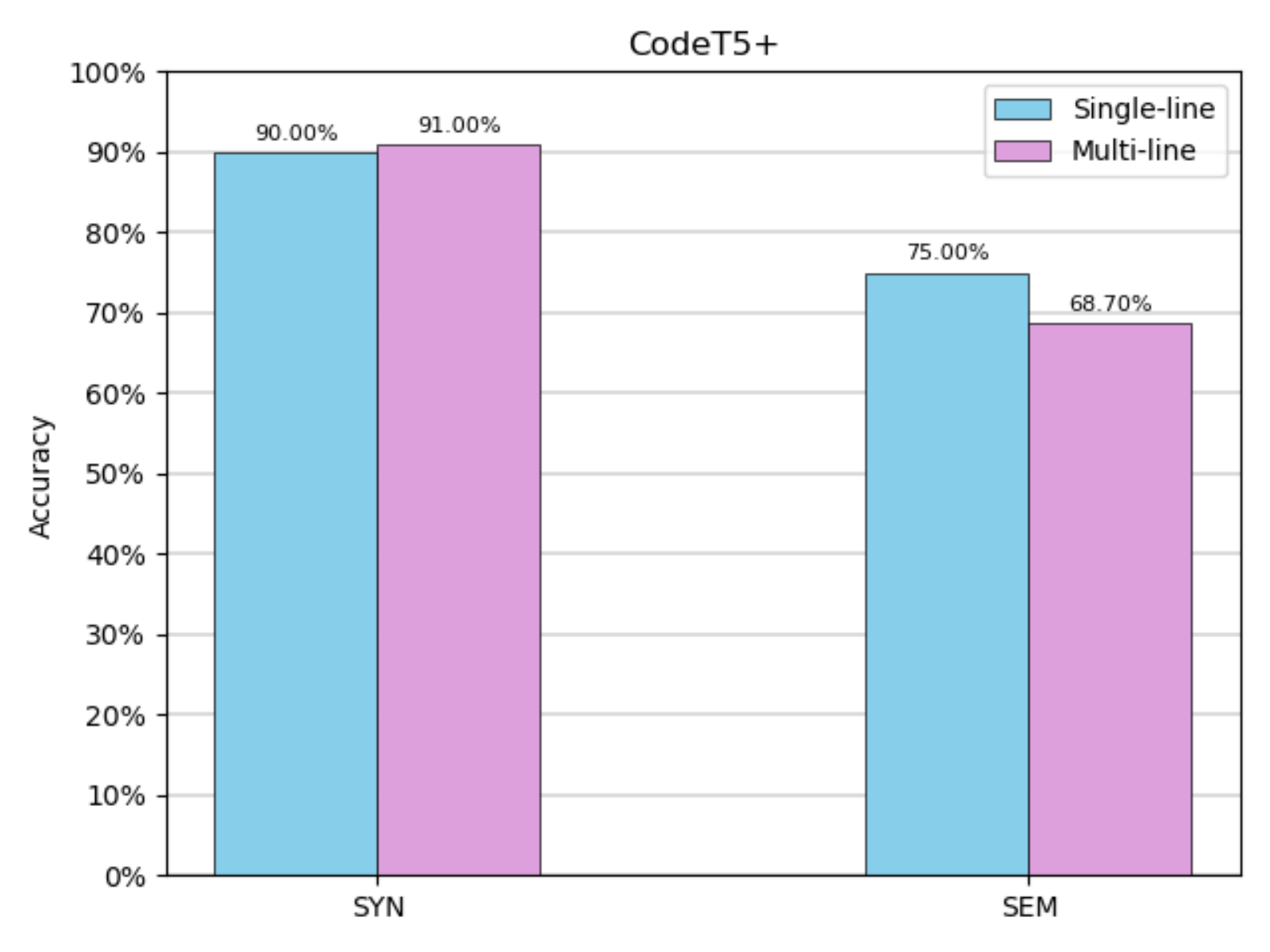}}

        \caption{Comparison between the models' performance on single-line code snippets \textit{vs.} multi-line code snippets when trained on partially augmented data (50\%) and tested on the original, clean data in terms of syntactic (SYN) and semantic (SEM) accuracy. Figures on the \textbf{left} show the performance when data is perturbed with \textit{word substitution}, while figures on the \textbf{right} show the performance when data is perturbed with \textit{word omission}.}
\label{fig:single_multi_63}
\end{figure}

\figurename~\ref{fig:clean_test} shows the results of the syntax and semantic evaluation for the three models. Again, we did not include robust accuracy in this analysis since the test set is not perturbed.
\figurename~\ref{fig:syntax_Test0} shows that the syntactic correctness of the models is pretty similar ($\geq 90\%$), regardless of the perturbation added in the training phase and the model. 
Notably, CodeT5+ provides an increment of the syntax accuracy by $1\%$ and $2\%$ when trained with word substitution and word omission, respectively.
For the semantic evaluation, \figurename~\ref{fig:semantics_Test0} shows that the usage of word substitution in the training phase provides a performance that is superior to the one achieved without perturbations for the Seq2Seq and CodeT5+ ($+4.6\%$ and $+3.1\%$, respectively). The usage of word omission in half of the training data boosts the performance of CodeT5+ ($+4.0\%$), while the other models decrease their semantic accuracy up to $\sim 2\%$.

As for the comparison between the models' performance on single-line versus multi-line snippets, \figurename{}~\ref{fig:single_multi_63} shows that when trained on partially augmented data (50\%) and tested on the original clean test set, for Seq2Seq and CodeBERT syntactic accuracy remains high across different perturbations, with single-line descriptions consistently performing slightly better than multi-line descriptions. Conversely, semantic accuracy on single-line snippets is significantly higher for both perturbation types.
To summarize, the overall trend suggests that these models generate syntactically correct snippets regardless of their length, whereas are more capable of generating code semantically equivalent to the ground truth when dealing with shorter snippets, regardless of the perturbation type.
On the other hand, a larger pre-trained model like CodeT5+ shows once again to benefit more from data augmentation than other models, becoming able to generate more complex multi-line snippets almost as correctly as single-line snippets.

\tablename~\ref{tab:qualitative_nonpert} shows two worth noticing examples in which, despite the non-perturbed test set, the models failed the prediction when trained with the original, non-perturbed training set, but returned syntactically and semantically correct predictions when trained with perturbed code descriptions.
The first example of the table shows that the use of word substitution in 50\% of the training set helped the model in predicting the right order of the instructions. In the second example, despite the model identifying the correct operands and their ordering, we found that the word \textit{move} confused the model since this verb is mostly used to transfer contents (e.g., from a register into another one). The removal of words in the training set helped the model to emphasize also other parts of the intent, such as the word \textit{right}. Therefore, after the data augmentation, the model correctly interpreted the \textit{move right} by generating the shift right operation (\texttt{shr} instruction).

\begin{table}[t]
\centering
\caption{Illustrative examples of predictions with and without data augmentation (DA) on the original, non-perturbed test set. The prediction errors are \textcolor{red}{\textbf{red}}.}
\label{tab:qualitative_nonpert}
\footnotesize
\begin{tabular}{  
>{\centering\arraybackslash}m{4cm} |
>{\centering\arraybackslash}m{3cm} |
>{\centering\arraybackslash}m{3cm} }
\toprule
\textbf{Original Intent}  & \textbf{Prediction w/o DA} & \textbf{Prediction with DA} \\ \midrule
\textit{push eax and edx on the stack} &
\texttt{push \textcolor{red}{edx} \textbackslash{n} push \textcolor{red}{eax}} &
\texttt{push eax \textbackslash{n} push edx}\\\midrule
\textit{move the bits of ax to the right by a number equal to the content of cl} &
\texttt{\textcolor{red}{mov} ax, cl} &
\texttt{shr ax, cl}\\ 
\bottomrule
\end{tabular}
\end{table}

\begin{mybox}{\parbox{10.5cm}{RQ3: Can data augmentation increase the performance of the models against non-perturbed code descriptions?}}
An important takeaway from our experiments is that the use of perturbations in the training data can increase the performance of the models in the code generation also against the original, non-perturbed code descriptions.
Indeed, although the syntactic correctness is almost the same for all models (with an improvement up to $1.8\%$), we found that the semantic correctness increased up to $4.6\%$ when using word substitution and up to $4.0\%$ when using word omission. In particular, a newer, pre-trained model such as CodeT5+,  benefits more from descriptions that are shorter or contain more variability.
Therefore, even if the test set is not perturbed, the models can improve the semantic correctness of the predictions by learning from the higher variability in the training set due to the use of perturbations. This is an effective solution to overcome a fixed structure of the NL intents derived from a single description style of the code used to fine-tune the models.

Overall, when fine-tuned on partially perturbed and then tested on the original data, Seq2Seq and CodeBERT show a behavior comparable to the baseline, proving a trade-off between robustness and performance, but their inability to fully leverage the benefits of data augmentation, due to the little (or zero) data seen during pre-training. On the other hand, models like CodeT5+, which are pre-trained on extensive datasets, can gain greater advantages from data augmentation. Their enhanced pre-training allows them to better generalize from the augmented data, improving their performance on both clean and perturbed data.
In conclusion, synonym substitution proves to be an effective data augmentation strategy, being the preferred solution to improve the models' ability to generate code when tested both on perturbed and non-perturbed NL descriptions.
\end{mybox}

\section{Threats to Validity}
\label{sec:threats}
\subsection{Construct Validity}


\noindent
\textbf{Exclusion of Closed Models:}  Regarding the closed models, the design choice of not including these models at this stage of our research was driven by three primary motivations. First, current research on offensive code generation typically relies on fine-tuning open-source models to enhance their abilities to generate shellcodes~\cite{yang2022dualsc,yang2023exploitgen,ruan2023prompt}. This approach allows for more granular control over the model's training process and the ability to tailor it specifically to the task at hand. Second, one of the primary goals of our work was to thoroughly assess the effectiveness of the proposed data augmentation strategy. This necessitated a fine-tuning approach, which is not feasible with closed models in a few-shot learning setting. Few-shot learning, while powerful, does not allow for a fair comparison with fine-tuning on an increasingly perturbed dataset, which was crucial for our study. Third, since both attackers and defenders need to avoid leaking their techniques and tactics to their counterparts (OPSEC), we consider the case of an attacker or defender who builds her own AI code generator by fine-tuning a model on a dataset of offensive code.
Using few-shot learning on closed models is a promising avenue for future research. It would enable us to evaluate the robustness of public models in the challenging domain of exploit generation, offering a broader perspective on their capabilities and limitations.

\noindent
\textbf{Use of Open-Source Models:} Since the goal of our investigation is to not only assess the robustness of AI code generators but also to thoroughly evaluate the effectiveness of our data augmentation strategy, our analysis requires the fine-tuning of the targeted AI models on the new, perturbed data. 
Therefore, we fine-tuned three open-source models, representative of the state-of-the-art, on a corpus of offensive code.
We plan to extend our assessment with other popular open-source LLMs, such as CodeLlama. Indeed, while fine-tuning open-source LLMs such as CodeLlama is extremely costly in terms of time and computing resources, it is an interesting path to explore for future work. 
However, we have also to consider that, although manipulable through prompt engineering, licenses for such models are designed to prevent misuse, including the generation of malicious code, and to promote the ethical use of AI technology.

\noindent
\textbf{Choice of Perturbation Types:} The choice of perturbation types is crucial for construct validity as it influences whether the perturbations effectively represent the variability in natural language descriptions.
The exploration of perturbation types is an integral consideration in assessing the robustness of AI-based code generators. Incorporating a more extensive range of perturbations would undoubtedly offer a more comprehensive understanding of the model's responsiveness to the variability of the NL. This aspect forms a crucial direction for future research. In Section~\ref{sec:methodology}, we detailed the rationale behind the perturbation types chosen, emphasizing their prevalence in practice. The use of word substitutions and omissions stems from their representative nature in capturing the challenges posed by the variability in NL descriptions. Finally, we remark that the usage of word-based perturbations in NL for code generation models has never been addressed before. Therefore, we consider the implementation of five different perturbations across two categories a valid means to establish a fundamental understanding before exploring a broader spectrum of perturbations in future work.

\subsection{Internal Validity}
\noindent
\textbf{Dataset Representativeness:} The dataset used in the study is vital for internal validity as it represents the basis for evaluating the robustness of AI-based solutions in generating software exploits. However, if the dataset is not comprehensive or representative of offensive code generation tasks, it may introduce bias, compromising the internal validity of the study's findings.
The dataset used for our experiments fits perfectly with the scope of this work since it is the largest collection of offensive code available to date for code generation. This manually curated dataset contains high-quality and detailed descriptions of code, that are often not available in larger corpora for code generation. Indeed, the dataset provides NL descriptions both at the block and statement levels that are closer to the descriptions needed by the models for complex programming tasks. This feature makes the dataset suitable for the injection of the perturbations in the code descriptions.

\noindent
\textbf{Realism of Shellcode Examples:} Regarding the simplicity and size of the examples in our dataset, it is important to note that shellcodes are typically concise sequences of assembly instructions designed to access low-level system components such as memory, stack, and registers. For instance, the lines of codes of shellcode programs used in previous work range between 16 and 46 (average is ~26)~\cite{liguori2021evil}. Consequently, the code descriptions are typically short as they provide detailed, expertly crafted explanations. Nevertheless, the dataset also includes 1,374 intents (approximately 23\% of the dataset) that generate multiple lines of assembly code, incorporating various assembly instructions, including complete functions.
We acknowledge that there is often an overlap between tokens in the natural language intent and the output code. This overlap is inherent to the nature of shellcode generation, where specific terminology and technical jargon are essential for accurately describing and implementing these low-level operations. This makes the introduction of perturbations such as word omission a valuable solution to make AI code generators less dependent on detailed NL descriptions.
The concern about the dataset's realism regarding how code exploits are found is valid. 
In practice, generating actual exploits often requires information from the targeted program, as not all potential exploits apply universally. However, creating a benchmark that includes such detailed contextual information is complex and resource-intensive. Although we acknowledge that the inclusion of this information can potentially lead to an improvement in performance, this investigation is not a specific goal of our work.
Our current dataset is composed of real-world shellcodes collected from reputable online exploit databases~\cite{shellstorm,exploitdb}. It represents the largest collection of assembly exploits annotated with natural language descriptions available to date. While compact, it offers a rich and authentic source of data for addressing the broader challenge of offensive code generation. To the best of our knowledge, there are no other comparable corpora in the context of offensive code generation from NL descriptions. As a matter of fact, our dataset is a widely used benchmark in the field of offensive code generation and is considered a high-quality corpora~\cite{yang2022dualsc,yang2023exploitgen}.
We also emphasize the critical importance of assessing and improving model robustness, especially given the lack of high-quality data for training AI models in this domain. Our proposed data augmentation strategy is tailored to enhance the model's ability to generalize from the available data, ensuring that it can maintain performance even when important information is missing or altered. 
In summary, while our dataset has limitations, it is a significant and realistic resource for the task at hand. Future work includes refining and expanding our approaches to further enhance the robustness and applicability of AI models in this challenging and crucial area.

\subsection{External Validity}
\noindent
\textbf{Generalization to Other Code Generation Scenarios:} Focusing solely on offensive code generation may limit the external validity of the study's findings, as they may not generalize to other code generation scenarios.
However, although offensive code is different from general-purpose ones in terms of programming languages and characteristics, the proposed method can be applied to different code generation scenarios. The decision to focus on offensive code generation is driven by the critical importance of robustness in this specific field. The use of models to generate offensive code, a research topic that is gaining increasing interest in software security, poses unique challenges, requiring models to be robust against NL variations while maintaining syntactic and semantic correctness. The few benchmark datasets for security tasks are limited in size and have detailed NL descriptions~\cite{liguori2022can,liguori2021evil}, hence constraining the generalization of results and making offensive code generation an ideal domain for our study. By focusing on exploits, the study addresses a specific and high-impact application of code generation, providing insights into the challenges for assessing and improving the robustness of models via data augmentation in scenarios with stringent requirements.

\noindent
\textbf{Generalization with Respect to Tokenization Process:}
In our study, we primarily focused on models using word-level tokenization to establish a consistent baseline and assess the effectiveness of our data augmentation strategy. Our perturbation process modifies the natural language description before tokenization, making it agnostic to the specific tokenizer used. This allows us to introduce variations in the input text, which are then captured by the model during training, hence, potentially generalizing to other tokenization methods, including subword-level tokenizers such as Byte Pair Encoding (BPE).
The reason for focusing on word-level tokenization in our experiments was primarily based on the availability and widespread use of pre-trained models that utilize this tokenization scheme. However, we acknowledge that exploring the applicability of our approach with different tokenizers, including BPE, could be an interesting avenue for future research.

\section{Limitations and Future Work}
\label{sec:limitation}
Despite the valuable insights gained from our study on the effects of linguistic perturbations on AI code generators, several limitations warrant discussion and pave the way for future research.

\vspace{0.1cm}
\noindent
\textbf{Controlled Data Augmentation Strategy:} Our methodology involved a controlled data augmentation strategy where we kept the dataset size constant. This approach allowed us to isolate the effects of different perturbation types on model performance without introducing confounding variables associated with dataset size changes. However, this also limits the scope of our findings concerning methodologies that involve increasing the training dataset size through data augmentation. Moreover, implementing this would require a significant reconfiguration of our experimental setup, including redoing all experiments under new conditions and substantially increasing the number of configurations and manual evaluations needed for assessing semantic correctness. Nevertheless, exploring the impact of expanding the training dataset with perturbed data remains an important direction for future work.

\vspace{0.1cm}
\noindent
\textbf{Fixed Ratio of Word Substitutions:} We set the word substitution ratio at 10\% of the words identified as suitable for substitution. This decision was based on balancing variability and sentence integrity, aligning with best practices and findings from prior work~\cite{wei2019eda}. Nonetheless, we acknowledge that experimenting with varying substitution ratios could offer deeper insights into the model's robustness to different levels of perturbation. Conducting such experiments would significantly expand the number of experimental configurations and further strain the manual evaluation process, which already involves a substantial workload due to the need for assessing semantic correctness. Future research could explore the effects of different substitution ratios, potentially utilizing automated evaluation methods to manage the increased complexity.

\vspace{0.1cm}
\noindent
\textbf{Combination of Perturbation Strategies:} In our study, we applied word substitution and word omission perturbations separately to emulate distinct real-world scenarios. Word substitution mimics using different but equivalent ways to describe a code snippet, while word omission represents missing or incomplete information in a code description. Both strategies were effective in increasing model robustness against perturbed test sets, with word substitution also improving performance on the original, non-perturbed test set. However, we did not investigate the combined effect of applying multiple perturbation types simultaneously. Assessing the impact of combined perturbations could provide a more comprehensive understanding of how models cope with complex linguistic variations. This presents a promising avenue for future research.

\section{Conclusion}
\label{sec:conclusion}

In this paper, we proposed a method to assess the robustness of AI offensive code generators through the injection of perturbations in the NL code descriptions.
We first showed that the perturbed descriptions preserve the semantics of the original ones.
Then, we applied the method to assess three state-of-the-art models in the automatic generation of offensive assembly code for software security exploits starting from the English language.
Our experiments pointed out that the models are not robust when the intents deviate from the ones used in the corpora.
In particular, the performance of the models drastically drops when developers do not explicitly specify all the information in the NL intents (\textit{word omission}).
To enhance the robustness of the code generators, we applied the method to perform \textit{data augmentation}, i.e., to increase the variability of the code descriptions, showing that models increase the performance when at least $50\%$ of the training data is perturbed.
Finally, we found that the performance of the code generation task improves when the model is trained with a perturbed version of the code descriptions containing new words (\textit{word substitution}) also against the original, non-perturbed test-set.

For practitioners looking to enhance model robustness, we recommend employing both word substitution and word omission perturbations in multiple rounds of data augmentation. If resources are constrained, prioritizing word substitution is advisable, as it has demonstrated benefits for both robustness and overall performance.

\section*{Data and/or Code availability}

The dataset, experimental results, and code to replicate the experiments are available at the following URL: \url{https://github.com/dessertlab/Robustness-of-AI-Offensive-Code-Generators/}.

\section*{Funding}

The authors have no relevant financial or non-financial interests to disclose.

\section*{Declarations}

\textbf{Conflicts of interests} The authors have no conflicts of interest to declare that are relevant to the content of this article.
All authors certify that they have no affiliations with or involvement in any organization or entity with any financial interest or non-financial interest in the subject matter or materials discussed in this manuscript.
The authors have no financial or proprietary interests in any material discussed in this article.

\begin{acknowledgements}
This work has been partially supported by the MUR PRIN 2022 program, project \textit{FLEGREA}, CUP E53D23007950001, the \textit{IDA—Information Disorder Awareness} Project funded by the European Union-Next Generation EU within the SERICS Program through the MUR National Recovery and Resilience Plan under Grant PE00000014, the \textit{SERENA-IIoT} project funded by MUR (Ministero dell’Università e della Ricerca) and European Union (Next Generation EU) under the PRIN 2022 program (project code 2022CN4EBH) and the \textit{GENIO} project (CUP B69J23005770005) funded by MIMIT.

We are grateful to the DESSERT Research Group at the University of Naples Federico II, Italy, the students of the Italian training program CyberChallenge.IT 2024, team Naples, and anyone who participated in our survey.
\end{acknowledgements}

%
%

\bibliographystyle{spmpsci}      
\bibliography{mybibfile}   

\begin{thebibliography}{10}
\providecommand{\url}[1]{{#1}}
\providecommand{\urlprefix}{URL }
\expandafter\ifx\csname urlstyle\endcsname\relax
  \providecommand{\doi}[1]{DOI~\discretionary{}{}{}#1}\else
  \providecommand{\doi}{DOI~\discretionary{}{}{}\begingroup \urlstyle{rm}\Url}\fi

\bibitem{ahmed2024automatic}
Ahmed, T., Pai, K.S., Devanbu, P., Barr, E.: Automatic semantic augmentation of language model prompts (for code summarization).
\newblock In: Proceedings of the IEEE/ACM 46th International Conference on Software Engineering, pp. 1--13 (2024)

\bibitem{akbik2018coling}
Akbik, A., Blythe, D., Vollgraf, R.: Contextual string embeddings for sequence labeling.
\newblock In: Proceedings of the 27th International Conference on Computational Linguistics, {COLING} 2018, Santa Fe, New Mexico, USA, August 20-26, 2018, pp. 1638--1649. Association for Computational Linguistics (2018).
\newblock \urlprefix\url{https://aclanthology.org/C18-1139/}

\bibitem{alzantotSEHSC18}
Alzantot, M., Sharma, Y., Elgohary, A., Ho, B., Srivastava, M.B., Chang, K.: Generating natural language adversarial examples.
\newblock In: E.~Riloff, D.~Chiang, J.~Hockenmaier, J.~Tsujii (eds.) Proceedings of the 2018 Conference on Empirical Methods in Natural Language Processing, Brussels, Belgium, October 31 - November 4, 2018, pp. 2890--2896. Association for Computational Linguistics (2018).
\newblock \doi{10.18653/v1/d18-1316}.
\newblock \urlprefix\url{https://doi.org/10.18653/v1/d18-1316}

\bibitem{avgerinos2014automatic}
Avgerinos, T., Cha, S.K., Rebert, A., Schwartz, E.J., Woo, M., Brumley, D.: Automatic exploit generation.
\newblock Communications of the ACM \textbf{57}(2), 74--84 (2014)

\bibitem{bahdanau2014neural}
Bahdanau, D., Cho, K., Bengio, Y.: Neural machine translation by jointly learning to align and translate.
\newblock In: 3rd International Conference on Learning Representations, {ICLR} 2015, San Diego, CA, USA, May 7-9, 2015, Conference Track Proceedings (2015).
\newblock \urlprefix\url{http://arxiv.org/abs/1409.0473}

\bibitem{belinkov2018synthetic}
Belinkov, Y., Bisk, Y.: Synthetic and natural noise both break neural machine translation.
\newblock In: 6th International Conference on Learning Representations, {ICLR} 2018, Vancouver, BC, Canada, April 30 - May 3, 2018, Conference Track Proceedings. OpenReview.net (2018).
\newblock \urlprefix\url{https://openreview.net/forum?id=BJ8vJebC-}

\bibitem{bird2006nltk}
Bird, S.: Nltk: the natural language toolkit.
\newblock In: Proceedings of the COLING/ACL 2006 Interactive Presentation Sessions, pp. 69--72 (2006)

\bibitem{botacin2023gpthreats}
Botacin, M.: Gpthreats-3: Is automatic malware generation a threat?
\newblock In: 2023 IEEE Security and Privacy Workshops (SPW), pp. 238--254. IEEE (2023)

\bibitem{boucher_2022_badchars}
Boucher, N., Shumailov, I., Anderson, R., Papernot, N.: Bad characters: Imperceptible nlp attacks.
\newblock In: 2022 IEEE Symposium on Security and Privacy (SP), pp. 1987--2004 (2022).
\newblock \doi{10.1109/SP46214.2022.9833641}

\bibitem{cheng2020seq2sick}
Cheng, M., Yi, J., Chen, P.Y., Zhang, H., Hsieh, C.J.: Seq2sick: Evaluating the robustness of sequence-to-sequence models with adversarial examples.
\newblock In: Proceedings of the AAAI conference on artificial intelligence, vol.~34, pp. 3601--3608 (2020)

\bibitem{chung2004identifying}
Chung, T.M., Nation, P.: Identifying technical vocabulary.
\newblock System \textbf{32}(2), 251--263 (2004)

\bibitem{COTRONEO2024112113}
Cotroneo, D., Foggia, A., Improta, C., Liguori, P., Natella, R.: Automating the correctness assessment of ai-generated code for security contexts.
\newblock Journal of Systems and Software \textbf{216}, 112113 (2024).
\newblock \doi{https://doi.org/10.1016/j.jss.2024.112113}.
\newblock \urlprefix\url{https://www.sciencedirect.com/science/article/pii/S0164121224001584}

\bibitem{exploitdb}
Exploit-db: {Exploit Database Shellcodes}.
\newblock \url{https://www.exploit-db.com/shellcodes?platform=linux\_x86/} (2023)

\bibitem{feng-etal-2021-survey}
Feng, S.Y., Gangal, V., Wei, J., Chandar, S., Vosoughi, S., Mitamura, T., Hovy, E.: A survey of data augmentation approaches for {NLP}.
\newblock In: C.~Zong, F.~Xia, W.~Li, R.~Navigli (eds.) Findings of the Association for Computational Linguistics: ACL-IJCNLP 2021, pp. 968--988. Association for Computational Linguistics, Online (2021).
\newblock \doi{10.18653/v1/2021.findings-acl.84}.
\newblock \urlprefix\url{https://aclanthology.org/2021.findings-acl.84}

\bibitem{feng2020codebert}
Feng, Z., Guo, D., Tang, D., Duan, N., Feng, X., Gong, M., Shou, L., Qin, B., Liu, T., Jiang, D., Zhou, M.: Codebert: {A} pre-trained model for programming and natural languages.
\newblock In: Findings of the Association for Computational Linguistics: {EMNLP} 2020, Online Event, 16-20 November 2020, \emph{Findings of {ACL}}, vol. {EMNLP} 2020, pp. 1536--1547. Association for Computational Linguistics (2020).
\newblock \doi{10.18653/v1/2020.findings-emnlp.139}.
\newblock \urlprefix\url{https://doi.org/10.18653/v1/2020.findings-emnlp.139}

\bibitem{foster2005sockets}
Foster, J.: Sockets, Shellcode, Porting, and Coding: Reverse Engineering Exploits and Tool Coding for Security Professionals.
\newblock Elsevier Science (2005).
\newblock \urlprefix\url{https://books.google.it/books?id=ZNI5dvBSfZoC}

\bibitem{gao2019soft}
Gao, F., Zhu, J., Wu, L., Xia, Y., Qin, T., Cheng, X., Zhou, W., Liu, T.Y.: Soft contextual data augmentation for neural machine translation.
\newblock In: Proceedings of the 57th Annual Meeting of the Association for Computational Linguistics, pp. 5539--5544 (2019)

\bibitem{DBLP:conf/icse/GaoZGW23}
Gao, S., Zhang, H., Gao, C., Wang, C.: Keeping pace with ever-increasing data: Towards continual learning of code intelligence models.
\newblock In: 45th {IEEE/ACM} International Conference on Software Engineering, {ICSE} 2023, Melbourne, Australia, May 14-20, 2023, pp. 30--42. {IEEE} (2023).
\newblock \doi{10.1109/ICSE48619.2023.00015}.
\newblock \urlprefix\url{https://doi.org/10.1109/ICSE48619.2023.00015}

\bibitem{gupta2023chatgpt}
Gupta, M., Akiri, C., Aryal, K., Parker, E., Praharaj, L.: From chatgpt to threatgpt: Impact of generative ai in cybersecurity and privacy.
\newblock IEEE Access  (2023)

\bibitem{han-etal-2021-translation}
Han, L., Smeaton, A., Jones, G.: Translation quality assessment: A brief survey on manual and automatic methods.
\newblock In: Proceedings for the First Workshop on Modelling Translation: Translatology in the Digital Age, pp. 15--33. Association for Computational Linguistics, online (2021).
\newblock \urlprefix\url{https://aclanthology.org/2021.motra-1.3}

\bibitem{DBLP:journals/corr/abs-2109-07403}
Hauser, J., Meng, Z., Pascual, D., Wattenhofer, R.: {BERT} is robust! {A} case against synonym-based adversarial examples in text classification.
\newblock CoRR \textbf{abs/2109.07403} (2021).
\newblock \urlprefix\url{https://arxiv.org/abs/2109.07403}

\bibitem{heigold-etal-2018-robust}
Heigold, G., Varanasi, S., Neumann, G., van Genabith, J.: How robust are character-based word embeddings in tagging and {MT} against wrod scramlbing or randdm nouse?
\newblock In: Proceedings of the 13th Conference of the Association for Machine Translation in the {A}mericas (Volume 1: Research Track), pp. 68--80. Association for Machine Translation in the Americas, Boston, MA (2018).
\newblock \urlprefix\url{https://aclanthology.org/W18-1807}

\bibitem{henkel2022semantic}
Henkel, J., Ramakrishnan, G., Wang, Z., Albarghouthi, A., Jha, S., Reps, T.: Semantic robustness of models of source code.
\newblock In: 2022 IEEE International Conference on Software Analysis, Evolution and Reengineering (SANER), pp. 526--537. IEEE (2022)

\bibitem{huang2021robustness}
Huang, S., Li, Z., Qu, L., Pan, L.: On robustness of neural semantic parsers.
\newblock In: Proceedings of the 16th Conference of the European Chapter of the Association for Computational Linguistics: Main Volume, {EACL} 2021, Online, April 19 - 23, 2021, pp. 3333--3342. Association for Computational Linguistics (2021).
\newblock \urlprefix\url{https://aclanthology.org/2021.eacl-main.292/}

\bibitem{jha2022codeattack}
Jha, A., Reddy, C.K.: Codeattack: Code-based adversarial attacks for pre-trained programming language models.
\newblock arXiv preprint arXiv:2206.00052  (2022)

\bibitem{jin2020bert}
Jin, D., Jin, Z., Zhou, J.T., Szolovits, P.: Is bert really robust? a strong baseline for natural language attack on text classification and entailment.
\newblock In: Proceedings of the AAAI conference on artificial intelligence, vol.~34, pp. 8018--8025 (2020)

\bibitem{kachru2009handbook}
Kachru, B.B., Kachru, Y., Nelson, C.L.: The handbook of world Englishes, vol.~48.
\newblock John Wiley \& Sons (2009)

\bibitem{kim2018artificial}
Kim, D., MacKinnon, T.: Artificial intelligence in fracture detection: transfer learning from deep convolutional neural networks.
\newblock Clinical radiology \textbf{73}(5), 439--445 (2018)

\bibitem{kingma2015adam}
Kingma, D.P., Ba, J.: Adam: {A} method for stochastic optimization.
\newblock In: 3rd International Conference on Learning Representations, {ICLR} 2015, San Diego, CA, USA, May 7-9, 2015, Conference Track Proceedings (2015).
\newblock \urlprefix\url{http://arxiv.org/abs/1412.6980}

\bibitem{li2018textbugger}
Li, J., Ji, S., Du, T., Li, B., Wang, T.: Textbugger: Generating adversarial text against real-world applications.
\newblock In: 26th Annual Network and Distributed System Security Symposium, {NDSS} 2019, San Diego, California, USA, February 24-27, 2019. The Internet Society (2019).
\newblock \urlprefix\url{https://www.ndss-symposium.org/ndss-paper/textbugger-generating-adversarial-text-against-real-world-applications/}

\bibitem{li2018named}
Li, Z., Wang, X., Aw, A., Chng, E.S., Li, H.: Named-entity tagging and domain adaptation for better customized translation.
\newblock In: Proceedings of the seventh named entities workshop, pp. 41--46 (2018)

\bibitem{li-etal-2021-searching}
Li, Z., Xu, J., Zeng, J., Li, L., Zheng, X., Zhang, Q., Chang, K.W., Hsieh, C.J.: Searching for an effective defender: Benchmarking defense against adversarial word substitution.
\newblock In: Proceedings of the 2021 Conference on Empirical Methods in Natural Language Processing, pp. 3137--3147. Association for Computational Linguistics, Online and Punta Cana, Dominican Republic (2021).
\newblock \doi{10.18653/v1/2021.emnlp-main.251}.
\newblock \urlprefix\url{https://aclanthology.org/2021.emnlp-main.251}

\bibitem{liguori2021shellcode_ia32}
Liguori, P., Al-Hossami, E., Cotroneo, D., Natella, R., Cukic, B., Shaikh, S.: {S}hellcode{\_}{IA}32: A dataset for automatic shellcode generation.
\newblock In: Proceedings of the 1st Workshop on Natural Language Processing for Programming (NLP4Prog 2021), pp. 58--64. Association for Computational Linguistics, Online (2021).
\newblock \doi{10.18653/v1/2021.nlp4prog-1.7}.
\newblock \urlprefix\url{https://aclanthology.org/2021.nlp4prog-1.7}

\bibitem{liguori2022can}
Liguori, P., Al-Hossami, E., Cotroneo, D., Natella, R., Cukic, B., Shaikh, S.: Can we generate shellcodes via natural language? an empirical study.
\newblock Automated Software Engineering \textbf{29}(1), 30 (2022).
\newblock \doi{10.1007/s10515-022-00331-3}.
\newblock \urlprefix\url{https://doi.org/10.1007/s10515-022-00331-3}

\bibitem{liguori2021evil}
Liguori, P., Al-Hossami, E., Orbinato, V., Natella, R., Shaikh, S., Cotroneo, D., Cukic, B.: Evil: Exploiting software via natural language.
\newblock In: 2021 IEEE 32nd International Symposium on Software Reliability Engineering (ISSRE), pp. 321--332 (2021).
\newblock \doi{10.1109/ISSRE52982.2021.00042}

\bibitem{liguori2023evaluates}
Liguori, P., Improta, C., Natella, R., Cukic, B., Cotroneo, D.: Who evaluates the evaluators? on automatic metrics for assessing ai-based offensive code generators.
\newblock Expert Systems with Applications \textbf{225}, 120073 (2023)

\bibitem{liu2019technical}
Liu, D., Lei, L.: Technical vocabulary.
\newblock In: The Routledge handbook of vocabulary studies, pp. 111--124. Routledge (2019)

\bibitem{DBLP:journals/corr/abs-1907-11692}
Liu, Y., Ott, M., Goyal, N., Du, J., Joshi, M., Chen, D., Levy, O., Lewis, M., Zettlemoyer, L., Stoyanov, V.: Roberta: {A} robustly optimized {BERT} pretraining approach.
\newblock CoRR \textbf{abs/1907.11692} (2019).
\newblock \urlprefix\url{http://arxiv.org/abs/1907.11692}

\bibitem{lutellier2020coconut}
Lutellier, T., Pham, H.V., Pang, L., Li, Y., Wei, M., Tan, L.: Coconut: combining context-aware neural translation models using ensemble for program repair.
\newblock In: Proceedings of the 29th ACM SIGSOFT international symposium on software testing and analysis, pp. 101--114 (2020)

\bibitem{DBLP:conf/msr/MashhadiH21}
Mashhadi, E., Hemmati, H.: Applying codebert for automated program repair of java simple bugs.
\newblock In: 18th {IEEE/ACM} International Conference on Mining Software Repositories, {MSR} 2021, Madrid, Spain, May 17-19, 2021, pp. 505--509. {IEEE} (2021).
\newblock \doi{10.1109/MSR52588.2021.00063}.
\newblock \urlprefix\url{https://doi.org/10.1109/MSR52588.2021.00063}

\bibitem{DBLP:conf/icse/MastropaoloPGCSOB23}
Mastropaolo, A., Pascarella, L., Guglielmi, E., Ciniselli, M., Scalabrino, S., Oliveto, R., Bavota, G.: On the robustness of code generation techniques: An empirical study on github copilot.
\newblock In: 45th {IEEE/ACM} International Conference on Software Engineering, {ICSE} 2023, Melbourne, Australia, May 14-20, 2023, pp. 2149--2160. {IEEE} (2023).
\newblock \doi{10.1109/ICSE48619.2023.00181}.
\newblock \urlprefix\url{https://doi.org/10.1109/ICSE48619.2023.00181}

\bibitem{mastropaolo2021studying}
Mastropaolo, A., Scalabrino, S., Cooper, N., Palacio, D.N., Poshyvanyk, D., Oliveto, R., Bavota, G.: Studying the usage of text-to-text transfer transformer to support code-related tasks.
\newblock In: 2021 IEEE/ACM 43rd International Conference on Software Engineering (ICSE), pp. 336--347. IEEE (2021)

\bibitem{DBLP:journals/corr/abs-2011-03901}
Mathai, A., Khare, S., Tamilselvam, S., Mani, S.: Adversarial black-box attacks on text classifiers using multi-objective genetic optimization guided by deep networks.
\newblock CoRR \textbf{abs/2011.03901} (2020).
\newblock \urlprefix\url{https://arxiv.org/abs/2011.03901}

\bibitem{megahed2018penetration}
Megahed, H.: Penetration Testing with Shellcode: Detect, exploit, and secure network-level and operating system vulnerabilities.
\newblock Packt Publishing (2018)

\bibitem{michel2019evaluation}
Michel, P., Li, X., Neubig, G., Pino, J.M.: On evaluation of adversarial perturbations for sequence-to-sequence models.
\newblock In: Proceedings of the 2019 Conference of the North American Chapter of the Association for Computational Linguistics: Human Language Technologies, {NAACL-HLT} 2019, Minneapolis, MN, USA, June 2-7, 2019, Volume 1 (Long and Short Papers), pp. 3103--3114. Association for Computational Linguistics (2019).
\newblock \doi{10.18653/v1/n19-1314}.
\newblock \urlprefix\url{https://doi.org/10.18653/v1/n19-1314}

\bibitem{modrzejewski2020incorporating}
Modrzejewski, M., Exel, M., Buschbeck, B., Ha, T.L., Waibel, A.: Incorporating external annotation to improve named entity translation in nmt.
\newblock In: Proceedings of the 22nd Annual Conference of the European Association for Machine Translation, pp. 45--51 (2020)

\bibitem{morris2020reevaluating}
Morris, J.X., Lifland, E., Lanchantin, J., Ji, Y., Qi, Y.: Reevaluating adversarial examples in natural language.
\newblock arXiv preprint arXiv:2004.14174  (2020)

\bibitem{morris2020textattack}
Morris, J.X., Lifland, E., Yoo, J.Y., Grigsby, J., Jin, D., Qi, Y.: Textattack: {A} framework for adversarial attacks, data augmentation, and adversarial training in {NLP}.
\newblock In: Proceedings of the 2020 Conference on Empirical Methods in Natural Language Processing: System Demonstrations, {EMNLP} 2020 - Demos, Online, November 16-20, 2020, pp. 119--126. Association for Computational Linguistics (2020).
\newblock \doi{10.18653/v1/2020.emnlp-demos.16}.
\newblock \urlprefix\url{https://doi.org/10.18653/v1/2020.emnlp-demos.16}

\bibitem{mrksic:2016:naacl}
Mrk\v{s}i\'c, N., {\'O S\'eaghdha}, D., Thomson, B., Ga\v{s}i\'c, M., Rojas-Barahona, L., Su, P.H., Vandyke, D., Wen, T.H., Young, S.: Counter-fitting word vectors to linguistic constraints.
\newblock In: Proceedings of HLT-NAACL (2016)

\bibitem{neubig18xnmt}
Neubig, G., Sperber, M., Wang, X., Felix, M., Matthews, A., Padmanabhan, S., Qi, Y., Sachan, D.S., Arthur, P., Godard, P., Hewitt, J., Riad, R., Wang, L.: {XNMT: The eXtensible Neural Machine Translation Toolkit}.
\newblock In: Conference of the Association for Machine Translation in the Americas (AMTA) Open Source Software Showcase. Boston, USA (2018).
\newblock \urlprefix\url{https://arxiv.org/pdf/1803.00188.pdf}

\bibitem{nguyen2020data}
Nguyen, X.P., Joty, S., Wu, K., Aw, A.T.: Data diversification: A simple strategy for neural machine translation.
\newblock Advances in Neural Information Processing Systems \textbf{33}, 10018--10029 (2020)

\bibitem{pa2023attacker}
Pa~Pa, Y.M., Tanizaki, S., Kou, T., Van~Eeten, M., Yoshioka, K., Matsumoto, T.: An attacker’s dream? exploring the capabilities of chatgpt for developing malware.
\newblock In: Proceedings of the 16th Cyber Security Experimentation and Test Workshop, pp. 10--18 (2023)

\bibitem{panichella2012mining}
Panichella, S., Aponte, J., Di~Penta, M., Marcus, A., Canfora, G.: Mining source code descriptions from developer communications.
\newblock In: 2012 20th IEEE International Conference on Program Comprehension (ICPC), pp. 63--72. IEEE (2012)

\bibitem{tokenize}
{Python}: {tokenize} (2023).
\newblock \urlprefix\url{https://docs.python.org/3/library/tokenize.html}

\bibitem{DBLP:journals/jmlr/RaffelSRLNMZLL20}
Raffel, C., Shazeer, N., Roberts, A., Lee, K., Narang, S., Matena, M., Zhou, Y., Li, W., Liu, P.J.: Exploring the limits of transfer learning with a unified text-to-text transformer.
\newblock J. Mach. Learn. Res. \textbf{21}, 140:1--140:67 (2020).
\newblock \urlprefix\url{http://jmlr.org/papers/v21/20-074.html}

\bibitem{st}
Reimers, N.: Sentencetransformers documentation.
\newblock \url{https://www.sbert.net/} (2022)

\bibitem{reimers2019sentence}
Reimers, N., Gurevych, I.: Sentence-bert: Sentence embeddings using siamese bert-networks.
\newblock In: K.~Inui, J.~Jiang, V.~Ng, X.~Wan (eds.) Proceedings of the 2019 Conference on Empirical Methods in Natural Language Processing and the 9th International Joint Conference on Natural Language Processing, {EMNLP-IJCNLP} 2019, Hong Kong, China, November 3-7, 2019, pp. 3980--3990. Association for Computational Linguistics (2019).
\newblock \doi{10.18653/v1/D19-1410}.
\newblock \urlprefix\url{https://doi.org/10.18653/v1/D19-1410}

\bibitem{DBLP:conf/emnlp/ReimersG20}
Reimers, N., Gurevych, I.: Making monolingual sentence embeddings multilingual using knowledge distillation.
\newblock In: B.~Webber, T.~Cohn, Y.~He, Y.~Liu (eds.) Proceedings of the 2020 Conference on Empirical Methods in Natural Language Processing, {EMNLP} 2020, Online, November 16-20, 2020, pp. 4512--4525. Association for Computational Linguistics (2020).
\newblock \doi{10.18653/v1/2020.emnlp-main.365}.
\newblock \urlprefix\url{https://doi.org/10.18653/v1/2020.emnlp-main.365}

\bibitem{ruan2023prompt}
Ruan, X., Yu, Y., Ma, W., Cai, B.: Prompt learning for developing software exploits.
\newblock In: Proceedings of the 14th Asia-Pacific Symposium on Internetware, pp. 154--164 (2023)

\bibitem{shellstorm}
Shell-storm: {Shellcodes database for study cases}.
\newblock \url{http://shell-storm.org/shellcode/} (2022)

\bibitem{1223656}
Silva, C., Ribeiro, B.: The importance of stop word removal on recall values in text categorization.
\newblock In: Proceedings of the International Joint Conference on Neural Networks, 2003., vol.~3, pp. 1661--1666 vol.3 (2003).
\newblock \doi{10.1109/IJCNN.2003.1223656}

\bibitem{spacy}
{spaCy}: {Industrial-Strength Natural Language Processing} (2023).
\newblock \urlprefix\url{https://spacy.io/}

\bibitem{tufano2019learning}
Tufano, M., Pantiuchina, J., Watson, C., Bavota, G., Poshyvanyk, D.: On learning meaningful code changes via neural machine translation.
\newblock In: 2019 IEEE/ACM 41st International Conference on Software Engineering (ICSE), pp. 25--36. IEEE (2019)

\bibitem{pops}
Tychonievich, L.: {Parts of Programming Speech}.
\newblock \url{https://luthert.web.illinois.edu/blog/posts/519.html} (2014)

\bibitem{vaswani2017attention}
Vaswani, A., Shazeer, N., Parmar, N., Uszkoreit, J., Jones, L., Gomez, A.N., Kaiser, {\L}., Polosukhin, I.: Attention is all you need.
\newblock In: Advances in neural information processing systems, pp. 5998--6008 (2017)

\bibitem{wang-etal-2023-recode}
Wang, S., Li, Z., Qian, H., Yang, C., Wang, Z., Shang, M., Kumar, V., Tan, S., Ray, B., Bhatia, P., Nallapati, R., Ramanathan, M.K., Roth, D., Xiang, B.: {R}e{C}ode: Robustness evaluation of code generation models.
\newblock In: A.~Rogers, J.~Boyd-Graber, N.~Okazaki (eds.) Proceedings of the 61st Annual Meeting of the Association for Computational Linguistics (Volume 1: Long Papers), pp. 13818--13843. Association for Computational Linguistics, Toronto, Canada (2023).
\newblock \doi{10.18653/v1/2023.acl-long.773}.
\newblock \urlprefix\url{https://aclanthology.org/2023.acl-long.773}

\bibitem{wang2018switchout}
Wang, X., Pham, H., Dai, Z., Neubig, G.: Switchout: an efficient data augmentation algorithm for neural machine translation.
\newblock In: Proceedings of the 2018 Conference on Empirical Methods in Natural Language Processing, pp. 856--861 (2018)

\bibitem{wang2023codet5+}
Wang, Y., Le, H., Gotmare, A.D., Bui, N.D., Li, J., Hoi, S.C.: Codet5+: Open code large language models for code understanding and generation.
\newblock arXiv preprint arXiv:2305.07922  (2023)

\bibitem{wang2013study}
Wang, Z.: Study on the importance of cultural context analysis in machine translation.
\newblock In: Proceedings of The Eighth International Conference on Bio-Inspired Computing: Theories and Applications (BIC-TA), 2013, pp. 29--35. Springer (2013)

\bibitem{wei2019eda}
Wei, J., Zou, K.: Eda: Easy data augmentation techniques for boosting performance on text classification tasks.
\newblock In: Proceedings of the 2019 Conference on Empirical Methods in Natural Language Processing and the 9th International Joint Conference on Natural Language Processing (EMNLP-IJCNLP), pp. 6382--6388 (2019)

\bibitem{wu2020evaluating}
Wu, W., Arendt, D., Volkova, S.: Evaluating neural machine comprehension model robustness to noisy inputs and adversarial attacks.
\newblock CoRR \textbf{abs/2005.00190} (2020).
\newblock \urlprefix\url{https://arxiv.org/abs/2005.00190}

\bibitem{wu2022study}
Wu, X., Xia, Y., Zhu, J., Wu, L., Xie, S., Qin, T.: A study of bert for context-aware neural machine translation.
\newblock Machine Learning \textbf{111}(3), 917--935 (2022)

\bibitem{xu2023autopwn}
Xu, D., Chen, K., Lin, M., Lin, C., Wang, X.: Autopwn: Artifact-assisted heap exploit generation for ctf pwn competitions.
\newblock IEEE Transactions on Information Forensics and Security  (2023)

\bibitem{yang2022dualsc}
Yang, G., Chen, X., Zhou, Y., Yu, C.: Dualsc: Automatic generation and summarization of shellcode via transformer and dual learning.
\newblock In: 2022 IEEE International Conference on Software Analysis, Evolution and Reengineering (SANER), pp. 361--372. IEEE (2022)

\bibitem{yang2023exploitgen}
Yang, G., Zhou, Y., Chen, X., Zhang, X., Han, T., Chen, T.: Exploitgen: Template-augmented exploit code generation based on codebert.
\newblock Journal of Systems and Software \textbf{197}, 111577 (2023)

\bibitem{yang2022natural}
Yang, Z., Shi, J., He, J., Lo, D.: Natural attack for pre-trained models of code.
\newblock In: Proceedings of the 44th International Conference on Software Engineering, pp. 1482--1493 (2022)

\bibitem{DBLP:journals/pacmpl/Yefet0Y20}
Yefet, N., Alon, U., Yahav, E.: Adversarial examples for models of code.
\newblock Proc. {ACM} Program. Lang. \textbf{4}({OOPSLA}), 162:1--162:30 (2020).
\newblock \doi{10.1145/3428230}.
\newblock \urlprefix\url{https://doi.org/10.1145/3428230}

\bibitem{yoo-qi-2021-towards-improving}
Yoo, J.Y., Qi, Y.: Towards improving adversarial training of {NLP} models.
\newblock In: Findings of the Association for Computational Linguistics: EMNLP 2021, pp. 945--956. Association for Computational Linguistics, Punta Cana, Dominican Republic (2021).
\newblock \doi{10.18653/v1/2021.findings-emnlp.81}.
\newblock \urlprefix\url{https://aclanthology.org/2021.findings-emnlp.81}

\bibitem{yu2022data}
Yu, S., Wang, T., Wang, J.: Data augmentation by program transformation.
\newblock Journal of Systems and Software \textbf{190}, 111304 (2022)

\bibitem{zeng2022extensive}
Zeng, Z., Tan, H., Zhang, H., Li, J., Zhang, Y., Zhang, L.: An extensive study on pre-trained models for program understanding and generation.
\newblock In: Proceedings of the 31st ACM SIGSOFT International Symposium on Software Testing and Analysis, pp. 39--51 (2022)

\bibitem{zhang2020training}
Zhang, X., Zhou, Y., Han, T., Chen, T.: Training deep code comment generation models via data augmentation.
\newblock In: Proceedings of the 12th Asia-Pacific Symposium on Internetware, pp. 185--188 (2020)

\bibitem{zhang2024rocoins}
Zhang, Y., Wang, X., Xi, Z., Xia, H., Gui, T., Zhang, Q., Huang, X.: Rocoins: Enhancing robustness of large language models through code-style instructions.
\newblock arXiv preprint arXiv:2402.16431  (2024)

\bibitem{DBLP:journals/tosem/ZhouZSHCG22}
Zhou, Y., Zhang, X., Shen, J., Han, T., Chen, T., Gall, H.C.: Adversarial robustness of deep code comment generation.
\newblock {ACM} Trans. Softw. Eng. Methodol. \textbf{31}(4), 60:1--60:30 (2022).
\newblock \doi{10.1145/3501256}.
\newblock \urlprefix\url{https://doi.org/10.1145/3501256}

\bibitem{10123534}
Zhu, R., Zhang, C.: How robust is a large pre-trained language model for code generationƒ a case on attacking gpt2.
\newblock In: 2023 IEEE International Conference on Software Analysis, Evolution and Reengineering (SANER), pp. 708--712. IEEE Computer Society, Los Alamitos, CA, USA (2023).
\newblock \doi{10.1109/SANER56733.2023.00076}.
\newblock \urlprefix\url{https://doi.ieeecomputersociety.org/10.1109/SANER56733.2023.00076}

\bibitem{zhuo-etal-2023-robustness}
Zhuo, T.Y., Li, Z., Huang, Y., Shiri, F., Wang, W., Haffari, G., Li, Y.F.: On robustness of prompt-based semantic parsing with large pre-trained language model: An empirical study on codex.
\newblock In: Proceedings of the 17th Conference of the European Chapter of the Association for Computational Linguistics, pp. 1090--1102. Association for Computational Linguistics, Dubrovnik, Croatia (2023).
\newblock \doi{10.18653/v1/2023.eacl-main.77}.
\newblock \urlprefix\url{https://aclanthology.org/2023.eacl-main.77}

\end{thebibliography}

\end{document}